\documentclass[letterpaper]{article} 
\usepackage{aaai2027}  

\usepackage[hyphens]{url}  
\usepackage{graphicx} 
\usepackage{natbib}  
\usepackage{caption} 
\usepackage[ruled,vlined]{algorithm2e}
\usepackage{algorithmic}
\usepackage{booktabs}
\usepackage{multirow}
\usepackage{amsmath}

\usepackage{lineno}
\usepackage{xspace}
\newcommand{\method}{\texttt{ReACT-CLIP}\xspace}

\usepackage[table]{xcolor}
\definecolor{bestcell}{rgb}{0.88, 0.95, 0.88}
\usepackage{subcaption}
\usepackage[table]{xcolor}
\usepackage{booktabs}
\usepackage{amssymb}

\definecolor{cvpr25col}{HTML}{1F6FB2}   
\definecolor{cvpr26col}{HTML}{1F6FB2}   
\definecolor{icmlcol}{HTML}{12A035}     
\definecolor{iclrcol}{HTML}{16A085}     
\definecolor{aaaicol}{HTML}{D35400}     
\definecolor{iccvwcol}{HTML}{C0392B}    
\definecolor{arxivcol}{HTML}{7F8C8D}    

\definecolor{bestcell}{HTML}{FFD54F}    
\definecolor{oursbg}{HTML}{E8F5E9}      
\definecolor{oursaccent}{HTML}{1B5E20}  
\definecolor{headerbg}{HTML}{263238}    
\definecolor{rowgray}{HTML}{F5F5F5}     

\definecolor{bestcell}{HTML}{FFD54F}    
\definecolor{oursbg}{HTML}{E8F5E9}      
\definecolor{oursaccent}{HTML}{1B5E20}  
\definecolor{headerbg}{HTML}{263238}    
\definecolor{deltapos}{HTML}{2E7D32}    
\definecolor{deltaneg}{HTML}{C62828}    

\definecolor{grouplabel}{HTML}{757575}  

\definecolor{rulecol}{HTML}{4A4A4A}     

\definecolor{groupbg}{HTML}{E3E7EC}     

\title{\texttt{ReACT-CLIP}: Response-Aware Test-Time Defense for
Vision--Language Models}

\author{
Hashmat Shadab Malik\textsuperscript{\rm 1},
Toluwani Aremu\textsuperscript{\rm 1},
Samuele Poppi\textsuperscript{\rm 1}, \\
Muzammal Naseer\textsuperscript{\rm 2,3},
Salman Khan\textsuperscript{\rm 1,4}
}

\affiliations{
\textsuperscript{\rm 1} Mohamed bin Zayed University of Artificial Intelligence (MBZUAI), UAE\\
\textsuperscript{\rm 2} Khalifa University, UAE\\
\textsuperscript{\rm 3} University of Western Australia, Australia\\
\textsuperscript{\rm 4} Australian National University, Australia\\
\texttt{\{hashmat.malik,toluwani.aremu,samuele.poppi,salman.khan\}@mbzuai.ac.ae}\\
\texttt{muhammadmuzammal.naseer@ku.ac.ae}
}

\begin{document}

\maketitle

\begin{abstract}

Training-free test-time defenses offer a practical way to improve the adversarial robustness of CLIP-style vision--language models without modifying the pretrained model. However, their correction strength is typically fixed for a narrow range of attack budgets, even though the attack budget is unknown at inference and the required correction varies across samples. We show that this mismatch causes existing defenses to degrade sharply as attacks strengthen. We introduce \method{}, a response-conditioned test-time defense that
separately determines how strongly each input should be corrected and whether
defensive intervention is necessary. Our key observation is that the relative increase in CLIP visual-feature drift between low- and high-noise probes provides a graded, sample-specific proxy for correction demand. \method{} maps this relative cross-noise drift to the Gaussian noise scale used to construct a stable, noise-averaged feature anchor, enabling the corrective reach to adapt to each input. To determine whether intervention is necessary, we further observe that clean
inputs retain stable class-probability distributions under  weak spatial
augmentations, whereas adversarial inputs exhibit greater variation. \method{}
quantifies this variation using a prediction-instability score computed by
Jensen--Shannon divergence and combines it with relative cross-noise drift to
form the defensive intervention score. \method{} requires no model or prompt training, and its correction-strength mapping is calibrated once and fixed across datasets and attack budgets. Across 12 downstream datasets, as well as ImageNet and its
distribution-shifted variants, \method{} delivers substantial robustness gains
across diverse attack types and perturbation budgets while largely preserving
clean accuracy. Code is publicly available at \url{https://github.com/HashmatShadab/ReACT-CLIP}.

\end{abstract}

\section{Introduction}
\label{sec:introduction}

CLIP-style vision--language models learn a shared representation space in which images are matched with natural-language descriptions \cite{radford2021learning}. Their strong zero-shot transfer has made frozen CLIP encoders common backbones for classification, retrieval, detection, segmentation, and broader multimodal systems. This reuse also concentrates adversarial risk: a visually small perturbation that corrupts the shared visual representation can affect multiple downstream applications built upon the same encoder \cite{szegedy2013intriguing,carlini2017towards}.

\begin{figure}[t]
\centering
\includegraphics[width=1\linewidth]{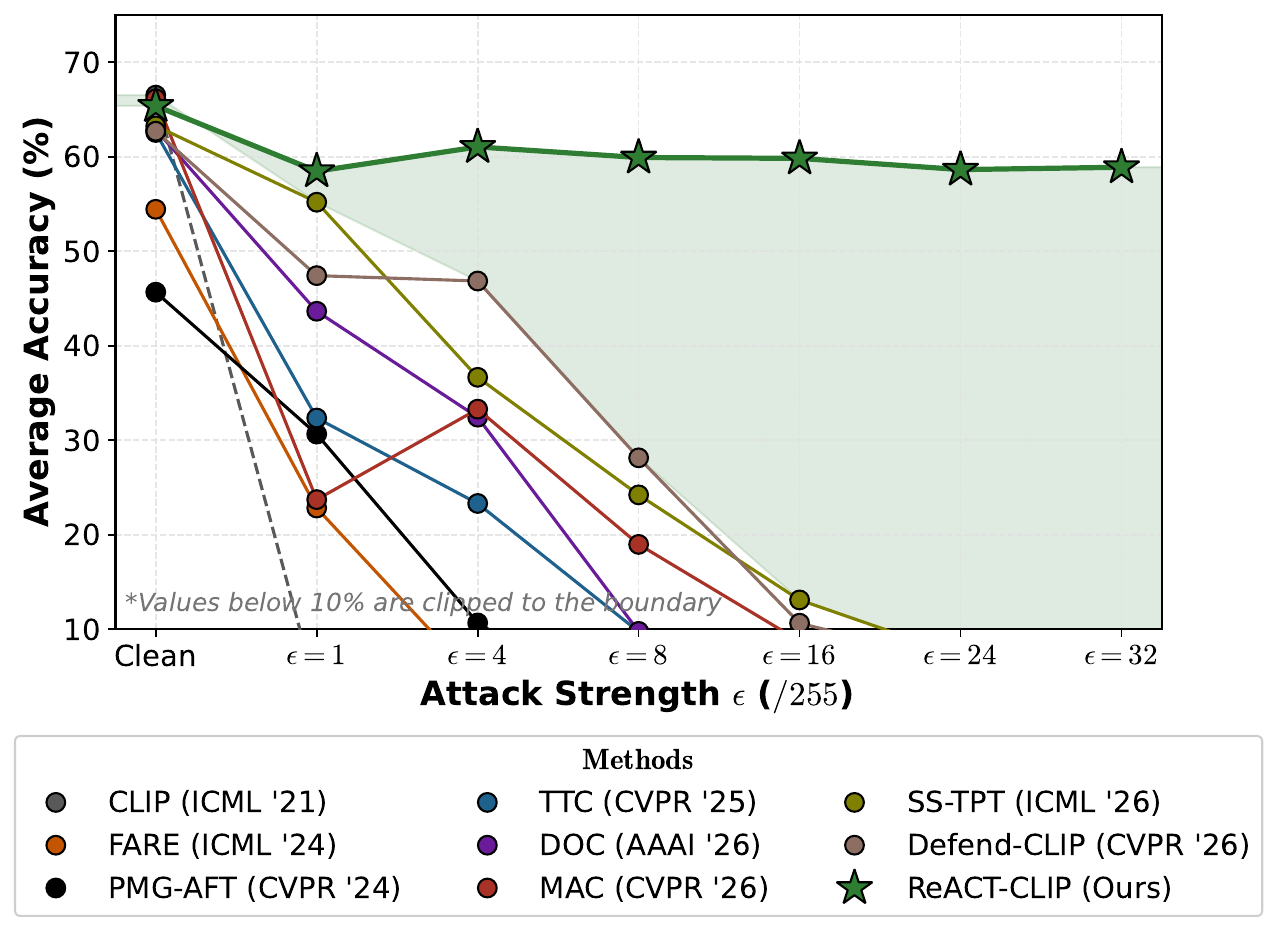}
\vspace{-2em}
\caption{\textbf{Robustness across attack strengths.}
Average accuracy across 12 datasets under PGD-10 attacks from
$\epsilon=1/255$ to $32/255$. Existing defenses degrade sharply as attacks
strengthen, while \method{} maintains stable performance across the full
range of attack budgets.}
\label{fig:teaser}
\vspace{-1.5em}
\end{figure}

Most defenses for CLIP have relied on adversarial training.
\emph{Adversarial finetuning} updates the visual encoder using adversarial examples generated through min--max optimization \cite{mao2022understanding,schlarmann2024robust,wang2024pre,li2024language}, whereas \emph{adversarial prompt tuning} keeps the encoder frozen and optimizes learnable textual tokens against a similar objective \cite{zhang2024adversarial}. Although effective within their training regime, both approaches require repeated adversarial-example generation, can overfit the finetuning distribution, and commonly trade clean accuracy for robustness \cite{zhang2019theoretically}. These costs have motivated training-free defenses that preserve the pretrained model and intervene only at inference.

\begin{figure}[t]
    \centering
    \includegraphics[width=1\linewidth]{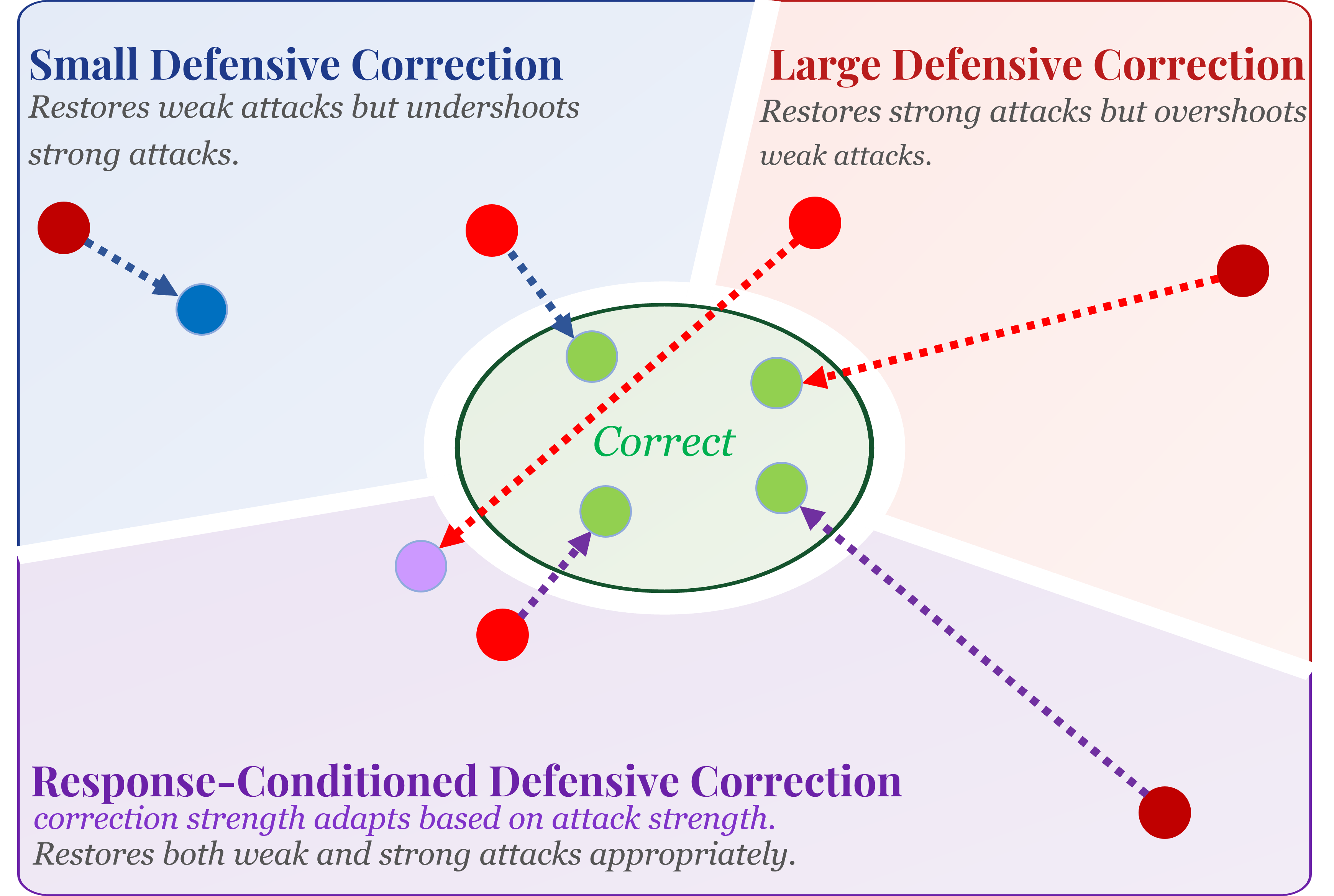}
\vspace{-2em}
\caption{\textbf{Fixed versus response-conditioned correction.}
A small correction undershoots strong attacks, while a large correction
overshoots weak ones. \method{} conditions the defensive correction to each
sample separately, allowing both weak and strong attacks to be
handled appropriately.}
    \label{fig:concept}
        \vspace{-1.5em}
\end{figure}

Existing test-time defenses follow three principal strategies. Prompt-based methods adapt textual prompts for each input \cite{sheng2025r}. Counterattack methods optimize an input-space perturbation intended to reverse the adversarial displacement \cite{xing2025clip,jiang2026diversifying}. Feature-correction methods average the representations of Gaussian-noised views to construct a stable anchor and move the original feature toward or beyond that anchor \cite{tong2025zero}. Although their update directions may depend on the input, the capacity of these interventions is largely set in advance through global hyperparameters, such as the number and step size of optimization updates, the Gaussian anchor scale, or the extrapolation strength. Consequently, a configuration selected for one attack budget may be unsuitable for another.

This fixed-intervention design creates a fundamental limitation at test time:
the defender does not know whether an incoming sample is clean or adversarial
or, if adversarial, the perturbation budget used to generate it as the attacker is not restricted to the budget for which the
defense was configured. Figure~\ref{fig:teaser} shows that representative
adversarially trained, prompt-based, counterattack, and feature-correction
defenses degrade sharply as attack strength increases.
Figure~\ref{fig:concept} illustrates this mismatch: a mild correction may
restore weak attacks but undershoot stronger ones, whereas a stronger correction
Furthermore, applying defensive
correction unconditionally can reduce accuracy on inputs that were already
correctly classified. A successful test-time defense must therefore determine
both \emph{how strongly to correct} and \emph{whether correction should be
applied at all}.

We introduce \textbf{\method{}}---Response-Aware Correction at Test Time for CLIP---a training-free, response-conditioned defense that determines both whether an input requires intervention and how strongly it should be corrected. To estimate the required correction strength, \method{} probes each input with low- and high-magnitude Gaussian noise and measures the resulting visual-feature drift. Prior work has used noise-induced drift as a binary detection signal, since weak noise can leave adversarial features deceptively stable whereas stronger noise exposes greater instability \cite{xing2025clip,brahma2026defending}. We find that drift measured at either noise level alone provides limited information about correction strength. In contrast, the relative increase from low- to high-noise drift exhibits a clearer progression from clean inputs to weak and strong attacks, while also capturing variation among samples generated under the same attack budget. We therefore use this \emph{relative cross-noise drift} as a graded, sample-specific proxy for the strength of defensive correction required.

In feature-correction methods, the anchor-noise scale determines how the feature
anchor is constructed, while the extrapolation strength controls how far the
visual feature moves toward and beyond that anchor. \method{} therefore maps
relative cross-noise drift to a sample-specific correction scale, which is used
to construct a noise-averaged feature anchor from Gaussian-noised views. The
original visual feature is then extrapolated toward and beyond this
response-conditioned anchor to counteract adversarial displacement. Low relative drift yields a smaller noise scale and a mild correction, whereas higher drift yields a larger scale and greater corrective reach. The anchor is thus constructed separately for each input rather than using a globally fixed scale. The mapping is calibrated once and then held fixed across test datasets and attack budgets, without model or prompt optimization.

In addition to defining how strongly an input should be corrected, a reliable activation signal is needed to determine whether defensive intervention should be applied at all, since unnecessary correction can alter otherwise correct predictions on clean inputs. Prior work has used relative cross-noise drift for this purpose, as it effectively separates clean inputs from moderate and strong attacks \cite{xing2025clip,brahma2026defending}. However, clean and weakly attacked inputs overlap in the low-drift regime, making relative drift alone insufficient for reliable activation.We address this ambiguity using the stability of CLIP's class-probability
distribution under  weak spatial augmentations. Clean inputs generally
retain stable predictions, whereas adversarial inputs exhibit greater
variation. \method{} quantifies this variation using a
prediction-instability score computed by Jensen--Shannon divergence and
combines it with relative cross-noise drift to form the defensive intervention
score. Thus, relative cross-noise drift determines \emph{how strongly} to
correct, while the defensive intervention score determines \emph{whether}
correction should be applied, avoiding unnecessary intervention on clean
inputs.
Our contributions are summarized as follows:

\begin{itemize}

    \item \textbf{Response-conditioned anchor construction.}
    We identify the limitation of globally fixed correction strength and use
    relative cross-noise drift as a graded, sample-specific proxy for  correction
    strength. A calibrated correction-strength mapping converts this response
    into the Gaussian noise scale used to construct each input's feature anchor,
    adapting the corrective reach across attack budgets.

\item \textbf{Selective defensive intervention.}
We introduce a defensive intervention score that combines relative cross-noise
drift with a prediction-instability score to improve the overall adversarial detection performance.

    \item \textbf{Extensive robustness evaluation.}
Across 12 downstream datasets, \method{} maintains approximately $60\%$ PGD-10 robust
accuracy from $\epsilon=1/255$ to $16/255$, outperforming the strongest
baseline by up to $46.50\%$ while maintaining high clean accuracy.

\end{itemize}

\section{Method}
\label{sec:method}

\method{} converts noise-anchored feature correction from a globally fixed
operation into a response-conditioned test-time defense. For each input, it
makes two distinct decisions: how strongly the representation should be
corrected and whether correction should be applied. Relative cross-noise drift determines the sample-specific correction scale used
for response-conditioned anchor construction, while the defensive intervention
score combines this drift with the prediction-instability score to determine
whether correction should be applied in the first place.

\subsection{Preliminaries}
\label{sec:preliminaries}

\paragraph{CLIP zero-shot classification.}
CLIP performs zero-shot recognition by matching image and text
representations in a shared embedding space. Given an input image $x$
and class names $\{c_k\}_{k=1}^{K}$, each class is converted into a
natural-language prompt using a prompt template $\Phi(\cdot)$, yielding $\mathcal{P}_k=\Phi(c_k)$. The image and prompts are encoded by the visual encoder $\mathcal{F}_v$ and text encoder $\mathcal{F}_t$, respectively, and their features are $\ell_2$-normalized as
\begin{equation}
\label{eq:clip-features}
f_v(x)
=
\frac{\mathcal{F}_v(x)}
{\|\mathcal{F}_v(x)\|_2},
\qquad
f_t^k
=
\frac{\mathcal{F}_t(\mathcal{P}_k)}
{\|\mathcal{F}_t(\mathcal{P}_k)\|_2}.
\end{equation}
The probability assigned to class $k$ is obtained by applying a
temperature-scaled softmax to the cosine similarities between the visual feature and all textual class features:
\begin{equation}
\label{eq:clip-prediction}
p(y=k\mid x)
=
\frac{
\exp\!\left(
\operatorname{sim}\!\left(f_v(x),f_t^k\right)/\gamma
\right)
}{
\sum_{j=1}^{K}
\exp\!\left(
\operatorname{sim}\!\left(f_v(x),f_t^j\right)/\gamma
\right)
},
\end{equation}
where $\gamma$ is the temperature parameter and
$\operatorname{sim}(\cdot,\cdot)$ denotes cosine similarity. The
predicted label is
$\hat y(x)=\arg\max_k p(y=k\mid x)$.

\paragraph{Adversarial attacks.}
Given a clean image $x$ with label $y$, a white-box adversary attacks the
frozen zero-shot CLIP classifier by constructing
$x_{\mathrm{adv}}=x+\delta$, where $\|\delta\|_p\leq\epsilon$.
Here, $\epsilon$ denotes the perturbation budget, which bounds the
maximum allowable input change under the $\ell_p$ norm. The perturbation
is obtained by maximizing the zero-shot classification loss:
\begin{equation}
\label{eq:attack}
\delta^\star
=
\arg\max_{\|\delta\|_p\leq\epsilon}
\mathcal{L}_{\mathrm{ce}}
\big(p(\cdot\mid x+\delta),y\big),
\qquad
x_{\mathrm{adv}}=x+\delta^\star.
\end{equation}
The attack is generated before the post-hoc test-time defense is applied and therefore does not observe the defensive intervention. The
adversarial objective in Eq.~\ref{eq:attack} is optimized using
projected gradient descent (PGD)~\citep{madry2017towards}.

\paragraph{Noise-anchored feature extrapolation.}
We build on the feature-correction operator used by
AOM~\citep{tong2025zero} and
DefendCLIP~\citep{brahma2026defending}. Given a test input $x$, an
anchor-noise scale $\sigma_{\mathrm{a}}$, and $M$ independent Gaussian
samples, we construct
\begin{equation}
\label{eq:noisy-views}
x_i^{(\sigma_{\mathrm{a}})}
=
x+\sigma_{\mathrm{a}}\eta_i,
\qquad
\eta_i\sim\mathcal{N}(0,I),
\qquad
i=1,\ldots,M.
\end{equation}
Their normalized visual features are averaged to form a noise-averaged
feature anchor:
\begin{equation}
\label{eq:noise-anchor}
f_{\mathrm{anc}}(x;\sigma_{\mathrm{a}})
=
\frac{1}{M}
\sum_{i=1}^{M}
f_v\!\left(x_i^{(\sigma_{\mathrm{a}})}\right).
\end{equation}
The original visual feature is then moved along the direction defined
by this anchor:
\begin{equation}
\label{eq:anchor-extrapolation}
\tilde f_v(x;\sigma_{\mathrm{a}},\alpha)
=
f_v(x)
+
\alpha
\left[
f_{\mathrm{anc}}(x;\sigma_{\mathrm{a}})-f_v(x)
\right].
\end{equation}
When $\alpha=1$, prediction is based directly on the
Gaussian-noise-averaged anchor, corresponding to an ensemble over noisy
views. When $\alpha>1$, the representation is moved beyond the anchor
along the same direction. We refer to this operation as
\emph{anchor-guided extrapolation}.

The defensive correction is jointly affected by two quantities. The
anchor-noise scale $\sigma_{\mathrm{a}}$ determines how the anchor is
constructed, while the extrapolation strength $\alpha$ determines how
far the original feature is moved toward and beyond that anchor.
Existing feature-correction methods fix both quantities globally and
therefore apply the same correction configuration regardless of the
displacement induced in the current test input.

\subsection{\method{}}
\label{sec:response-policy}

\method{} separates test-time defense into two input-specific decisions.
It first estimates the correction demand of the current sample and uses
this estimate to construct a response-conditioned feature anchor.
It then determines whether the resulting correction should be
activated.

\begin{figure}[t]
    \centering
    \includegraphics[width=\linewidth]
    {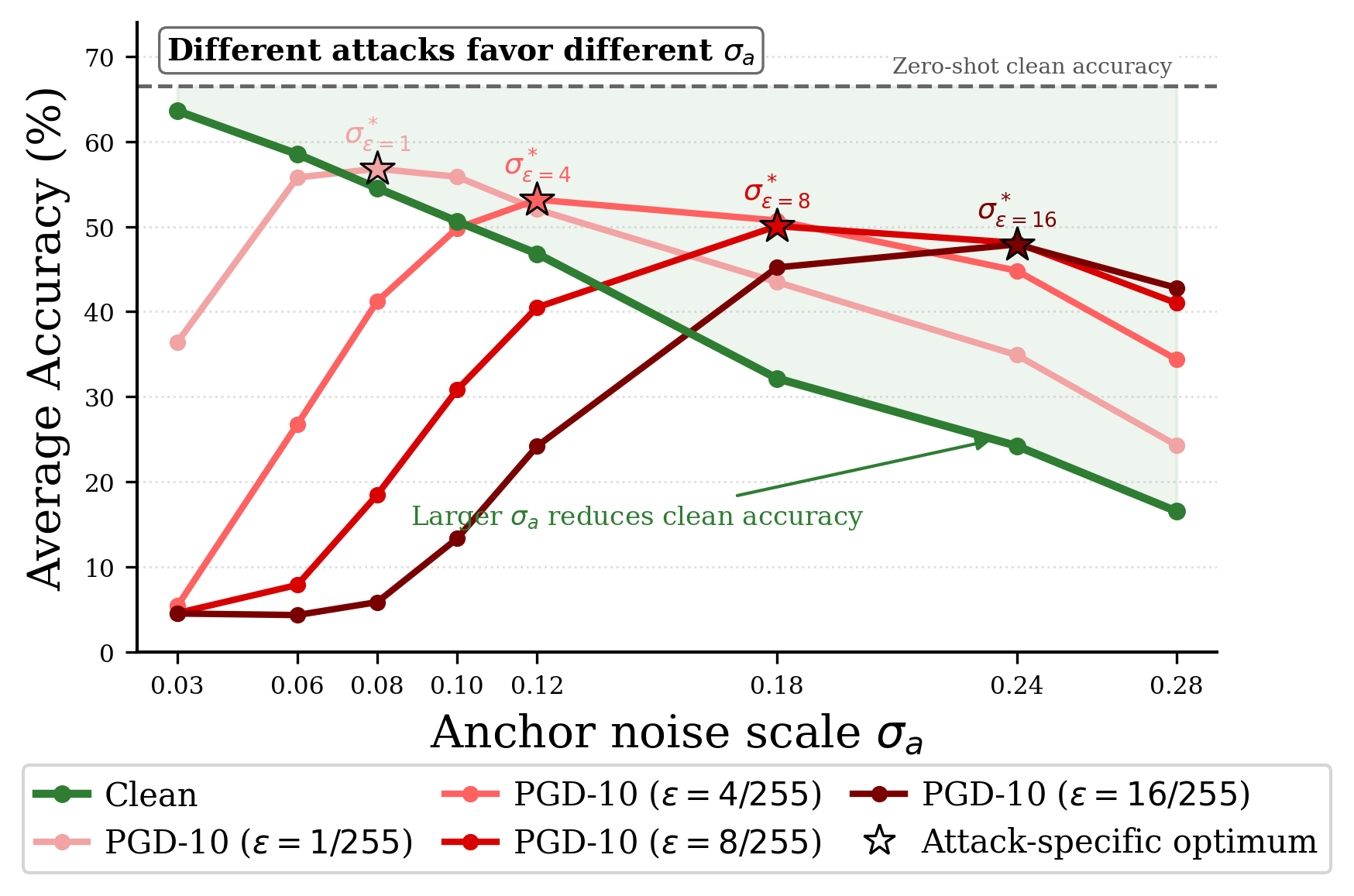}
\vspace{-2em}
    \caption{\textbf{A fixed anchor-noise scale is suboptimal.}
    Under the same extrapolation strength, different PGD-10 perturbation
    budgets achieve their highest defended accuracy at different
    anchor-noise scales, while clean accuracy decreases as the scale
    increases. A single fixed scale therefore cannot preserve clean
    performance while providing sufficient corrective reach across
    perturbation budgets.}
    \label{fig:sigma_tradeoff}
        \vspace{-1.5em}
\end{figure}

\paragraph{Limitation of fixed defensive correction.}
Figure~\ref{fig:sigma_tradeoff} reports defended classification accuracy
averaged across the 12 downstream datasets described in the
experimental setup while varying the anchor-noise scale
$\sigma_{\mathrm{a}}$ under four PGD-10 perturbation budgets. Weak
attacks achieve their highest accuracy at relatively small scales,
whereas stronger attacks require progressively larger scales.
Increasing $\sigma_{\mathrm{a}}$ also reduces clean accuracy and can
degrade robust accuracy at budgets whose optimal correction requires a
smaller scale.

Consequently, no single anchor scale performs well across the complete
perturbation range. A scale selected for clean or weakly attacked inputs
lacks sufficient corrective reach for stronger attacks, while a scale
selected for strong attacks can overcorrect clean or weakly perturbed
inputs. Moreover, samples generated under the same nominal budget can
experience different amounts of displacement in CLIP's visual feature
space. The correction scale should therefore be inferred from each
input rather than fixed globally.

\paragraph{Noise-induced feature dynamics.}
Adapting the correction strength requires an observable signal that
reflects the current input's correction demand. Prior work has shown
that clean and adversarial visual features respond differently to
Gaussian noise. Under weak noise, adversarial features can appear
falsely stable~\citep{xing2025clip}, whereas stronger noise exposes
greater instability and produces larger feature drift
\citep{malik2026beyond}. Defend-CLIP\citep{brahma2026defending} measures the change between these
regimes to decide whether a fixed correction should be activated
. We investigate whether this change can
instead provide a graded, sample-specific signal for controlling the
correction itself.

For each input, we perturb the image using two fixed Gaussian-noise
scales, $\sigma_{\mathrm{low}}<\sigma_{\mathrm{high}}$. For
$s\in\{\mathrm{low},\mathrm{high}\}$, the corresponding noisy view is
\begin{equation}
\label{eq:probe-view}
x^{(\sigma_s)}
=
x+\sigma_s\eta_s,
\qquad
\eta_s\sim\mathcal{N}(0,I).
\end{equation}
We then measure the displacement between its CLIP visual feature and
the original visual feature:
\begin{equation}
\label{eq:probe-drift}
d_s(x)
=
\left\|
f_v\!\left(x^{(\sigma_s)}\right)-f_v(x)
\right\|_2,
\qquad
s\in\{\mathrm{low},\mathrm{high}\}.
\end{equation}

\begin{figure}[t]
    \centering
    \includegraphics[width=\linewidth]
    {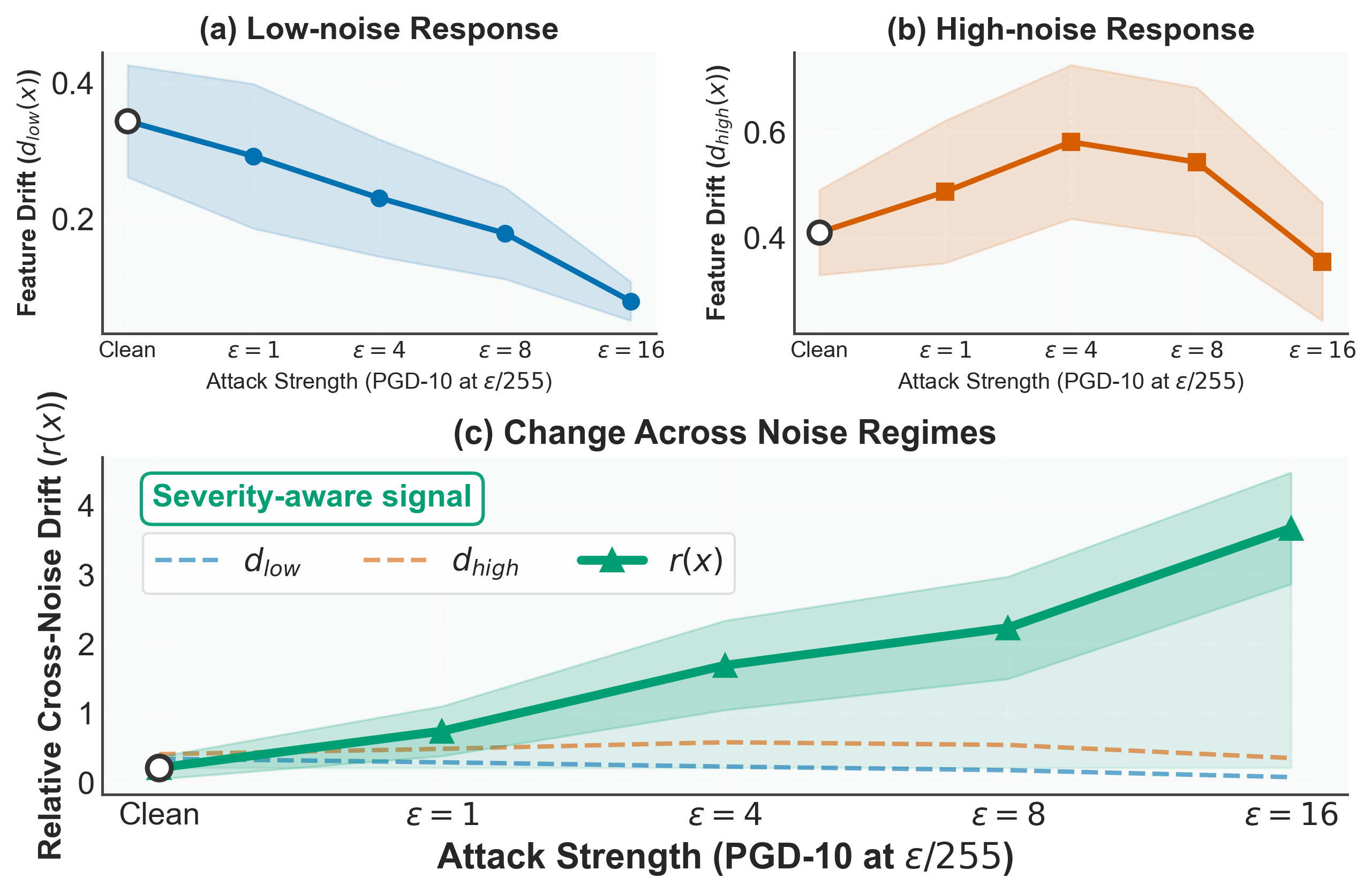}
\vspace{-2em}
    \caption{\textbf{Relative cross-noise drift provides a graded
    response across perturbation budgets.}
   The low- and high-noise probes show different drift magnitudes, but neither
alone exhibits a clear progression from clean inputs to stronger attacks.
Their relative change, $r(x)$, amplifies this variation and increases with the
perturbation budget, providing a sample-specific signal for adapting correction
strength.}
    \label{fig:drift-signal}
        \vspace{-1.5em}
\end{figure}

As shown in Figure~\ref{fig:drift-signal}, neither absolute drift
reliably orders clean inputs and different perturbation budgets. Under
$\sigma_{\mathrm{low}}$, adversarial features can exhibit false
stability. Under $\sigma_{\mathrm{high}}$, their drift increases but
still overlaps across attack budgets. The change between these regimes
is therefore more informative than either measurement alone. We define
the \emph{relative cross-noise drift} as:
\begin{equation}
\label{eq:relative-drift}
r(x)
=
\frac{
d_{\mathrm{high}}(x)-d_{\mathrm{low}}(x)
}{
d_{\mathrm{low}}(x)
}.
\end{equation}
Relative cross-noise drift generally increases from clean inputs to
weakly and strongly attacked inputs. It also varies among samples
generated under the same nominal budget, since a fixed $\epsilon$ does
not displace every sample equally in visual feature space. We therefore
use $r(x)$ as a sample-specific proxy for correction demand rather than
as a direct estimate of the perturbation budget.
\begin{figure}[t]
    \centering
    \includegraphics[width=\linewidth]
    {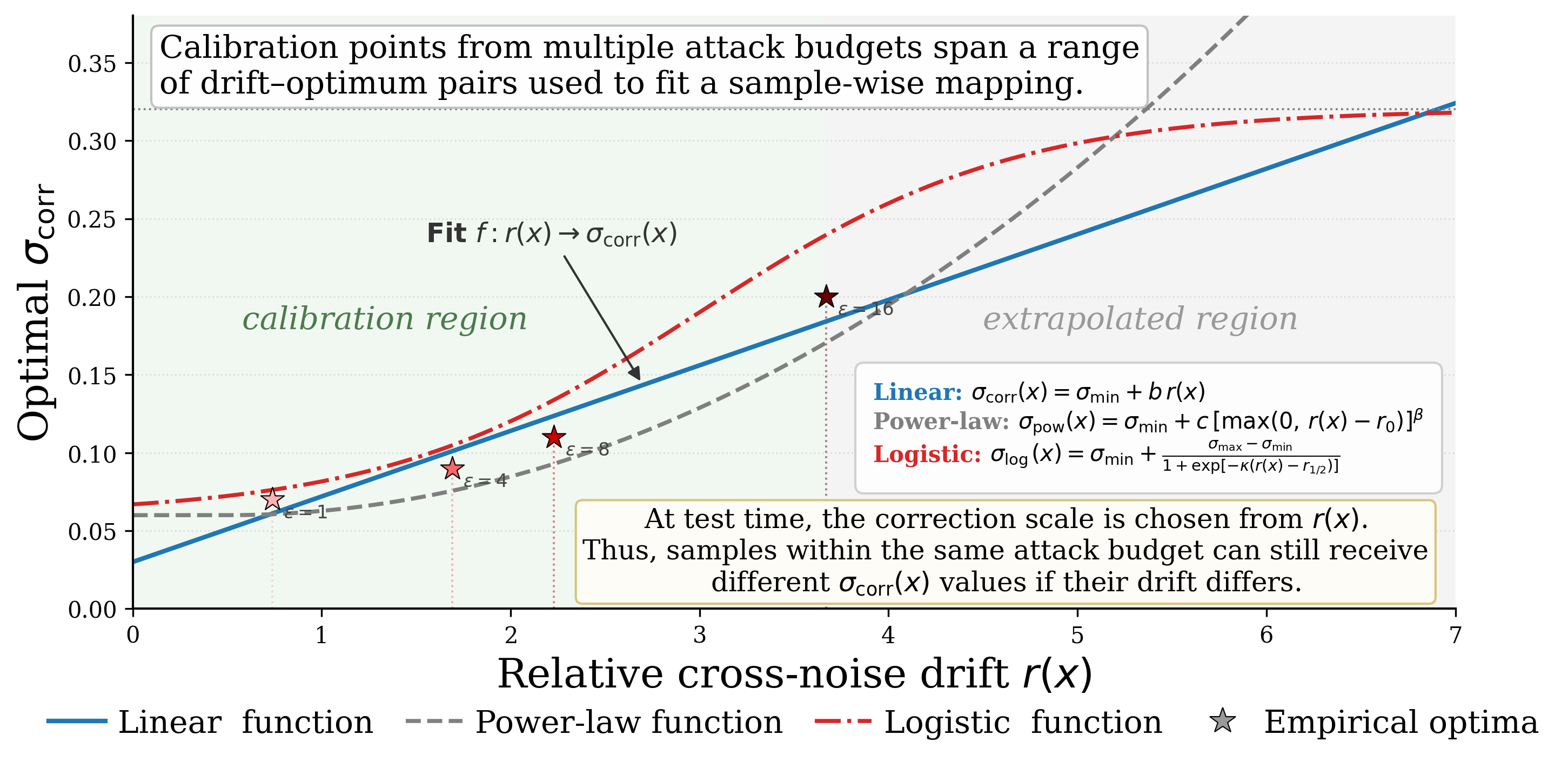}
\vspace{-2em}
    \caption{\textbf{Calibrating relative cross-noise drift to defensive
    correction strength.}
    Each star denotes a calibration pair
    $(\bar r_{\epsilon},\sigma_{\epsilon}^{\star})$, where
    $\bar r_{\epsilon}$ is the average relative cross-noise drift and
    $\sigma_{\epsilon}^{\star}$ is the optimal anchor-noise scale at that budget.}
    \label{fig:drift-calibration}
    \vspace{-1em}
\end{figure}

\paragraph{Response-conditioned anchor construction.}
We next convert relative cross-noise drift into the correction strength
assigned to each input. To calibrate this relationship, we generate
PGD-10 adversarial examples on Caltech256 at
$\epsilon\in\{1,4,8,16\}/255$. For each budget, we compute the average
relative cross-noise drift:
\begin{equation}
\bar r_{\epsilon}
=
\frac{1}{|\mathcal{D}_{\epsilon}|}
\sum_{x\in\mathcal{D}_{\epsilon}}
r(x),
\end{equation}
where $\mathcal{D}_{\epsilon}$ denotes the adversarial samples generated
at budget $\epsilon$. At each budget, we sweep the anchor-noise scale
under a fixed extrapolation strength and select
$\sigma_{\epsilon}^{\star}$, the value that maximizes average defended
classification accuracy. This produces four calibration pairs
$\{(\bar r_{\epsilon},\sigma_{\epsilon}^{\star})\}$, shown as stars in
Figure~\ref{fig:drift-calibration}. These pairs trace a monotone
relationship between the observed response and the anchor scale that
provides the most effective correction at that budget.

We fit a linear mapping to these calibration points:
\begin{equation}
\label{eq:linear-map}
\sigma_{\mathrm{corr}}(x)
=
a+b\,r(x),
\qquad
a=0.03,
\quad
b=0.042,
\end{equation}
where $\sigma_{\mathrm{corr}}(x)$ is the sample-specific correction
scale used to construct the anchor. \method{} adopts this linear mapping
throughout the main results. Figure~\ref{fig:drift-calibration}
additionally shows power-law and logistic alternatives fitted
independently to the same calibration points. We evaluate these
alternatives in the experiment section to determine the behavior of
\method{} across different functions used for mapping this relation.

Substituting
$\sigma_{\mathrm{a}}=\sigma_{\mathrm{corr}}(x)$ into
Eqs.~\ref{eq:noisy-views}--\ref{eq:noise-anchor} gives the
response-conditioned anchor
\begin{equation}
\label{eq:adaptive-anchor}
f_{\mathrm{anc}}(x)
=
\frac{1}{M}
\sum_{i=1}^{M}
f_v\!\left(
x_i^{(\sigma_{\mathrm{corr}}(x))}
\right).
\end{equation}
A small relative cross-noise drift produces a low correction scale and a mild
correction, whereas a larger drift constructs an anchor with greater corrective
reach. Because Eq.~\ref{eq:linear-map} is applied to
each sample separately, \method{} adapts both across perturbation
budgets and among samples generated under the same budget.

\begin{figure}[t]
    \centering
    \includegraphics[width=\linewidth]
    {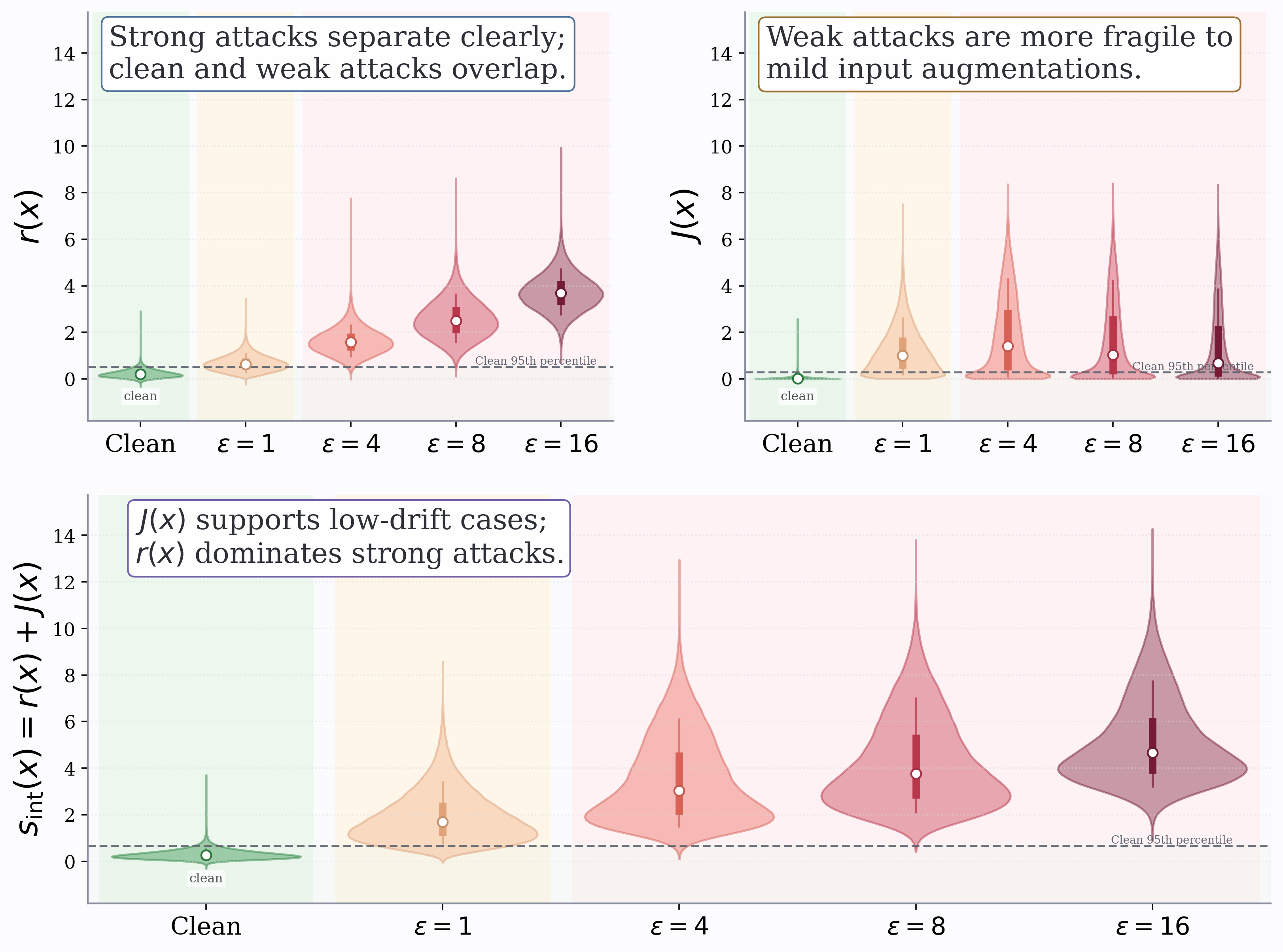}
\vspace{-2em}
\caption{\textbf{Complementary evidence for selective intervention.}
$r(x)$ separates moderate and strong attacks, while  $J(x)$ provides separation between
clean and weaker attacks. Their combination $s_{int}(x)$ forms a more reliable intervention
score for deciding whether correction should be applied.}
    \label{fig:fused-gate}
        \vspace{-2em}
\end{figure}

\paragraph{Selective defensive intervention.}
The correction scale $\sigma_{\mathrm{corr}}(x)$ determines how strongly
an input would be corrected, but not whether correction should be
applied. Under Eq.~\ref{eq:linear-map}, even inputs with low relative
drift receive a mild, non-zero anchor scale.Applying the resulting extrapolation
unconditionally can therefore alter an otherwise correct clean prediction.
Determining whether to intervene is thus distinct from determining correction
strength.

DefendCLIP addresses this intervention decision using the relative drift score,
applying correction when $r(x)$ exceeds a fixed threshold. Although relative
cross-noise drift effectively separates clean inputs from moderate and strong
attacks, clean and weakly attacked inputs overlap in the low-drift regime.
Consequently, a drift-only decision must trade unnecessary intervention on
clean inputs against missed intervention on weak attacks.
We reduce this ambiguity using a prediction-instability score under a weak
spatial augmentations $\mathcal{T}(x)$. Let
\begin{equation}
p_x
=
p(\cdot\mid x),
\qquad
p_{\mathcal{T}}
=
p(\cdot\mid\mathcal{T}(x)),
\qquad
m
=
\frac{p_x+p_{\mathcal{T}}}{2}.
\end{equation}
The prediction-instability score is computed using Jensen--Shannon divergence:
\begin{equation}
\label{eq:js-score}
J(x)
=
\frac{1}{2}
\operatorname{KL}(p_x\|m)
+
\frac{1}{2}
\operatorname{KL}(p_{\mathcal{T}}\|m).
\end{equation}
Clean inputs generally retain stable class probabilities under weak
spatial augmentations, whereas adversarial inputs exhibit greater
prediction variation. Hence, $J(x)$ provides complementary evidence in the weak-attack regime, where
relative cross-noise drift alone remains ambiguous.
As shown in Figure~\ref{fig:fused-gate}, we combine the two signals into the
defensive intervention score:
\begin{equation}
\label{eq:intervention-score}
s_{\mathrm{int}}(x)
=
r(x)+J(x),
\qquad
g(x)
=
\mathbb{I}\!\left[
s_{\mathrm{int}}(x)\geq\tau
\right],
\end{equation}
where $\tau$ is the intervention threshold and $g(x)$ is the correction gate.
The final defended feature is
\begin{equation}
\label{eq:final-feature}
f_{\mathrm{out}}(x)
=
f_v(x)
+
g(x)\,
\alpha
\left[
f_{\mathrm{anc}}(x)-f_v(x)
\right].
\end{equation}
When $g(x)=0$, the original feature is retained; otherwise, the
response-conditioned correction is applied. Relative cross-noise drift therefore
controls correction strength through Eq.~\ref{eq:linear-map} and contributes to the intervention decision as well, while $J(x)$ helps specifically to
distinguish clean samples from adversarial samples of lower strength.

\begin{table*}[!t]
\centering
\renewcommand{\arraystretch}{1.30}
\setlength{\tabcolsep}{4pt}
\resizebox{\textwidth}{!}{
\begin{tabular}{l| >{\color{gray}\columncolor{gray!10}}c|ccc|ccccccc|>{\columncolor{oursbg}}c}
\toprule
\multirow{3}{*}{\bfseries Setting} & \multicolumn{1}{c|}{\scriptsize\itshape\textcolor{gray}{Zero-shot}} & \multicolumn{3}{c|}{\itshape\textcolor{grouplabel}{Adversarial Finetuning}} & \multicolumn{8}{c}{\itshape\textcolor{grouplabel}{Test-time Defense}}  \\
\cmidrule(lr){2-13}
 & \textcolor{gray}{\bfseries CLIP}  & \bfseries TeCoA & \bfseries PMG-AFT & \bfseries FARE & \bfseries Anti-Adv & \bfseries TTC & \bfseries DOC & \bfseries SS-TPT & \bfseries AOM & \bfseries MAC & \bfseries Defend-CLIP & \bfseries \method{} \\
 & {\scriptsize\itshape\textcolor{gray}{(ICML '21)}} &  {\scriptsize\itshape\textcolor{iclrcol}{(ICLR '23)}} & {\scriptsize\itshape\textcolor{cvpr25col}{(CVPR '24)}} & {\scriptsize\itshape\textcolor{icmlcol}{(ICML '24)}} & {\scriptsize\itshape\textcolor{aaaicol}{(AAAI '22)}} & {\scriptsize\itshape\textcolor{cvpr25col}{(CVPR '25)}} & {\scriptsize\itshape\textcolor{aaaicol}{(AAAI '26)}} & {\scriptsize\itshape\textcolor{icmlcol}{(ICML '26)}} & {\scriptsize\itshape\textcolor{cvpr25col}{(CVPR '25)}} & {\scriptsize\itshape\textcolor{cvpr26col}{(CVPR '26)}} & {\scriptsize\itshape\textcolor{cvpr26col}{(CVPR '26)}} & \textcolor{oursaccent}{~$\bigstar$}\\
\midrule
Clean & 66.52  & 43.15 & 45.66 & 54.41 & 61.93 & 62.57 & 62.87 & 63.27 & 50.60 & \cellcolor{bestcell}\textbf{66.14} & 62.70 & \underline{65.38}~\textcolor{deltaneg}{\scriptsize(-0.76)} \\
$\epsilon=1/255$ & 3.40  & 26.40 & 30.67 & 22.83 & 14.15 & 32.35 & 43.65 & 55.19 & \underline{55.86} & 23.72 & 47.42 & \cellcolor{bestcell}\textbf{58.47}~\textcolor{deltapos}{\scriptsize(+2.61)} \\
$\epsilon=4/255$ & 0.07  & 4.73 & 5.63 & 0.87 & 0.67 & 23.31 & 32.44 & 36.66 & \underline{49.62} & 33.29 & 46.85 & \cellcolor{bestcell}\textbf{61.04}~\textcolor{deltapos}{\scriptsize(+11.42)} \\
$\epsilon=8/255$ & 0.03  & 0.40 & 0.47 & 0.13 & 0.22 & 9.74 & 9.77 & 24.22 & \underline{30.83} & 18.98 & 28.14 & \cellcolor{bestcell}\textbf{59.92}~\textcolor{deltapos}{\scriptsize(+29.09)} \\
$\epsilon=16/255$ & 0.02  & 0.12 & 0.13 & 0.08 & 0.08 & 5.52 & 2.08 & 13.10 & \underline{13.33} & 8.75 & 10.64 & \cellcolor{bestcell}\textbf{59.83}~\textcolor{deltapos}{\scriptsize(+46.50)} \\
\bottomrule
\end{tabular}
}
\vspace{-0.5em}
\caption{{Clean and PGD-10 robust accuracy across  attack strengths.} Results are averaged over 12 downstream datasets.
}
\label{tab:main_results_avg}
\vspace{-0.5em}
\end{table*}

\begin{figure*}[!t]
\centering
\setlength{\abovecaptionskip}{4pt}
\setlength{\belowcaptionskip}{0pt}
\begin{minipage}[t]{0.485\textwidth}\vspace{0pt}
\centering
\resizebox{\linewidth}{!}{%
\renewcommand{\arraystretch}{1.3}
\setlength{\tabcolsep}{7pt}
\begin{tabular}{l|l|cc|ccc}
\toprule
\bfseries Anchor construction
& \bfseries Intervention Metric $s_{\mathrm{int}}(x)$
& \bfseries Clean
& \bfseries $\epsilon{=}1$
& \bfseries $\epsilon{=}4$
& \bfseries $\epsilon{=}8$
& \bfseries $\epsilon{=}16$ \\
\midrule

\rowcolor{rowgray}
Fixed anchor ($\sigma_{\mathrm{a}}{=}0.1$)
& $r(x)$
& \underline{62.70}
& 47.42
& 46.85
& 28.14
& 10.64 \\

\rowcolor{oursbg}
$\sigma_{\mathrm{corr}}(x)$
& $r(x)$
& 62.52
& \underline{53.82}
& \cellcolor{bestcell}\textbf{61.76}
& \cellcolor{bestcell}\textbf{60.60}
& \cellcolor{bestcell}\textbf{60.50} \\

\rowcolor{oursbg}
$\sigma_{\mathrm{corr}}(x)$
& $r(x)+J(x)$
& \cellcolor{bestcell}\textbf{65.38}
& \cellcolor{bestcell}\textbf{58.47}
& \underline{61.04}
& \underline{59.92}
& \underline{59.83} \\

\midrule

\multicolumn{2}{l|}{
\itshape\textcolor{gray}{
$\Delta$ Response-conditioned anchor construction
}}
& \textcolor{deltaneg}{$-0.18$}
& \textcolor{deltapos}{$+6.40$}
& \textcolor{deltapos}{$+14.91$}
& \textcolor{deltapos}{$+32.46$}
& \textcolor{deltapos}{$+49.86$} \\

\multicolumn{2}{l|}{
\itshape\textcolor{gray}{
$\Delta$ Defensive intervention score
}}
& \textcolor{deltapos}{$+2.86$}
& \textcolor{deltapos}{$+4.65$}
& \textcolor{deltaneg}{$-0.72$}
& \textcolor{deltaneg}{$-0.68$}
& \textcolor{deltaneg}{$-0.67$} \\

\bottomrule
\end{tabular}
}
\captionof{table}{{Effect of response-conditioned correction and selective defensive
intervention.}}
\label{tab:adaptive_correction}

\vspace{-0.1em}

\includegraphics[width=\linewidth]{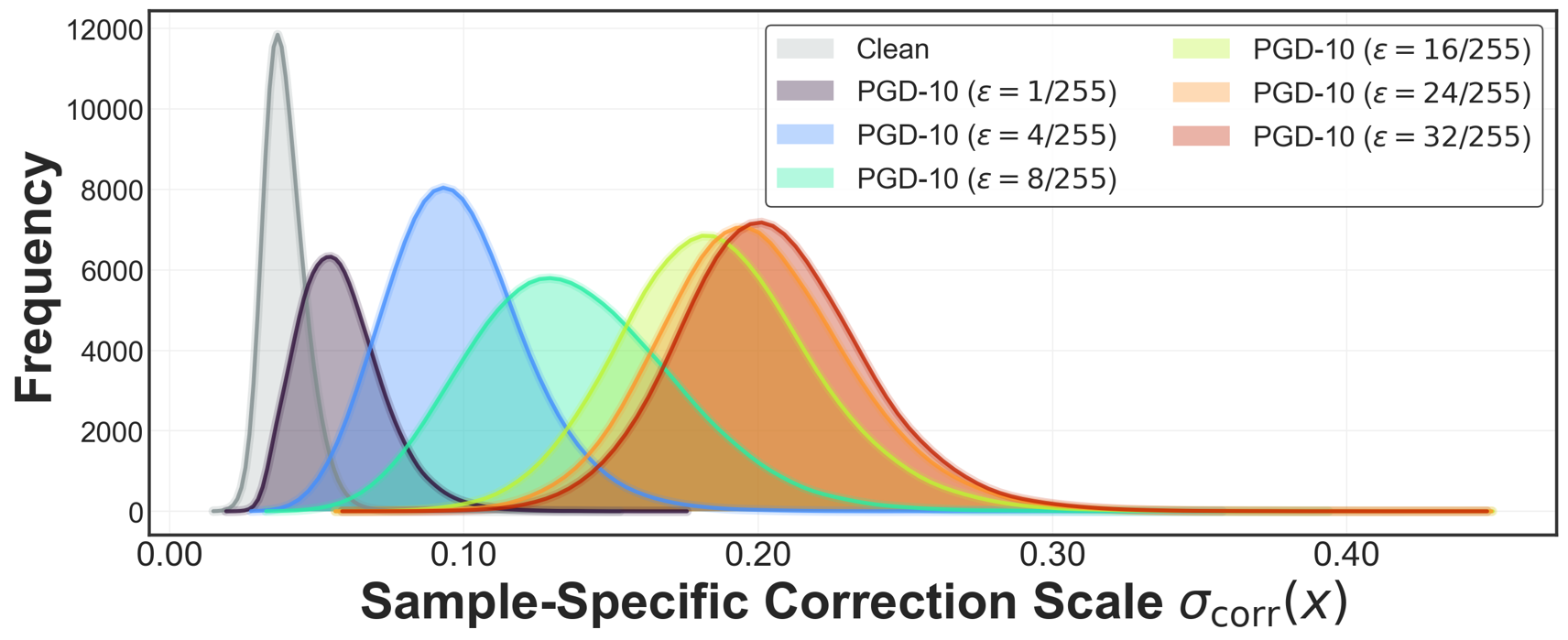}
\vspace{-1.4em}
\captionof{figure}{{Distribution of sample-specific correction scales.}}
\label{fig:sigma_k_grid}

\vspace{-0.2em}

\includegraphics[width=\linewidth]{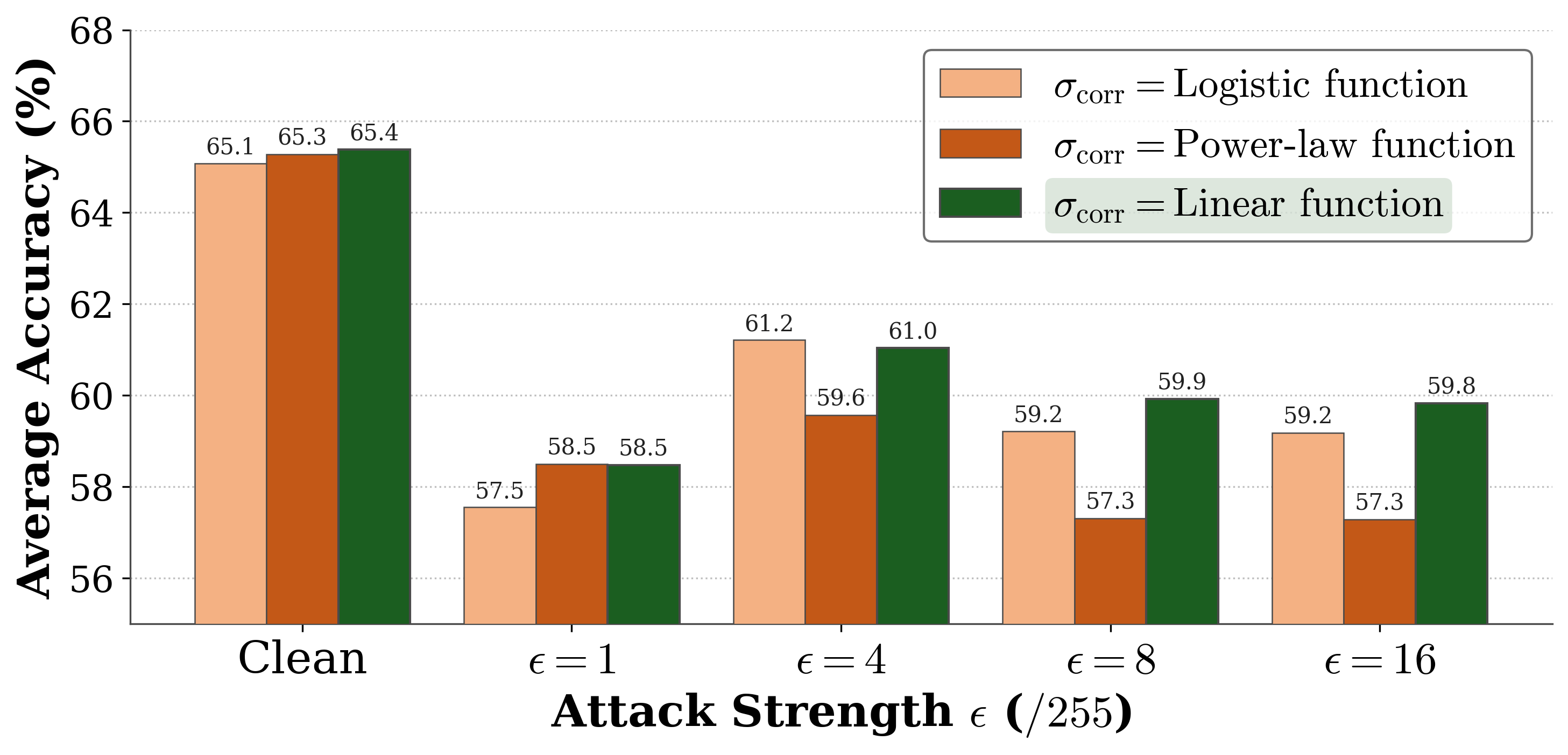}
\vspace{-1.5em}
\captionof{figure}{{Behavior across correction-strength mappings.}}
\label{fig:sigma_mapping_ablation}
\end{minipage}%
\hfill
\begin{minipage}[t]{0.485\textwidth}\vspace{0pt}
\centering
\resizebox{\linewidth}{!}{%
\renewcommand{\arraystretch}{1.3}
\setlength{\tabcolsep}{7pt}
\begin{tabular}{l| cc|ccc}
\toprule
\bfseries Intervention Metric &\bfseries Clean &\bfseries $\epsilon{=}1$ &\bfseries $\epsilon{=}4$ &\bfseries $\epsilon{=}8$ &\bfseries $\epsilon{=}16$ \\
\midrule
\rowcolor{rowgray}
$s_{\mathrm{int}}(x)=r(x), \tau=0.35$ & 62.52 & 53.82 & \cellcolor{bestcell}\textbf{61.76} & \cellcolor{bestcell}\textbf{60.60} & \cellcolor{bestcell}\textbf{60.50} \\
$s_{\mathrm{int}}(x)=r(x)+J(x), \tau=0.9$ & \cellcolor{bestcell}\textbf{66.15} & 55.27 & 60.58 & 59.75 & 59.67 \\
\rowcolor{rowgray}
$s_{\mathrm{int}}(x)=r(x)+J(x), \tau=0.85$ & \underline{66.05} & 56.17 & 60.73 & 59.78 & 59.70 \\
$s_{\mathrm{int}}(x)=r(x)+J(x), \tau=0.80$ & 65.89 & 57.02 & 60.83 & 59.81 & 59.72 \\
\rowcolor{rowgray}
$s_{\mathrm{int}}(x)=r(x)+J(x), \tau=0.75$ & 65.64 & 57.78 & 60.93 & 59.86 & 59.77 \\
\rowcolor{oursbg}
\textcolor{oursaccent}{\textbf{$s_{\mathrm{int}}(x)=r(x)+J(x), \tau=0.70$}~$\bigstar$} & 65.38 & 58.47 & 61.04 & 59.92 & 59.83 \\
$s_{\mathrm{int}}(x)=r(x)+J(x), \tau=0.65$ & 65.05 & \underline{59.05} & 61.14 & 59.99 & 59.89 \\
\rowcolor{rowgray}
$s_{\mathrm{int}}(x)=r(x)+J(x), \tau=0.60$ & 64.54 & \cellcolor{bestcell}\textbf{59.62} & \underline{61.29} & \underline{60.11} & \underline{60.02} \\
\midrule
{\itshape\textcolor{gray}{$\Delta$ (adopted $-$ drift only~$r(x)$)}} & \textcolor{deltapos}{$+2.86$} & \textcolor{deltapos}{$+4.65$} & \textcolor{deltaneg}{$-0.72$} & \textcolor{deltaneg}{$-0.68$} & \textcolor{deltaneg}{$-0.67$} \\
\bottomrule
\end{tabular}}
\captionof{table}{{Sensitivity to the intervention threshold.}}
\label{tab:gate_sweep}

\vspace{1.2em}
\resizebox{\linewidth}{!}{%
\renewcommand{\arraystretch}{1.20}
\setlength{\tabcolsep}{8pt}
\begin{tabular}{l|cc|ccc}
\toprule
\bfseries Intervention Metric & \bfseries Clean & \bfseries $\epsilon{=}1$ & \bfseries $\epsilon{=}4$ & \bfseries $\epsilon{=}8$ & \bfseries $\epsilon{=}16$ \\
\midrule
$s_{\mathrm{int}}(x)=r(x)$ & 74.22 & 85.15 & \cellcolor{bestcell}\textbf{99.87} & \cellcolor{bestcell}\textbf{99.99} & \cellcolor{bestcell}\textbf{100} \\
\rowcolor{oursbg}
\textcolor{oursaccent}{\textbf{$s_{\mathrm{int}}(x)=r(x)+J(x)$}~$\bigstar$} & \cellcolor{bestcell}\textbf{95.44} & \cellcolor{bestcell}\textbf{93.24} & 99.76 & 99.98 & \cellcolor{bestcell}\textbf{100} \\
\midrule
{\itshape\textcolor{gray}{$\Delta$}} & \textcolor{deltapos}{$+$21.22} & \textcolor{deltapos}{$+$8.09} & \textcolor{deltaneg}{$-$0.11} & \textcolor{deltaneg}{$-$0.01} & \textcolor{grouplabel}{0.00} \\
\bottomrule
\end{tabular}}
\captionof{table}{
Percentage of clean and adversarial inputs correctly detected using different intervention scores.
}
\vspace{1.2em}
\label{tab:detection_accuracy}
\includegraphics[width=\linewidth]{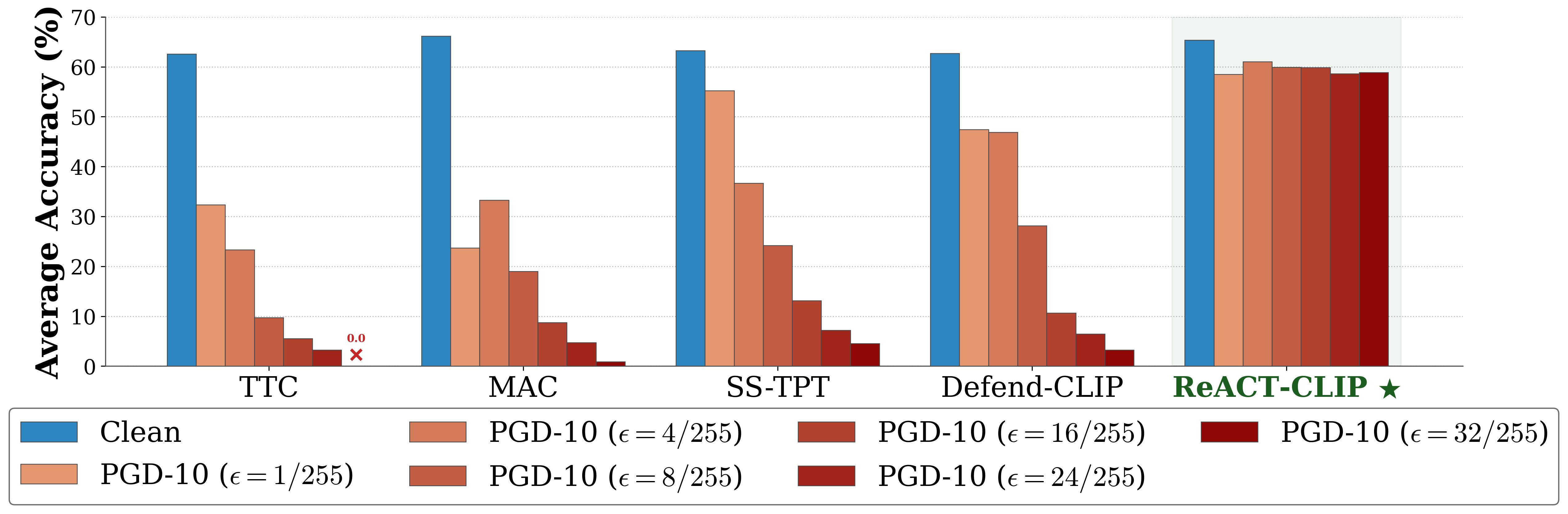}
\captionof{figure}{{Robustness beyond the calibrated mapping range.}}
\label{fig:extended_eps}
\end{minipage}
\vspace{-0.75em}
\end{figure*}
\section{Experiments}
\label{sec:experiments} 

\subsection{Experimental Setup}

\noindent \textbf{Datasets.}
We evaluate on 12 downstream  datasets spanning diverse
domains: Caltech101, Caltech256, CIFAR10, CIFAR100, STL10, OxfordPets,
Flowers102, Food101, StanfordCars, FGVC-Aircraft, DTD, and
EuroSAT~\citep{fei2004learning,griffin2007caltech,krizhevsky2009learning,coates2011analysis,parkhi2012cats,nilsback2008automated,bossard2014food,krause20133d,maji2013fine,cimpoi2014describing,helber2019eurosat}.
 We also evaluate on ImageNet~\citep{deng2009imagenet} and
four OOD variants: ImageNet-A~\citep{hendrycks2021natural},
ImageNet-V2~\citep{recht2019imagenet}, ImageNet-R~\citep{hendrycks2021many},
and ImageNet-Sketch~\citep{wang2019learning}.

\noindent \textbf{Baselines.}
We use zero-shot CLIP as the undefended reference and compare against
adversarially finetuned models-
TeCoA~\citep{mao2022understanding},
PMG-AFT~\citep{wang2024pre}, and
FARE~\citep{schlarmann2024robust} as well as test-time
defenses, including Anti-Adv~\citep{alfarra2022combating},
TTC~\citep{xing2025clip},
DOC~\citep{jiang2026diversifying},
SS-TPT~\citep{kim2026ss},
MAC~\citep{kim2026mac}, AOM~\citep{tong2025zero} and
Defend-CLIP~\citep{brahma2026defending}.

\begin{table*}[t]
\centering
\begin{minipage}[t]{0.32\textwidth}
\centering
\resizebox{\linewidth}{!}{%
\renewcommand{\arraystretch}{1.0}
\setlength{\tabcolsep}{7pt}
\begin{tabular}{l| ccccc}
\toprule
\bfseries Method & \bfseries Clean & \bfseries $\epsilon{=}1$ & \bfseries $\epsilon{=}4$ & \bfseries $\epsilon{=}8$ & \bfseries $\epsilon{=}16$ \\
\midrule
\rowcolor{rowgray}
TTC {\scriptsize\textcolor{cvpr25col}{\textit{(CVPR '25)}}} & 60.12 & 31.45 & 31.33 & 11.44 & 4.07 \\
DOC {\scriptsize\textcolor{aaaicol}{\textit{(AAAI '26)}}} & 60.70 & 43.14 & 43.02 & 15.28 & 4.28 \\
SS-TPT {\scriptsize\textcolor{icmlcol}{\textit{(ICML '26)}}} & 64.18 & \underline{58.10} & 41.14 & 28.12 & \underline{18.23} \\
AOM {\scriptsize\textcolor{cvpr25col}{\textit{(CVPR '25)}}} & 51.86 & 57.60 & \underline{61.07} & \underline{46.73} & 12.63 \\
\rowcolor{rowgray}
MAC {\scriptsize\textcolor{cvpr26col}{\textit{(CVPR '26)}}} & \cellcolor{bestcell}\textbf{65.23} & 27.94 & 48.64 & 28.48 & 8.95 \\
D-CLIP {\scriptsize\textcolor{cvpr26col}{\textit{(CVPR '26)}}} & 62.34 & 44.72 & 58.02 & 43.87 & 9.77 \\
\midrule
\rowcolor{oursbg}
\textcolor{oursaccent}{\textbf{\method{}}~$\bigstar$} & \underline{64.59} & \cellcolor{bestcell}\textbf{58.99} & \cellcolor{bestcell}\textbf{63.67} & \cellcolor{bestcell}\textbf{63.37} & \cellcolor{bestcell}\textbf{64.25} \\
\midrule
{\itshape\textcolor{gray}{$\Delta$}} & \textcolor{deltaneg}{$-0.64$} & \textcolor{deltapos}{$+0.89$} & \textcolor{deltapos}{$+2.60$} & \textcolor{deltapos}{$+16.64$} & \textcolor{deltapos}{$+46.02$} \\
\bottomrule
\end{tabular}}
\vspace{-1em}
\caption{Clean and PGD-100 robust accuracy (\%) averaged across 
datasets.}
\label{tab:pgd100}
    \vspace{-1.5em}
\end{minipage}%
\hfill
\begin{minipage}[t]{0.32\textwidth}
\centering
\resizebox{\linewidth}{!}{%
\renewcommand{\arraystretch}{1.0}
\setlength{\tabcolsep}{7pt}
\begin{tabular}{l| ccccc}
\toprule
\bfseries Method & \bfseries Clean & \bfseries $\epsilon{=}1$ & \bfseries $\epsilon{=}4$ & \bfseries $\epsilon{=}8$ & \bfseries $\epsilon{=}16$ \\
\midrule
\rowcolor{rowgray}
TTC {\scriptsize\textcolor{cvpr25col}{\textit{(CVPR '25)}}} & 60.12 & 30.06 & 22.83 & 8.72 & 5.35 \\
DOC {\scriptsize\textcolor{aaaicol}{\textit{(AAAI '26)}}} & 60.70 & 40.67 & 33.24 & 11.18 & 6.17 \\
SS-TPT {\scriptsize\textcolor{icmlcol}{\textit{(ICML '26)}}} & 64.18 & 53.45 & 34.26 & 21.24 & 12.20 \\
AOM {\scriptsize\textcolor{cvpr25col}{\textit{(CVPR '25)}}} & 51.86 & \cellcolor{bestcell}\textbf{56.77} & \underline{52.23} & \underline{30.87} & \underline{16.42} \\
\rowcolor{rowgray}
MAC {\scriptsize\textcolor{cvpr26col}{\textit{(CVPR '26)}}} & \cellcolor{bestcell}\textbf{65.23} & 23.28 & 32.45 & 17.06 & 10.09 \\
Defend-CLIP {\scriptsize\textcolor{cvpr26col}{\textit{(CVPR '26)}}} & 62.34 & 47.62 & 49.30 & 28.01 & 13.55 \\
\midrule
\rowcolor{oursbg}
\textcolor{oursaccent}{\textbf{\method{}}~$\bigstar$} & \underline{64.59} & \underline{56.40} & \cellcolor{bestcell}\textbf{59.69} & \cellcolor{bestcell}\textbf{57.29} & \cellcolor{bestcell}\textbf{57.19} \\
\midrule
{\itshape\textcolor{gray}{$\Delta$}} & \textcolor{deltaneg}{$-0.64$} & \textcolor{deltaneg}{$-0.37$} & \textcolor{deltapos}{$+7.46$} & \textcolor{deltapos}{$+26.42$} & \textcolor{deltapos}{$+40.77$} \\
\bottomrule
\end{tabular}}
\vspace{-1em}
\caption{Clean and C\&W-10 robust accuracy (\%) averaged across 
datasets.}
\label{tab:cw}
    \vspace{-1.5em}
\end{minipage}%
\hfill
\begin{minipage}[t]{0.32\textwidth}
\centering
\resizebox{\linewidth}{!}{%
\renewcommand{\arraystretch}{1.0}
\setlength{\tabcolsep}{7pt}
\begin{tabular}{l| ccccc}
\toprule
\bfseries Method & \bfseries Clean & \bfseries $\epsilon{=}1$ & \bfseries $\epsilon{=}4$ & \bfseries $\epsilon{=}8$ & \bfseries $\epsilon{=}16$ \\
\midrule
\rowcolor{rowgray}
TTC {\scriptsize\textcolor{cvpr25col}{\textit{(CVPR '25)}}} & 60.12 & 32.44 & 23.56 & 10.01 & 6.12 \\
DOC {\scriptsize\textcolor{aaaicol}{\textit{(AAAI '26)}}} & 60.70 & 44.92 & 35.12 & 9.20 & 3.41 \\
SS-TPT {\scriptsize\textcolor{icmlcol}{\textit{(ICML '26)}}} & 64.18 & 55.64 & 37.12 & 25.16 & 13.56 \\
AOM {\scriptsize\textcolor{cvpr25col}{\textit{(CVPR '25)}}} & 51.86 & \underline{56.41} & \underline{52.47} & \underline{31.73} & \underline{17.14} \\
\rowcolor{rowgray}
MAC {\scriptsize\textcolor{cvpr26col}{\textit{(CVPR '26)}}} & \cellcolor{bestcell}\textbf{65.23} & 25.21 & 34.81 & 18.11 & 10.13 \\
Defend-CLIP {\scriptsize\textcolor{cvpr26col}{\textit{(CVPR '26)}}} & 62.34 & 47.81 & 48.90 & 29.40 & 12.81 \\
\midrule
\rowcolor{oursbg}
\textcolor{oursaccent}{\textbf{\method{}}~$\bigstar$} & \underline{64.59} & \cellcolor{bestcell}\textbf{57.85} & \cellcolor{bestcell}\textbf{60.84} & \cellcolor{bestcell}\textbf{58.89} & \cellcolor{bestcell}\textbf{58.96} \\
\midrule
{\itshape\textcolor{gray}{$\Delta$}} & \textcolor{deltaneg}{$-0.64$} & \textcolor{deltapos}{$+1.44$} & \textcolor{deltapos}{$+8.37$} & \textcolor{deltapos}{$+27.16$} & \textcolor{deltapos}{$+41.82$} \\
\bottomrule
\end{tabular}}
\vspace{-1em}
\caption{Clean and AutoAttack robust accuracy (\%) averaged across
 datasets. }
\label{tab:autoattack}
    \vspace{-1.5em}
\end{minipage}
\end{table*}

\noindent \textbf{Attack and evaluation protocol.}
We adopt the adversarial threat setting established by prior training-free
test-time defenses, including TTC, DOC, MAC, AOM, and DefendCLIP. Adversarial
examples are crafted against the frozen CLIP model before any test-time defense is applied. This setting reflects practical deployment scenarios in which
attackers have access to the open-source model but areunaware of the user-side defensive intervention.
Our main evaluation reports results averaged across the 12 downstream
datasets under PGD-10 attacks with an $\ell_\infty$ constraint and
$\epsilon\in\{1,4,8,16\}/255$. Unless stated otherwise, all reported results
refer to this setting. We additionally evaluate robustness under PGD-100,
C\&W-10, and AutoAttack.
Following standard adversarial evaluation, we craft adversarial examples only
for clean samples that are correctly classified by the undefended CLIP model.
Samples that are initially misclassified remain unperturbed and are
passed as such through the defense.

\noindent \textbf{Implementation details.}
All experiments use CLIP ViT-B/32 for evaluation. For each
baseline, we use the available official implementation and its recommended configuration (details provided in the appendix). AOM and
Defend-CLIP use a fixed anchor scale $\sigma_{\mathrm{a}}{=}0.1$ and
extrapolation strength $\alpha{=}1.2$. AOM applies correction unconditionally,
whereas Defend-CLIP uses the drift-only intervention score
$s_{\mathrm{int}}(x)=r(x)$ with $\tau{=}0.35$. We reproduce AOM using the
DefendCLIP implementation with the correction gate disabled.
For \method{}, the linear mapping from $r(x)$ to the sample-specific correction
scale $\sigma_{\mathrm{corr}}(x)$ is calibrated once on Caltech256 and fixed
across all datasets and attacks. We use $M{=}10$ noisy views,
$\sigma_{\mathrm{low}}{=}0.02$, $\sigma_{\mathrm{high}}{=}0.05$, and the
defensive intervention score $s_{\mathrm{int}}(x)=r(x)+J(x)$ with threshold
$\tau{=}0.7$.  Inspired by
AugMix~\citep{hendrycks2019augmix}, the weak stochastic augmentation
$\mathcal{T}$ used to compute $J(x)$ applies a mild random affine
transformation followed by independently sampled Gaussian blur, additive
Gaussian noise, and color jitter. Since $\sigma_{\mathrm{corr}}(x)$ constructs an anchor better
matched to each input, it reduces the risk of overshooting associated with a
globally fixed anchor and enables reliable use of a stronger global
extrapolation strength $\alpha{=}2.0$.

\subsection{Results}

\noindent \textbf{Comparison with Existing Defenses.}
Table~\ref{tab:main_results_avg} compares \method{} with zero-shot CLIP,
adversarially finetuned models, and training-free test-time defenses under
PGD-10, averaged across 12 datasets. The strongest baseline falls from
$55.86\%$ at $\epsilon{=}1/255$ to $13.33\%$ at
$\epsilon{=}16/255$, whereas \method{} maintains approximately $60\%$
robust accuracy across all budgets. It surpasses the strongest baseline by
$2.61\%$, $11.42\%$, $29.09\%$, and $46.50\%$ at
$\epsilon\in\{1,4,8,16\}/255$, respectively, while remaining within
$0.76\%$ of the strongest defense on clean inputs. The slightly lower
performance at $\epsilon{=}1/255$ than at larger budgets reflects the greater
difficulty of distinguishing weak adversarial examples from clean inputs for
selective intervention, causing some weak attacks to remain uncorrected.

\noindent \textbf{Ablation on Response-Conditioned Anchor Construction and
Selective Defensive Intervention.}
Table~\ref{tab:adaptive_correction} isolates the two contributions of
\method{}. Response-conditioned anchor construction improves robust accuracy
over the fixed-anchor configuration by $6.40\%$, $14.91\%$, $32.46\%$, and
$49.86\%$ as $\epsilon$ increases from $1/255$ to $16/255$. Replacing the drift-only score with the
defensive intervention score further improves clean and
$\epsilon{=}1/255$ accuracy by $2.86\%$ and $4.65\%$, respectively, while
maintaining approximately $60\%$ robustness at stronger budgets. These results
show that response-conditioned anchor construction drives robustness across
perturbation budgets, whereas selective defensive intervention primarily
protects clean inputs and improves performance in the weak-attack regime.

\noindent \textbf{Behavior of the Correction-Strength Mapping.}
Figure~\ref{fig:sigma_k_grid} shows the distribution of sample-specific
correction scales $\sigma_{\mathrm{corr}}(x)$ aggregated across 12
downstream datasets. Clean and weakly attacked inputs generally receive
smaller scales, while the distributions shift toward larger scales as the
perturbation budget increases. The spread within each budget further shows
that \method{} adapts anchor construction to individual samples rather
than assigning a single scale based on the perturbation budget alone.
Figure~\ref{fig:sigma_mapping_ablation} compares performance when the
correction-strength mapping is instantiated as a linear, logistic, or
power-law function of relative cross-noise drift. All three mappings yield
similarly strong gains, with our adopted  linear mapping providing
high robust accuracy across all the attack strengths.

\noindent \textbf{Intervention Threshold and Detection Performance.}
Table~\ref{tab:gate_sweep} sweeps the intervention threshold $\tau$ under
the defensive intervention score $s_{\mathrm{int}}(x)=r(x)+J(x)$ and
identifies $\tau{=}0.7$ as a balanced operating point: lowering $\tau$
below this value trades clean accuracy for further gains at
$\epsilon{=}1/255$, while accuracy at $\epsilon\geq4/255$ remains largely
unaffected by the choice of threshold. Using this threshold,
Table~\ref{tab:detection_accuracy} reports how reliably the correction
gate separates clean from adversarial inputs, restricting evaluation to
samples correctly classified by undefended CLIP, since these are the only
inputs for which a corresponding adversarial example exists under our
protocol. Relative to the drift-only score $r(x)$, the defensive
intervention score raises the clean pass-through rate from $74.22\%$ to
$95.44\%$ and the weak-attack detection rate from $85.15\%$ to $93.24\%$,
confirming that $J(x)$ resolves much of the ambiguity between clean and
weakly attacked inputs that relative cross-noise drift alone leaves
unresolved. This gain comes at negligible cost to stronger attacks, with
detection remaining at or above $99.76\%$ for $\epsilon\geq4/255$.

\noindent \textbf{Robustness Beyond the Calibrated Mapping Range.}
Figure~\ref{fig:extended_eps} evaluates PGD-10 at
$\epsilon\in\{24,32\}/255$, beyond the budgets
$\epsilon\in\{1,4,8,16\}/255$ used to calibrate the linear
correction-strength mapping. While existing defenses continue to degrade as
the perturbation budget increases, \method{} maintains stable robustness in
this extrapolated region.

\noindent \textbf{Robustness and Generalization.}
We further evaluate \method{} under PGD-100, C\&W-10, and AutoAttack on nine
downstream datasets (excluding EuroSAT, Caltech256, and Food101). Tables~\ref{tab:pgd100},
\ref{tab:cw}, and~\ref{tab:autoattack} show that \method{} remains stable as
attack strength increases, surpassing the strongest baseline at
$\epsilon{=}16/255$ by $46.02\%$, $40.77\%$, and $41.82\%$, respectively.
Table~\ref{tab:ood} evaluates ImageNet and its four out-of-distribution
variants, where \method{} sustains $41.34$--$42.97\%$ robust accuracy and
improves over the strongest baseline by up to $31.46\%$.

\begin{table}[h]
\centering
\resizebox{\linewidth}{!}{%
\renewcommand{\arraystretch}{1.0}
\setlength{\tabcolsep}{7pt}
\begin{tabular}{l| ccccc}
\toprule
\bfseries Method & \bfseries Clean & \bfseries $\epsilon{=}1$ & \bfseries $\epsilon{=}4$ & \bfseries $\epsilon{=}8$ & \bfseries $\epsilon{=}16$ \\
\midrule
\rowcolor{rowgray}
SS-TPT {\scriptsize\textcolor{icmlcol}{\textit{(ICML '26)}}} & \cellcolor{bestcell}\textbf{50.67} & 39.75 & 25.99 & \underline{18.20} & \underline{10.38} \\
DOC {\scriptsize\textcolor{aaaicol}{\textit{(AAAI '26)}}} & 35.66 & 33.16 & 22.41 & 7.21 & 4.15 \\
TTC {\scriptsize\textcolor{cvpr25col}{\textit{(CVPR '25)}}} & 35.60 & 24.06 & 13.75 & 4.63 & 2.85 \\
AOM {\scriptsize\textcolor{cvpr25col}{\textit{(CVPR '25)}}} & 39.07 & \underline{40.50} & \underline{31.32} & {12.68} & {6.79} \\
\rowcolor{rowgray}
MAC {\scriptsize\textcolor{cvpr26col}{\textit{(CVPR '26)}}} & \underline{47.49} & 15.65 & 23.03 & 10.73 & 5.03 \\
Defend-CLIP {\scriptsize\textcolor{cvpr26col}{\textit{(CVPR '26)}}} & 43.56 & 34.74 & 28.41 & 10.05 & 4.16 \\
\midrule
\rowcolor{oursbg}
\textcolor{oursaccent}{\textbf{\method{}}~$\bigstar$} & 45.78 & \cellcolor{bestcell}\textbf{41.34} & \cellcolor{bestcell}\textbf{42.97} & \cellcolor{bestcell}\textbf{41.50} & \cellcolor{bestcell}\textbf{41.84} \\
\midrule
{\itshape\textcolor{gray}{$\Delta$}} & \textcolor{deltaneg}{$-4.89$} & \textcolor{deltapos}{$+0.84$} & \textcolor{deltapos}{$+11.65$} & \textcolor{deltapos}{$+23.30$} & \textcolor{deltapos}{$+31.46$} \\
\bottomrule
\end{tabular}}
\vspace{-0.5em}
\caption{Clean and PGD-10 robust accuracy (\%) averaged on ImageNet and its four
out-of-distribution variants.}
\label{tab:ood}
\vspace{-1.5em}
\end{table}

\section{Discussion and Conclusion}
\label{sec:discussion_conclusion}

We introduced \method{}, a training-free test-time defense that reframes
correction as a per-input decision rather than a globally fixed
intervention: for each sample, it separately determines how strongly to
correct and whether intervention is necessary, without requiring any
knowledge of the attack budget at inference. Relative cross-noise drift
determines the sample-specific correction scale used for
response-conditioned anchor construction, while the defensive
intervention score combines this drift with the prediction-instability
score to decide whether correction should be applied. This enables
effective adaptation across a wide range of perturbation budgets.
\method{} consistently outperforms existing defenses across attack
objectives, perturbation budgets, and distribution shifts, surpassing the
strongest baseline by up to $46.50$ percentage points while maintaining
competitive clean accuracy. The appendix provides a detailed related-work
discussion, complete per-dataset results, component-wise ablations, an
analysis of relative cross-noise drift across datasets, and the
associated computational cost. Overall, these results establish our
approach as an effective training-free test-time defense for CLIP.

\bibliography{aaai2027}

@inproceedings{zhang2019theoretically,
  title={Theoretically principled trade-off between robustness and accuracy},
  author={Zhang, Hongyang and Yu, Yaodong and Jiao, Jiantao and Xing, Eric and El Ghaoui, Laurent and Jordan, Michael},
  booktitle={International conference on machine learning},
  pages={7472--7482},
  year={2019},
  organization={PMLR}
}

@article{zhang2024vision,
  title={Vision-language models for vision tasks: A survey},
  author={Zhang, Jingyi and Huang, Jiaxing and Jin, Sheng and Lu, Shijian},
  journal={IEEE transactions on pattern analysis and machine intelligence},
  volume={46},
  number={8},
  pages={5625--5644},
  year={2024},
  publisher={IEEE}
}

@inproceedings{fares2026mirrorcheck,
  title={Mirrorcheck: Efficient adversarial defense for vision-language models},
  author={Fares, Samar and Ziu, Klea and Aremu, Toluwani and Durasov, Nikita and Tak{\'a}{\v{c}}, Martin and Fua, Pascal and Laptev, Ivan and Nandakumar, Karthik},
  booktitle={Proceedings of the IEEE/CVF Conference on Computer Vision and Pattern Recognition},
  pages={496--506},
  year={2026}
}

@inproceedings{jiang2026diversifying,
  title={Diversifying Counterattacks: Orthogonal Exploration for Robust CLlP Inference},
  author={Jiang, Chengze and Dong, Minjing and Shi, Xinli and Gui, Jie},
  booktitle={Proceedings of the AAAI Conference on Artificial Intelligence},
  volume={40},

  pages={5359--5368},
  year={2026}
}

@inproceedings{kim2026mac,
  title={When CLIP Sees More, It Fights Back Harder: Multi-View Guided Adaptive Counterattacks for Test-Time Adversarial Robustness},
  author={Kim, Sunoh and Um, Daeho},
  booktitle={Proceedings of the IEEE/CVF Conference on Computer Vision and Pattern Recognition (CVPR)},
  year={2026}
}

@article{yu2022coca,
  title={Coca: Contrastive captioners are image-text foundation models},
  author={Yu, Jiahui and Wang, Zirui and Vasudevan, Vijay and Yeung, Legg and Seyedhosseini, Mojtaba and Wu, Yonghui},
  journal={arXiv preprint arXiv:2205.01917},
  year={2022}
}

@article{saharia2022photorealistic,
  title={Photorealistic text-to-image diffusion models with deep language understanding},
  author={Saharia, Chitwan and Chan, William and Saxena, Saurabh and Li, Lala and Whang, Jay and Denton, Emily L and Ghasemipour, Kamyar and Gontijo Lopes, Raphael and Karagol Ayan, Burcu and Salimans, Tim and others},
  journal={Advances in neural information processing systems},
  volume={35},
  pages={36479--36494},
  year={2022}
}

@inproceedings{ramesh2021zero,
  title={Zero-shot text-to-image generation},
  author={Ramesh, Aditya and Pavlov, Mikhail and Goh, Gabriel and Gray, Scott and Voss, Chelsea and Radford, Alec and Chen, Mark and Sutskever, Ilya},
  booktitle={International conference on machine learning},
  pages={8821--8831},
  year={2021},
  organization={Pmlr}
}

@inproceedings{radford2021learning,
  title={Learning transferable visual models from natural language supervision},
  author={Radford, Alec and Kim, Jong Wook and Hallacy, Chris and Ramesh, Aditya and Goh, Gabriel and Agarwal, Sandhini and Sastry, Girish and Askell, Amanda and Mishkin, Pamela and Clark, Jack and others},
  booktitle={International conference on machine learning},
  pages={8748--8763},
  year={2021},
  organization={PmLR}
}

@article{shin2022reco,
  title={Reco: Retrieve and co-segment for zero-shot transfer},
  author={Shin, Gyungin and Xie, Weidi and Albanie, Samuel},
  journal={Advances in Neural Information Processing Systems},
  volume={35},
  pages={33754--33767},
  year={2022}
}

@inproceedings{zhou2022extract,
  title={Extract free dense labels from clip},
  author={Zhou, Chong and Loy, Chen Change and Dai, Bo},
  booktitle={European conference on computer vision},
  pages={696--712},
  year={2022},
  organization={Springer}
}

@inproceedings{zhang2023simple,
  title={A simple framework for open-vocabulary segmentation and detection},
  author={Zhang, Hao and Li, Feng and Zou, Xueyan and Liu, Shilong and Li, Chunyuan and Yang, Jianwei and Zhang, Lei},
  booktitle={Proceedings of the IEEE/CVF International Conference on Computer Vision},
  pages={1020--1031},
  year={2023}
}

@inproceedings{zhao2022exploiting,
  title={Exploiting unlabeled data with vision and language models for object detection},
  author={Zhao, Shiyu and Zhang, Zhixing and Schulter, Samuel and Zhao, Long and Vijay Kumar, BG and Stathopoulos, Anastasis and Chandraker, Manmohan and Metaxas, Dimitris N},
  booktitle={European conference on computer vision},
  pages={159--175},
  year={2022},
  organization={Springer}
}

@inproceedings{cai2023semantic,
  title={Semantic-enhanced image clustering},
  author={Cai, Shaotian and Qiu, Liping and Chen, Xiaojun and Zhang, Qin and Chen, Longteng},
  booktitle={Proceedings of the AAAI conference on artificial intelligence},
  volume={37},
  pages={6869--6878},
  year={2023}
}

@inproceedings{li2024language,
  title={Language-driven anchors for zero-shot adversarial robustness},
  author={Li, Xiao and Zhang, Wei and Liu, Yining and Hu, Zhanhao and Zhang, Bo and Hu, Xiaolin},
  booktitle={Proceedings of the IEEE/CVF Conference on Computer Vision and Pattern Recognition},
  pages={24686--24695},
  year={2024}
}

@article{mao2022understanding,
  title={Understanding zero-shot adversarial robustness for large-scale models},
  author={Mao, Chengzhi and Geng, Scott and Yang, Junfeng and Wang, Xin and Vondrick, Carl},
  journal={arXiv preprint arXiv:2212.07016},
  year={2022}
}

@article{schlarmann2024robust,
  title={Robust clip: Unsupervised adversarial fine-tuning of vision embeddings for robust large vision-language models},
  author={Schlarmann, Christian and Singh, Naman Deep and Croce, Francesco and Hein, Matthias},
  journal={arXiv preprint arXiv:2402.12336},
  year={2024}
}

@inproceedings{wang2024pre,
  title={Pre-trained model guided fine-tuning for zero-shot adversarial robustness},
  author={Wang, Sibo and Zhang, Jie and Yuan, Zheng and Shan, Shiguang},
  booktitle={Proceedings of the IEEE/CVF conference on computer vision and pattern recognition},
  pages={24502--24511},
  year={2024}
}

@inproceedings{zhang2024adversarial,
  title={Adversarial prompt tuning for vision-language models},
  author={Zhang, Jiaming and Ma, Xingjun and Wang, Xin and Qiu, Lingyu and Wang, Jiaqi and Jiang, Yu-Gang and Sang, Jitao},
  booktitle={European conference on computer vision},
  pages={56--72},
  year={2024},
  organization={Springer}
}

@article{zhou2024few,
  title={Few-shot adversarial prompt learning on vision-language models},
  author={Zhou, Yiwei and Xia, Xiaobo and Lin, Zhiwei and Han, Bo and Liu, Tongliang},
  journal={Advances in Neural Information Processing Systems},
  volume={37},
  pages={3122--3156},
  year={2024}
}

@article{szegedy2013intriguing,
  title={Intriguing properties of neural networks},
  author={Szegedy, Christian and Zaremba, Wojciech and Sutskever, Ilya and Bruna, Joan and Erhan, Dumitru and Goodfellow, Ian and Fergus, Rob},
  journal={arXiv preprint arXiv:1312.6199},
  year={2013}
}

@inproceedings{papernot2016limitations,
  title={The limitations of deep learning in adversarial settings},
  author={Papernot, Nicolas and McDaniel, Patrick and Jha, Somesh and Fredrikson, Matt and Celik, Z Berkay and Swami, Ananthram},
  booktitle={2016 IEEE European symposium on security and privacy (EuroS\&P)},
  pages={372--387},
  year={2016},
  organization={IEEE}
}

@inproceedings{carlini2017towards,
  title={Towards evaluating the robustness of neural networks},
  author={Carlini, Nicholas and Wagner, David},
  booktitle={2017 ieee symposium on security and privacy (sp)},
  pages={39--57},
  year={2017},
  organization={Ieee}
}

@inproceedings{croce2020reliable,
  title={Reliable evaluation of adversarial robustness with an ensemble of diverse parameter-free attacks},
  author={Croce, Francesco and Hein, Matthias},
  booktitle={International conference on machine learning},
  pages={2206--2216},
  year={2020},
  organization={PMLR}
}

@inproceedings{tong2025zero,
  title={On the Zero-shot Adversarial Robustness of Vision-Language Models: A Truly Zero-shot and Training-free Approach},
  author={Tong, Baoshun and Lai, Hanjiang and Pan, Yan and Yin, Jian},
  booktitle={Proceedings of the Computer Vision and Pattern Recognition Conference},
  pages={19921--19930},
  year={2025}
}

@article{shu2022test,
  title={Test-time prompt tuning for zero-shot generalization in vision-language models},
  author={Shu, Manli and Nie, Weili and Huang, De-An and Yu, Zhiding and Goldstein, Tom and Anandkumar, Anima and Xiao, Chaowei},
  journal={Advances in Neural Information Processing Systems},
  volume={35},
  pages={14274--14289},
  year={2022}
}

@inproceedings{feng2023diverse,
  title={Diverse data augmentation with diffusions for effective test-time prompt tuning},
  author={Feng, Chun-Mei and Yu, Kai and Liu, Yong and Khan, Salman and Zuo, Wangmeng},
  booktitle={Proceedings of the IEEE/CVF International Conference on Computer Vision},
  pages={2704--2714},
  year={2023}
}

@article{abdul2023align,
  title={Align your prompts: Test-time prompting with distribution alignment for zero-shot generalization},
  author={Abdul Samadh, Jameel and Gani, Mohammad Hanan and Hussein, Noor and Khattak, Muhammad Uzair and Naseer, Muhammad Muzammal and Shahbaz Khan, Fahad and Khan, Salman H},
  journal={Advances in Neural Information Processing Systems},
  volume={36},
  pages={80396--80413},
  year={2023}
}

@inproceedings{xing2025clip,
  title={Clip is strong enough to fight back: Test-time counterattacks towards zero-shot adversarial robustness of clip},
  author={Xing, Songlong and Zhao, Zhengyu and Sebe, Nicu},
  booktitle={Proceedings of the Computer Vision and Pattern Recognition Conference},
  pages={15172--15182},
  year={2025}
}

@article{zhao2023evaluating,
  title={On evaluating adversarial robustness of large vision-language models},
  author={Zhao, Yunqing and Pang, Tianyu and Du, Chao and Yang, Xiao and Li, Chongxuan and Cheung, Ngai-Man Man and Lin, Min},
  journal={Advances in Neural Information Processing Systems},
  volume={36},
  pages={54111--54138},
  year={2023}
}

@inproceedings{sheng2025r,
  title={R-TPT: Improving Adversarial Robustness of Vision-Language Models through Test-Time Prompt Tuning},
  author={Sheng, Lijun and Liang, Jian and Wang, Zilei and He, Ran},
  booktitle={Proceedings of the Computer Vision and Pattern Recognition Conference},
  pages={29958--29967},
  year={2025}
}

@article{wang2025double,
  title={Double visual defense: Adversarial pre-training and instruction tuning for improving vision-language model robustness},
  author={Wang, Zeyu and Xie, Cihang and Bartoldson, Brian and Kailkhura, Bhavya},
  journal={arXiv preprint arXiv:2501.09446},
  year={2025}
}

@article{madry2017towards,
  title={Towards deep learning models resistant to adversarial attacks},
  author={Madry, Aleksander and Makelov, Aleksandar and Schmidt, Ludwig and Tsipras, Dimitris and Vladu, Adrian},
  journal={arXiv preprint arXiv:1706.06083},
  year={2017}
}

@article{malik2025robust,
  title={Robust-llava: On the effectiveness of large-scale robust image encoders for multi-modal large language models},
  author={Malik, Hashmat Shadab and Shamshad, Fahad and Naseer, Muzammal and Nandakumar, Karthik and Khan, Fahad and Khan, Salman},
  journal={arXiv preprint arXiv:2502.01576},
  year={2025}
}

@article{nie2022diffusion,
  title={Diffusion models for adversarial purification},
  author={Nie, Weili and Guo, Brandon and Huang, Yujia and Xiao, Chaowei and Vahdat, Arash and Anandkumar, Anima},
  journal={arXiv preprint arXiv:2205.07460},
  year={2022}
}

@article{wang2022guided,
  title={Guided diffusion model for adversarial purification},
  author={Wang, Jinyi and Lyu, Zhaoyang and Lin, Dahua and Dai, Bo and Fu, Hongfei},
  journal={arXiv preprint arXiv:2205.14969},
  year={2022}
}

@inproceedings{deng2009imagenet,
  title = {ImageNet: A Large-Scale Hierarchical Image Database},
  author = {Deng, Jia and Dong, Wei and Socher, Richard and Li, Li-Jia and Li, Kai and Fei-Fei, Li},
  booktitle = {2009 IEEE Conference on Computer Vision and Pattern Recognition},
  pages = {248--255},
  year = {2009},
  organization = {IEEE}
}

@inproceedings{recht2019imagenet,
  title = {Do ImageNet Classifiers Generalize to ImageNet?},
  author = {Recht, Benjamin and Roelofs, Rebecca and Schmidt, Ludwig and Shankar, Vaishaal},
  booktitle = {International Conference on Machine Learning},
  pages = {5389--5400},
  year = {2019},
  organization = {PMLR}
}

@article{wang2019learning,
  title = {Learning Robust Global Representations by Penalizing Local Predictive Power},
  author = {Wang, Haohan and Ge, Songwei and Lipton, Zachary and Xing, Eric P},
  journal = {Advances in Neural Information Processing Systems},
  volume = {32},
  year = {2019}
}

@inproceedings{hendrycks2021natural,
  title = {Natural Adversarial Examples},
  author = {Hendrycks, Dan and Zhao, Kevin and Basart, Steven and Steinhardt, Jacob and Song, Dawn},
  booktitle = {Proceedings of the IEEE/CVF Conference on Computer Vision and Pattern Recognition},
  pages = {15262--15271},
  year = {2021}
}

@inproceedings{hendrycks2021many,
  title = {The Many Faces of Robustness: A Critical Analysis of Out-of-Distribution Generalization},
  author = {Hendrycks, Dan and Basart, Steven and Mu, Norman and Kadavath, Saurav and Wang, Frank and Dorundo, Evan and Desai, Rahul and Zhu, Tyler and Parajuli, Samyak and Guo, Mike and others},
  booktitle = {Proceedings of the IEEE/CVF International Conference on Computer Vision},
  pages = {8340--8349},
  year = {2021}
}

@inproceedings{fei2004learning,
  title = {Learning Generative Visual Models from Few Training Examples: An Incremental Bayesian Approach Tested on 101 Object Categories},
  author = {Fei-Fei, Li and Fergus, Rob and Perona, Pietro},
  booktitle = {2004 Conference on Computer Vision and Pattern Recognition Workshop},
  pages = {178--178},
  year = {2004},
  organization = {IEEE}
}

@inproceedings{parkhi2012cats,
  title = {Cats and Dogs},
  author = {Parkhi, Omkar M and Vedaldi, Andrea and Zisserman, Andrew and Jawahar, CV},
  booktitle = {2012 IEEE Conference on Computer Vision and Pattern Recognition},
  pages = {3498--3505},
  year = {2012},
  organization = {IEEE}
}

@inproceedings{krause20133d,
  title = {3D Object Representations for Fine-Grained Categorization},
  author = {Krause, Jonathan and Stark, Michael and Deng, Jia and Fei-Fei, Li},
  booktitle = {Proceedings of the IEEE International Conference on Computer Vision Workshops},
  pages = {554--561},
  year = {2013}
}

@inproceedings{nilsback2008automated,
  title = {Automated Flower Classification Over a Large Number of Classes},
  author = {Nilsback, Maria-Elena and Zisserman, Andrew},
  booktitle = {2008 Sixth Indian Conference on Computer Vision, Graphics \& Image Processing},
  pages = {722--729},
  year = {2008},
  organization = {IEEE}
}

@article{maji2013fine,
  title = {Fine-Grained Visual Classification of Aircraft},
  author = {Maji, Subhransu and Rahtu, Esa and Kannala, Juho and Blaschko, Matthew and Vedaldi, Andrea},
  journal = {arXiv preprint arXiv:1306.5151},
  year = {2013}
}

@inproceedings{cimpoi2014describing,
  title = {Describing Textures in the Wild},
  author = {Cimpoi, Mircea and Maji, Subhransu and Kokkinos, Iasonas and Mohamed, Sammy and Vedaldi, Andrea},
  booktitle = {Proceedings of the IEEE Conference on Computer Vision and Pattern Recognition},
  pages = {3606--3613},
  year = {2014}
}

@article{helber2019eurosat,
  title = {Eurosat: A Novel Dataset and Deep Learning Benchmark for Land Use and Land Cover Classification},
  author = {Helber, Patrick and Bischke, Benjamin and Dengel, Andreas and Borth, Damian},
  journal = {IEEE Journal of Selected Topics in Applied Earth Observations and Remote Sensing},
  volume = {12},
  number = {7},
  pages = {2217--2226},
  year = {2019},
  publisher = {IEEE}
}

@inproceedings{brahma2026defending,
  title={Defending CLIP via Noise-Induced Feature Dynamics for Training-Free, Zero-shot Adversarial Robustness},
  author={Brahma, Debarshi and Biswas, Soma},
  booktitle={Proceedings of the IEEE/CVF Conference on Computer Vision and Pattern Recognition},
  pages={656--665},
  year={2026}
}

@inproceedings{alfarra2022combating,
  title={Combating adversaries with anti-adversaries},
  author={Alfarra, Motasem and P{\'e}rez, Juan C and Thabet, Ali and Bibi, Adel and Torr, Philip HS and Ghanem, Bernard},
  booktitle={Proceedings of the AAAI Conference on Artificial Intelligence},
  volume={36},

  pages={5992--6000},
  year={2022}
}

@inproceedings{perez2021enhancing,
  title={Enhancing adversarial robustness via test-time transformation ensembling},
  author={P{\'e}rez, Juan C and Alfarra, Motasem and Jeanneret, Guillaume and Rueda, Laura and Thabet, Ali and Ghanem, Bernard and Arbel{\'a}ez, Pablo},
  booktitle={Proceedings of the IEEE/CVF International Conference on Computer Vision},
  pages={81--91},
  year={2021}
}

@article{wu2021attacking,
  title={Attacking adversarial attacks as a defense},
  author={Wu, Boxi and Pan, Heng and Shen, Li and Gu, Jindong and Zhao, Shuai and Li, Zhifeng and Cai, Deng and He, Xiaofei and Liu, Wei},
  journal={arXiv preprint arXiv:2106.04938},
  year={2021}
}

@inproceedings{bossard2014food,
  title={Food-101--mining discriminative components with random forests},
  author={Bossard, Lukas and Guillaumin, Matthieu and Van Gool, Luc},
  booktitle={European conference on computer vision},
  pages={446--461},
  year={2014},
  organization={Springer}
}

@inproceedings{coates2011analysis,
  title={An analysis of single-layer networks in unsupervised feature learning},
  author={Coates, Adam and Ng, Andrew and Lee, Honglak},
  booktitle={Proceedings of the fourteenth international conference on artificial intelligence and statistics},
  pages={215--223},
  year={2011},
  organization={JMLR Workshop and Conference Proceedings}
}

@article{griffin2007caltech,
  title={Caltech-256 object category dataset},
  author={Griffin, Gregory and Holub, Alex and Perona, Pietro},
  year={2007},
  journal={Caltech Technical Report}
}

@article{krizhevsky2009learning,
  title={Learning multiple layers of features from tiny images},
  author={Krizhevsky, Alex and Hinton, Geoffrey and others},
  year={2009},
  publisher={Toronto, ON, Canada},
  journal={Technical Report}
}

@article{kim2026ss,
  title={SS-TPT: Stability and Suitability-Guided Test-Time Prompt Tuning for Adversarially Robust Vision-Language Models},
  author={Kim, Sunoh and Um, Daeho},
  journal={arXiv preprint arXiv:2606.06943},
  year={2026}
}

@article{malik2026beyond,
  title={Beyond False Stability: High-Noise Drift Gating for Test-Time Adversarial Defenses in Vision-Language Models},
  author={Malik, Hashmat Shadab and Naseer, Muzammal and Khan, Salman},
  journal={arXiv preprint arXiv:2606.03730},
  year={2026}
}

@article{hendrycks2019augmix,
  title={Augmix: A simple data processing method to improve robustness and uncertainty},
  author={Hendrycks, Dan and Mu, Norman and Cubuk, Ekin D and Zoph, Barret and Gilmer, Justin and Lakshminarayanan, Balaji},
  journal={arXiv preprint arXiv:1912.02781},
  year={2019}
}
\appendix

\section*{Overview of the Appendix}
\label{sec:appendix-overview}
\addcontentsline{toc}{section}{Overview of the Appendix}

This appendix provides extended related work, full experimental detail,
per-dataset and per-mapping breakdowns of every result summarized in the
main paper, a formal statement of the threat model, and the full
specification of \method{}. It is organized as follows:

\begin{itemize}
\item \textbf{Appendix~\ref{sec:related_work} (Related Work)} situates
\method{} relative to adversarial finetuning, prompt-based and
counterattack test-time defenses, and the noise-response literature that
\method{} builds on most directly, and clarifies how \method{}
repurposes noise-response drift from a binary trigger into a graded,
sample-specific signal.

\item \textbf{Appendix~\ref{app:datasets} (Datasets)} describes the 12
downstream datasets and the ImageNet-based evaluation suite (ImageNet
plus four out-of-distribution variants), including domain coverage,
class counts, and test-set sizes (Table~\ref{tab:app_datasets}).

\item \textbf{Appendix~\ref{app:extended_main_results} (Detailed and
Extended Results Corresponding to the Main Paper)} reports the
per-dataset breakdown underlying  main-paper table
(Tables~\ref{tab:main_table1_per_dataset_results}--\ref{tab:main_table8_results_per_dataset}),
an extended comparison against three further baselines
(Table~\ref{tab:main_table1_results_with_other_baselines}), and results
at $\epsilon\in\{24,32\}/255$, beyond the range used to calibrate the
correction-strength mapping
(Table~\ref{tab:main_table1_results_with_eps24_and_eps32}).

\item \textbf{Appendix~\ref{sec:appendix_hyperparameter_ablation}
(Ablation of \method{}'s Hyperparameters)} sweeps the intervention
threshold $\tau$, the number of anchor views $M$, and the extrapolation
strength $\alpha$ independently, motivating the main configuration
($\tau{=}0.7$, $M{=}10$, $\alpha{=}2.0$) and showing in particular that
the fixed $\alpha{=}1.2$ used by AOM and Defend-CLIP is far too weak
once paired with \method{}'s response-conditioned anchor.

\item \textbf{Appendix~\ref{sec:appendix_cross_noise_drift} (Analysis of
Relative Cross-Noise Drift)} examines the correction-demand signal
$r(x)$ separately on every dataset, under both PGD-10 and PGD-100, and
shows that its monotone relationship with attack strength holds
consistently across the full evaluation suite and both optimizers.

\item \textbf{Appendix~\ref{sec:appendix_correction_scale_distributions}
(Analysis of Correction-Scale Distributions)} compares the resulting
correction-scale distributions produced by the linear, logistic, and
power-law mappings, aggregated across datasets and broken down
per-dataset, under PGD-10, PGD-100, and on the ImageNet-based datasets.

\item \textbf{Appendix~\ref{sec:appendix_defensive_intervention}
(Analysis of Defensive Intervention)} reports per-dataset distributions
of $r(x)$, $J(x)$, and the fused intervention score $s_{\mathrm{int}}(x)$,
confirming that the two signals play complementary roles consistently
across every dataset in the suite.

\item \textbf{Appendix~\ref{sec:appendix-threat-scope} (Adversarial
Threat Scope)} states the attacker's goal, knowledge, and capability in
full, specifies the attack suite and perturbation budgets evaluated, and
details the evaluation protocol used throughout the paper.

\item \textbf{Appendix~\ref{app:label_leakage} (Label Leakage)}
identifies and quantifies a reporting artifact specific to
feature-correction defenses, in which attacking already-misclassified
clean samples lets the defensive correction claim credit for higher robustness, and shows this artifact is not corrected for by
AOM or Defend-CLIP in their own reported evaluations.

\item \textbf{Appendix~\ref{sec:appendix-algorithm} (Algorithmic
Details of \method{})} gives the complete  calibration procedure
(Algorithm~\ref{alg:calibration}) and test-time inference procedure
(Algorithm~\ref{alg:react-clip}) for \method{}, along with full notation
and a per-image computational-cost analysis.
\end{itemize}

\section{Related Work}
\label{sec:related_work}
 
\paragraph{Vision--language models (VLMs).}
We use \emph{VLM} broadly for any model that couples visual inputs with
language representations or language outputs, including contrastive
image--text encoders such as CLIP, captioning and text-to-image generative
models, and multimodal large language models (MLLMs) that attach a visual
encoder or adapter to an LLM~\citep{zhang2024vision,yu2022coca,ramesh2021zero,saharia2022photorealistic,zhao2023evaluating,wang2025double,malik2025robust}.
Among these, CLIP remains the central testbed for zero-shot robustness
because its frozen visual encoder is reused, without further training, as
the backbone for classification, retrieval, detection, and a growing range
of downstream and multimodal
systems~\citep{radford2021learning,shin2022reco,zhou2022extract,zhang2023simple,zhao2022exploiting,cai2023semantic}.
This reuse is precisely what makes robustness delicate: altering the
backbone risks damaging the broad transfer that makes CLIP valuable in the
first place, while leaving it undefended exposes every downstream system
built on its visual representation to a single point of failure.
 
\paragraph{Adversarial robustness of zero-shot VLMs.}
Small $\ell_\infty$-bounded perturbations are well known to sharply change
neural-network
predictions~\citep{szegedy2013intriguing,papernot2016limitations,carlini2017towards,croce2020reliable},
and CLIP exhibits the same failure mode when attacked directly through its
zero-shot classification objective~\citep{mao2022understanding}. Most
existing defenses respond with some form of adversarial training.
\emph{Adversarial finetuning} generates adversarial examples during
training and updates the visual encoder itself, so that robustness
transfers to classes and datasets never seen during
finetuning~\citep{madry2017towards,mao2022understanding,schlarmann2024robust,wang2024pre,li2024language}.
\emph{Adversarial prompt tuning} instead keeps the visual and text
encoders fixed and optimizes learnable prompts or prompt tokens against a
similar adversarial objective~\citep{zhang2024adversarial,zhou2024few}.
Both families can meaningfully improve robustness, but both require
optimization beyond the original pretrained model, repeated
adversarial-example generation during training, and commonly trade away
clean accuracy in the process~\citep{zhang2019theoretically}.
 
\paragraph{Training-free test-time defenses.}
A separate line of work preserves the frozen VLM entirely and intervenes
only at inference. General-purpose test-time adaptation methods update
prompts or prediction statistics from augmented views of the test
input~\citep{shu2022test,feng2023diverse,abdul2023align,sheng2025r}, while
purification methods project the input or its representation toward a
cleaner manifold, sometimes with the help of a generative
model~\citep{nie2022diffusion,wang2022guided,fares2026mirrorcheck}. Within
the specific setting of test-time defense for zero-shot CLIP, this
literature is commonly organized into three strategies. \emph{Prompt-based}
methods re-optimize the textual prompt for each input, as in
R-TPT~\citep{sheng2025r} and the stability- and suitability-guided prompt
tuning of SS-TPT~\citep{kim2026ss}. \emph{Counterattack} methods instead
perturb the input itself in a direction expected to reverse the adversarial
displacement, beginning with the anti-adversarial perturbations of
Anti-Adv~\citep{alfarra2022combating} and continuing through TTC's
noise-triggered counterattack~\citep{xing2025clip}, DOC's diversified
search directions~\citep{jiang2026diversifying}, and MAC's counterattacks
guided by multiple augmented views~\citep{kim2026mac}. \emph{Feature-correction}
methods act directly on the visual feature: AOM averages the
representations of several Gaussian-noised views of the input into a
stable anchor and moves the original feature toward
it~\citep{tong2025zero}.
 
A further sub-thread within this literature studies how CLIP's visual
features respond to injected Gaussian noise, and uses that response as a
detection signal. Under weak noise, adversarial features can appear
deceptively stable, drifting no more than clean features -- a reversal
that \citet{xing2025clip} term \emph{false stability} and use to decide
when to trigger a counterattack. \citet{malik2026beyond} instead probe a
stronger noise scale, at which this reversal disappears and adversarial
features drift more than clean ones, and use that gap as a gating signal.
Defend-CLIP~\citep{brahma2026defending} combines both regimes directly: it
measures the change in feature drift between a weak and a strong noise
scale and thresholds this change to decide, in binary fashion, whether to
activate AOM's anchor-based correction.
 
This body of work establishes two things that \method{} builds on: that
training-free, noise-response signals can reliably separate clean from
adversarial CLIP features, and that noise-averaged anchors provide an
effective, training-free correction operator. Across all three strategies
above, however, the capacity of the intervention -- the prompt-update
step, the counterattack budget, or the anchor-noise scale and
extrapolation strength -- is fixed in advance and does not vary with the
input; where a gate is used, as in Defend-CLIP, it decides only
\emph{whether} a fixed correction is applied, not \emph{how strong} that
correction should be. \method{} retains the same noise-response principle
underlying this literature, but repurposes the relative change in drift
from a binary trigger into a graded, sample-specific signal that jointly
determines whether an input requires correction and how strong that
correction should be once triggered, allowing a single training-free
procedure to remain effective from very weak to very strong attacks
without any knowledge of the attack budget at test time.

\section{Datasets}
\label{app:datasets}

The datasets cover a wide range of domains: general object categories
(Caltech101, Caltech256, CIFAR10, CIFAR100, STL10), animal images
(OxfordPets), plant images (Flowers102), food images (Food101), vehicle
images (StanfordCars, FGVC-Aircraft), texture images (DTD), and remote
sensing images (EuroSAT)~\citep{fei2004learning,griffin2007caltech,krizhevsky2009learning,coates2011analysis,parkhi2012cats,nilsback2008automated,bossard2014food,krause20133d,maji2013fine,cimpoi2014describing,helber2019eurosat}.
To test generalization beyond the data used to calibrate the
drift-to-anchor mapping, we additionally evaluate on
ImageNet~\citep{deng2009imagenet} and four out-of-distribution variants:
ImageNet-A~\citep{hendrycks2021natural},
ImageNet-V2~\citep{recht2019imagenet},
ImageNet-R~\citep{hendrycks2021many}, and
ImageNet-Sketch~\citep{wang2019learning}. Because crafting adversarial
examples at multiple attack budgets across all five ImageNet-scale
datasets is computationally expensive, for ImageNet and each of its four
out-of-distribution variants we randomly select 5 samples per class for
evaluation, reducing the computational cost of adversarial example
generation while preserving class-balanced coverage of each dataset.
Table~\ref{tab:app_datasets} summarizes the statistics.

\begin{table}[t]
\centering
\renewcommand{\arraystretch}{1.3}
\setlength{\tabcolsep}{3pt}
\resizebox{\linewidth}{!}{
\begin{tabular}{llrr}
\toprule
\rowcolor{headerbg}
\color{white}\bfseries Dataset & \color{white}\bfseries Description & \color{white}\bfseries \# Classes & \color{white}\bfseries \# Test \\
\midrule
\multicolumn{4}{l}{\bfseries Fine-Grained Datasets} \\
\midrule
\rowcolor{rowgray}
CIFAR10       & Low-resolution object images  & 10    & 10,000 \\
CIFAR100      & Low-resolution object images  & 100   & 10,000 \\
\rowcolor{rowgray}
STL10         & Object images                 & 10    & 8,000  \\
DTD           & Texture images                & 47    & 1,880  \\
\rowcolor{rowgray}
OxfordPets    & Pet images (cats and dogs)     & 37    & 3,669  \\
Flowers102    & Flower species images          & 102   & 6,149  \\
\rowcolor{rowgray}
FGVC-Aircraft & Aircraft model images          & 100   & 3,333  \\
StanfordCars  & Car model images               & 196   & 8,041  \\
\rowcolor{rowgray}
Caltech101    & Object category images         & 101   & 8,677  \\
Caltech256    & Object category images         & 257   & 30,607 \\
\rowcolor{rowgray}
Food101       & Food images                    & 101   & 25,250 \\
EuroSAT       & Satellite images               & 10    & 27,000 \\
\cmidrule(lr){1-4}
\multicolumn{2}{l}{\itshape Total -- 12 datasets} & -- & \bfseries 142{,}606 \\
\midrule
\multicolumn{4}{l}{\bfseries ImageNet \& OOD Variants} \\
\midrule
\rowcolor{rowgray}
ImageNet      & Large-scale object images      & 1,000 & 5,000 \\
ImageNet-A    & Adversarial natural images      & 200   & 1,000 \\
\rowcolor{rowgray}
ImageNet-V2   & Re-collected ImageNet images    & 1,000 & 5,000 \\
ImageNet-R    & Artistic rendition images       & 200   & 1,000 \\
\rowcolor{rowgray}
ImageNet-S    & Sketch images                   & 1,000 & 5,000 \\
\cmidrule(lr){1-4}
\multicolumn{2}{l}{\itshape Total -- 5 datasets} & -- & \bfseries 17{,}000 \\
\bottomrule
\end{tabular}
}
\caption{Datasets used in our evaluation, grouped into 12 fine-grained downstream benchmarks and the ImageNet zero-shot generalization suite (ImageNet plus four out-of-distribution variants). We report the number of classes, the test-split size, and the total test samples per group.}
\label{tab:app_datasets}
\end{table}

\section{Detailed and Extended Results Corresponding to the Main Paper}
\label{app:extended_main_results}

This appendix reports per-dataset and extended-setting results underlying
the aggregate numbers presented in the main paper. These breakdowns serve
two purposes: first, to verify that the robustness gains of \method{} are
distributed across the evaluation suite rather than driven by a small
subset of datasets; and second, to examine the defense under conditions
not covered by the main aggregate tables, including perturbation budgets
beyond the calibration range, additional baselines, and a larger number
of attack iterations.

\begin{table*}[!htbp]
\centering
\renewcommand{\arraystretch}{1.20}
\setlength{\tabcolsep}{12pt}
\resizebox{\textwidth}{!}{
\begin{tabular}{ll| >{\color{gray}\columncolor{gray!10}}c|ccc|cccccc|>{\columncolor{oursbg}}c}
\toprule
\multirow{3}{*}{\bfseries Dataset} & \multirow{3}{*}{\bfseries Setting} & \multicolumn{1}{c|}{\itshape\textcolor{gray}{Zero-shot}} & \multicolumn{3}{c|}{\itshape\textcolor{grouplabel}{Adversarial Finetuning}} & \multicolumn{7}{c}{\itshape\textcolor{grouplabel}{Test-time Defense}} \\
\cmidrule(lr){3-3} \cmidrule(lr){4-6} \cmidrule(lr){7-12} \cmidrule(lr){13-13}
 & & \textcolor{gray}{\bfseries CLIP}  & \bfseries TeCoA & \bfseries PMG-AFT & \bfseries FARE & \bfseries TTC & \bfseries DOC & \bfseries SS-TPT & \bfseries AOM & \bfseries MAC & \bfseries Defend-CLIP & \bfseries \method{} \\
 & & {\scriptsize\itshape\textcolor{gray}{(ICML '21)}}  & {\scriptsize\itshape\textcolor{iclrcol}{(ICLR '23)}} & {\scriptsize\itshape\textcolor{cvpr25col}{(CVPR '24)}} & {\scriptsize\itshape\textcolor{icmlcol}{(ICML '24)}} & {\scriptsize\itshape\textcolor{cvpr25col}{(CVPR '25)}} & {\scriptsize\itshape\textcolor{aaaicol}{(AAAI '26)}} & {\scriptsize\itshape\textcolor{icmlcol}{(ICML '26)}} & {\scriptsize\itshape\textcolor{cvpr25col}{(CVPR '25)}} & {\scriptsize\itshape\textcolor{cvpr26col}{(CVPR '26)}} & {\scriptsize\itshape\textcolor{cvpr26col}{(CVPR '26)}} & \bfseries \textcolor{oursaccent}{~$\bigstar$} \\
\midrule
\multirow{5}{*}{CIFAR10}
 & Clean & 85.05  & 64.68 & 70.69 & 74.41 & 81.30 & 81.15 & 82.42 & 55.10 & \underline{84.09} & 83.61 & \cellcolor{bestcell}{84.49}~\textcolor{deltapos}{\scriptsize(+0.40)} \\
 & $\epsilon=1/255$ & 0.66  & 33.65 & 40.63 & 19.61 & 27.66 & 43.35 & \cellcolor{bestcell}\textbf{73.68} & \underline{72.66} & 34.98 & 49.05 & 71.93~\textcolor{deltaneg}{\scriptsize(-1.75)} \\
 & $\epsilon=4/255$ & 0.00  & 0.61 & 1.18 & 0.01 & 30.00 & 37.73 & 36.11 & \underline{74.26} & 36.02 & 70.13 & \cellcolor{bestcell}\textbf{77.67}~\textcolor{deltapos}{\scriptsize(+3.41)} \\
 & $\epsilon=8/255$ & 0.00  & 0.00 & 0.00 & 0.00 & 12.22 & 12.51 & 13.96 & \underline{58.17} & 13.14 & 54.15 & \cellcolor{bestcell}\textbf{76.33}~\textcolor{deltapos}{\scriptsize(+18.16)} \\
 & $\epsilon=16/255$ & 0.00  & 0.00 & 0.00 & 0.00 & 7.35 & 4.39 & 4.94 & \underline{31.69} & 6.02 & 27.67 & \cellcolor{bestcell}\textbf{78.05}~\textcolor{deltapos}{\scriptsize(+46.36)} \\
\midrule
\multirow{5}{*}{CIFAR100}
 & Clean & 57.22  & 35.98 & 40.31 & 46.64 & 56.37 & 56.43 & 52.38 & 27.68 & \cellcolor{bestcell}{\textbf{57.25}} & 56.13 & 56.06~\textcolor{deltaneg}{\scriptsize(-1.19)} \\
 & $\epsilon=1/255$ & 0.18  & 18.94 & 22.53 & 11.37 & 17.35 & 22.14 & \cellcolor{bestcell}\textbf{43.93} & 36.53 & 20.39 & 22.10 & \underline{43.55}~\textcolor{deltaneg}{\scriptsize(-0.38)} \\
 & $\epsilon=4/255$ & 0.00 & 1.59 & 1.73 & 0.06 & 15.05 & 15.20 & 18.71 & \underline{39.12} & 18.77 & 34.40 & \cellcolor{bestcell}\textbf{48.12}~\textcolor{deltapos}{\scriptsize(+9.00)} \\
 & $\epsilon=8/255$ & 0.00  & 0.06 & 0.04 & 0.00 & 9.66 & 2.58 & 7.79 & \underline{28.99} & 5.50 & 24.31 & \cellcolor{bestcell}\textbf{43.89}~\textcolor{deltapos}{\scriptsize(+14.90)} \\
 & $\epsilon=16/255$ & 0.00 & 0.00 & 0.00 & 0.00 & 8.39 & 0.58 & 2.75 & \underline{15.79} & 2.29 & 11.11 & \cellcolor{bestcell}\textbf{44.57}~\textcolor{deltapos}{\scriptsize(+28.78)} \\
\midrule
\multirow{5}{*}{STL10}
 & Clean & 96.40  & 87.38 & 88.59 & 91.75 & 95.96 & 95.90 & 95.72 & 90.09 & \cellcolor{bestcell}\textbf{96.28} & 95.07 & \underline{96.32}~\textcolor{deltapos}{\scriptsize(+0.04)} \\
 & $\epsilon=1/255$ & 11.07 & 70.06 & 73.01 & 59.01 & 69.28 & 82.48 & 91.25 & \underline{92.22} & 60.09 & 83.55 & \cellcolor{bestcell}\textbf{93.11}~\textcolor{deltapos}{\scriptsize(+0.89)} \\
 & $\epsilon=4/255$ & 0.03 & 13.04 & 15.10 & 1.89 & 52.00 & 69.03 & 76.08 & \underline{86.29} & 72.78 & 85.55 & \cellcolor{bestcell}\textbf{93.77}~\textcolor{deltapos}{\scriptsize(+7.48)} \\
 & $\epsilon=8/255$ & 0.00  & 0.43 & 0.34 & 0.03 & 17.15 & 22.71 & 57.17 & \underline{57.73} & 45.96 & 57.01 & \cellcolor{bestcell}\textbf{93.19}~\textcolor{deltapos}{\scriptsize(+35.46)} \\
 & $\epsilon=16/255$ & 0.00 & 0.03 & 0.00 & 0.00 & 7.61 & 7.96 & \underline{30.41} & 27.47 & 29.02 & 26.76 & \cellcolor{bestcell}\textbf{93.21}~\textcolor{deltapos}{\scriptsize(+62.80)} \\
\midrule
\multirow{5}{*}{DTD}
 & Clean & 40.64  & 25.00 & 21.81 & 31.97 & 35.85 & 35.05 & \cellcolor{bestcell}\textbf{42.71} & 26.65 & \underline{40.00} & 35.48 & 39.63~\textcolor{deltaneg}{\scriptsize(-3.08)} \\
 & $\epsilon=1/255$ & 2.93  & 17.50 & 15.05 & 15.69 & 23.88 & \underline{32.55} & \cellcolor{bestcell}\textbf{34.94} & 30.74 & 16.49 & 26.54 & 31.17~\textcolor{deltaneg}{\scriptsize(-3.77)} \\
 & $\epsilon=4/255$ & 0.11 & 4.20 & 4.26 & 0.90 & 13.40 & 17.55 & \underline{25.58} & 25.00 & 17.98 & 21.86 & \cellcolor{bestcell}\textbf{34.10}~\textcolor{deltapos}{\scriptsize(+8.52)} \\
 & $\epsilon=8/255$ & 0.00  & 0.37 & 0.59 & 0.21 & 5.16 & 3.29 & \underline{19.94} & 12.23 & 9.04 & 9.10 & \cellcolor{bestcell}\textbf{33.03}~\textcolor{deltapos}{\scriptsize(+13.09)} \\
 & $\epsilon=16/255$ & 0.00  & 0.21 & 0.27 & 0.11 & 3.94 & 0.85 & \underline{12.71} & 6.33 & 3.72 & 3.19 & \cellcolor{bestcell}\textbf{32.61}~\textcolor{deltapos}{\scriptsize(+19.90)} \\
\midrule
\multirow{5}{*}{OxfordPets}
 & Clean & 87.38  & 62.14 & 65.88 & 79.37 & 65.49 & 71.98 & 83.51 & 78.60 & \cellcolor{bestcell}\textbf{87.35} & 80.89 & \underline{85.36}~\textcolor{deltaneg}{\scriptsize(-1.99)} \\
 & $\epsilon=1/255$ & 1.01  & 38.38 & 41.32 & 31.15 & 46.03 & 64.15 & 73.53 & \underline{82.86} & 16.41 & 77.32 & \cellcolor{bestcell}\textbf{83.70}~\textcolor{deltapos}{\scriptsize(+0.84)} \\
 & $\epsilon=4/255$ & 0.00  & 0.90 & 1.77 & 0.22 & 25.78 & 44.67 & 46.36 & \underline{74.54} & 39.68 & 73.29 & \cellcolor{bestcell}\textbf{84.55}~\textcolor{deltapos}{\scriptsize(+10.01)} \\
 & $\epsilon=8/255$ & 0.00  & 0.00 & 0.00 & 0.14 & 4.17 & 3.76 & 27.09 & \underline{30.72} & 16.14 & 29.52 & \cellcolor{bestcell}\textbf{81.98}~\textcolor{deltapos}{\scriptsize(+51.26)} \\
 & $\epsilon=16/255$ & 0.00  & 0.00 & 0.00 & 0.14 & 2.15 & 0.41 & \underline{9.23} & 8.91 & 6.35 & 7.71 & \cellcolor{bestcell}\textbf{80.49}~\textcolor{deltapos}{\scriptsize(+71.26)} \\
\midrule
\multirow{5}{*}{Flowers102}
 & Clean & 65.43  & 36.82 & 37.05 & 47.99 & 63.08 & 62.49 & 60.96 & 50.67 & \cellcolor{bestcell}\textbf{64.95} & 61.72 & \underline{64.92}~\textcolor{deltaneg}{\scriptsize(-0.03)} \\
 & $\epsilon=1/255$ & 1.11  & 2.49 & 23.52 & 17.17 & 28.85 & 43.63 & 51.35 & \underline{55.50} & 11.12 & 49.37 & \cellcolor{bestcell}\textbf{58.97}~\textcolor{deltapos}{\scriptsize(+3.47)} \\
 & $\epsilon=4/255$ & 0.00  & 1.90 & 2.59 & 0.03 & 15.44 & 27.40 & 32.39 & \underline{44.46} & 19.89 & 40.66 & \cellcolor{bestcell}\textbf{61.74}~\textcolor{deltapos}{\scriptsize(+17.27)} \\
 & $\epsilon=8/255$ & 0.00  & 0.00 & 0.00 & 0.03 & 3.90 & 2.11 & \underline{19.51} & 14.80 & 8.38 & 10.99 & \cellcolor{bestcell}\textbf{60.16}~\textcolor{deltapos}{\scriptsize(+40.65)} \\
 & $\epsilon=16/255$ & 0.00  & 0.00 & 0.00 & 0.03 & 2.91 & 0.19 & \underline{8.53} & 6.31 & 3.37 & 2.50 & \cellcolor{bestcell}\textbf{59.94}~\textcolor{deltapos}{\scriptsize(+51.41)} \\
\midrule
\multirow{5}{*}{FGVCAircraft}
 & Clean & 20.10 & 5.43 & 5.40 & 10.83 & 15.78 & 15.30 & 17.94 & 12.60 & \cellcolor{bestcell}\textbf{19.71} & 15.57 & \underline{19.26}~\textcolor{deltaneg}{\scriptsize(-0.45)} \\
 & $\epsilon=1/255$ & 0.00 & 2.49 & 2.16 & 1.35 & 8.19 & \cellcolor{bestcell}\textbf{15.90} & 13.29 & \underline{14.55} & 2.01 & 10.80 & 13.95~\textcolor{deltaneg}{\scriptsize(-1.95)} \\
 & $\epsilon=4/255$ & 0.00  & 0.03 & 0.03 & 0.00 & 7.20 & 9.57 & 7.44 & \underline{11.43} & 9.84 & 9.24 & \cellcolor{bestcell}\textbf{14.94}~\textcolor{deltapos}{\scriptsize(+3.51)} \\
 & $\epsilon=8/255$ & 0.00 & 0.00 & 0.00 & 0.00 & 3.21 & 0.87 & 4.38 & \underline{5.16} & 3.09 & 2.97 & \cellcolor{bestcell}\textbf{15.24}~\textcolor{deltapos}{\scriptsize(+10.09)} \\
 & $\epsilon=16/255$ & 0.00  & 0.00 & 0.00 & 0.00 & 2.79 & 0.00 & 2.22 & \underline{4.20} & 1.14 & 2.01 & \cellcolor{bestcell}\textbf{15.33}~\textcolor{deltapos}{\scriptsize(+11.13)} \\
\midrule
\multirow{5}{*}{StanfordCars}
 & Clean & 52.10 & 20.98 & 25.51 & 38.61 & 41.20 & 41.82 & \cellcolor{bestcell}\textbf{56.00} & 43.45 & \underline{51.65} & 47.27 & 49.84~\textcolor{deltaneg}{\scriptsize(-6.16)} \\
 & $\epsilon=1/255$ & 0.02  & 8.76 & 11.73 & 6.79 & 14.48 & 28.64 & 43.50 & \cellcolor{bestcell}\textbf{45.38} & 5.67 & 40.59 & \underline{45.40}~\textcolor{deltapos}{\scriptsize(+0.12)} \\
 & $\epsilon=4/255$ & 0.00  & 0.15 & 0.12 & 0.01 & 13.65 & 24.01 & 20.96 & \underline{28.34} & 26.30 & 24.98 & \cellcolor{bestcell}\textbf{45.53}~\textcolor{deltapos}{\scriptsize(+17.19)} \\
 & $\epsilon=8/255$ & 0.00  & 0.00 & 0.00 & 0.01 & 4.02 & 0.78 & \underline{10.08} & 8.27 & 8.01 & 4.91 & \cellcolor{bestcell}\textbf{44.70}~\textcolor{deltapos}{\scriptsize(+34.62)} \\
 & $\epsilon=16/255$ & 0.00  & 0.00 & 0.00 & 0.01 & 3.36 & 0.04 & 3.09 & \underline{6.85} & 2.92 & 3.49 & \cellcolor{bestcell}\textbf{44.78}~\textcolor{deltapos}{\scriptsize(+37.93)} \\
\midrule
\multirow{5}{*}{Caltech101}
 & Clean & 85.69  & 71.79 & 75.45 & 80.93 & \underline{86.06} & \cellcolor{bestcell}\textbf{86.19} & 86.02 & 81.85 & 85.81 & 85.27 & 85.42~\textcolor{deltaneg}{\scriptsize(-0.77)} \\
 & $\epsilon=1/255$ & 14.68  & 55.53 & 61.06 & 50.73 & 51.76 & 65.84 & 80.68 & \cellcolor{bestcell}\textbf{83.19} & 48.92 & 72.13 & \underline{81.33}~\textcolor{deltaneg}{\scriptsize(-1.86)} \\
 & $\epsilon=4/255$ & 0.55  & 15.73 & 19.45 & 5.14 & 37.66 & 53.00 & 68.11 & \underline{74.62} & 57.36 & 71.56 & \cellcolor{bestcell}\textbf{82.14}~\textcolor{deltapos}{\scriptsize(+7.52)} \\
 & $\epsilon=8/255$ & 0.29  & 2.98 & 3.30 & 0.99 & 13.61 & 17.27 & \underline{55.95} & 41.71 & 38.60 & 39.03 & \cellcolor{bestcell}\textbf{80.07}~\textcolor{deltapos}{\scriptsize(+23.12)} \\
 & $\epsilon=16/255$ & 0.24  & 0.90 & 1.06 & 0.67 & 6.63 & 5.86 & \underline{40.40} & 20.48 & 26.40 & 17.81 & \cellcolor{bestcell}\textbf{80.35}~\textcolor{deltapos}{\scriptsize(+39.95)} \\
\midrule
\multirow{5}{*}{Caltech256}
 & Clean & 81.72  & 61.15 & 62.27 & 73.30 & 76.52 & 76.53 & \cellcolor{bestcell}\textbf{82.75} & 73.04 & \underline{81.49} & 79.60 & 80.51~\textcolor{deltaneg}{\scriptsize(-2.24)} \\
 & $\epsilon=1/255$ & 8.43  & 43.17 & 45.94 & 38.74 & 43.74 & 60.59 & \underline{75.20} & \underline{75.20} & 41.36 & 65.51 & \cellcolor{bestcell}\textbf{76.14}~\textcolor{deltapos}{\scriptsize(+0.94)} \\
 & $\epsilon=4/255$ & 0.12  & 8.30 & 10.61 & 2.14 & 27.78 & 44.35 & 61.00 & \underline{65.12} & 52.27 & 62.35 & \cellcolor{bestcell}\textbf{77.10}~\textcolor{deltapos}{\scriptsize(+11.98)} \\
 & $\epsilon=8/255$ & 0.05  & 0.96 & 1.30 & 0.12 & 15.11 & 26.82 & 48.18 & \underline{54.27} & 44.02 & 51.76 & \cellcolor{bestcell}\textbf{76.57}~\textcolor{deltapos}{\scriptsize(+22.30)} \\
 & $\epsilon=16/255$ & 0.02  & 0.24 & 0.25 & 0.05 & 3.30 & 2.08 & \underline{31.10} & 16.51 & 16.86 & 14.01 & \cellcolor{bestcell}\textbf{75.65}~\textcolor{deltapos}{\scriptsize(+44.55)} \\
\midrule
\multirow{5}{*}{Food101}
 & Clean & 83.88 & 30.00 & 36.61 & 55.30 & 80.15 & 78.49 & \underline{81.57} & 61.45 & \cellcolor{bestcell}\textbf{83.23} & 74.50 & 81.72~\textcolor{deltaneg}{\scriptsize(-1.51)} \\
 & $\epsilon=1/255$ & 0.71  & 13.88 & 18.57 & 11.66 & 35.27 & 50.24 & \underline{67.87} & 66.14 & 16.96 & 62.88 & \cellcolor{bestcell}\textbf{76.27}~\textcolor{deltapos}{\scriptsize(+8.40)} \\
 & $\epsilon=4/255$ & 0.00  & 0.56 & 1.03 & 0.06 & 19.20 & 32.55 & 40.72 & \underline{49.86} & 40.34 & 48.30 & \cellcolor{bestcell}\textbf{76.09}~\textcolor{deltapos}{\scriptsize(+26.23)} \\
 & $\epsilon=8/255$ & 0.00  & 0.00 & 0.01 & 0.00 & 8.41 & 14.34 & 21.22 & \underline{35.46} & 30.41 & 33.91 & \cellcolor{bestcell}\textbf{75.38}~\textcolor{deltapos}{\scriptsize(+39.92)} \\
 & $\epsilon=16/255$ & 0.00  & 0.00 & 0.00 & 0.00 & 2.26 & 0.31 & \underline{7.04} & 4.64 & 5.84 & 3.09 & \cellcolor{bestcell}\textbf{72.40}~\textcolor{deltapos}{\scriptsize(+65.36)} \\
\midrule
\multirow{5}{*}{EuroSAT}
 & Clean & 42.59  & 16.49 & 18.40 & 21.79 & \cellcolor{bestcell}\textbf{53.13} & \underline{53.11} & 17.31 & 6.06 & 41.85 & 37.30 & 41.04~\textcolor{deltaneg}{\scriptsize(-12.09)} \\
 & $\epsilon=1/255$ & 0.04  & 11.95 & 12.55 & 10.69 & \underline{21.72} & 14.30 & 13.03 & 15.39 & 10.27 & 9.23 & \cellcolor{bestcell}\textbf{25.95}~\textcolor{deltapos}{\scriptsize(+4.23)} \\
 & $\epsilon=4/255$ & 0.00  & 9.79 & 9.63 & 0.00 & \underline{22.55} & 14.20 & 6.48 & 22.36 & 8.31 & 19.90 & \cellcolor{bestcell}\textbf{36.75}~\textcolor{deltapos}{\scriptsize(+14.22)} \\
 & $\epsilon=8/255$ & 0.00 & 0.01 & 0.00 & 0.00 & 20.21 & 10.14 & 5.38 & \underline{22.50} & 5.51 & 20.07 & \cellcolor{bestcell}\textbf{38.49}~\textcolor{deltapos}{\scriptsize(+15.99)} \\
 & $\epsilon=16/255$ & 0.00  & 0.00 & 0.00 & 0.00 & \underline{16.61} & 2.32 & 4.81 & 10.79 & 1.05 & 8.37 & \cellcolor{bestcell}\textbf{40.54}~\textcolor{deltapos}{\scriptsize(+23.93)} \\
\midrule
\multirow{5}{*}{\bfseries Average}
 & Clean & 66.52  & 43.15 & 45.66 & 54.41 & 62.57 & 62.87 & 63.27 & 50.60 & \cellcolor{bestcell}\textbf{66.14} & 62.70 & \underline{65.38}~\textcolor{deltaneg}{\scriptsize(-0.76)} \\
 & $\epsilon=1/255$ & 3.40  & 26.40 & 30.67 & 22.83 & 32.35 & 43.65 & 55.19 & \underline{55.86} & 23.72 & 47.42 & \cellcolor{bestcell}\textbf{58.47}~\textcolor{deltapos}{\scriptsize(+2.61)} \\
 & $\epsilon=4/255$ & 0.07  & 4.73 & 5.63 & 0.87 & 23.31 & 32.44 & 36.66 & \underline{49.62} & 33.29 & 46.85 & \cellcolor{bestcell}\textbf{61.04}~\textcolor{deltapos}{\scriptsize(+11.42)} \\
 & $\epsilon=8/255$ & 0.03  & 0.40 & 0.47 & 0.13 & 9.74 & 9.77 & 24.22 & \underline{30.83} & 18.98 & 28.14 & \cellcolor{bestcell}\textbf{59.92}~\textcolor{deltapos}{\scriptsize(+29.09)} \\
 & $\epsilon=16/255$ & 0.02  & 0.12 & 0.13 & 0.08 & 5.52 & 2.08 & 13.10 & \underline{13.33} & 8.75 & 10.64 & \cellcolor{bestcell}\textbf{59.83}~\textcolor{deltapos}{\scriptsize(+46.50)} \\
\bottomrule
\end{tabular}
}
\caption{\textbf{Per-dataset breakdown of Table~1~\emph{(main paper)}.} Clean and PGD-10
robust accuracy (\%) at $\epsilon\in\{1,4,8,16\}/255$ for each of the 12
downstream datasets, alongside the 12-dataset average reported in Table~1
of the main paper. Values in parentheses denote the difference between
\method{} and the strongest baseline in each row. This breakdown verifies
that the robustness gains of \method{} under PGD-10 attacks are
distributed across datasets and visual domains rather than concentrated
in a small subset of the evaluation suite.}
\label{tab:main_table1_per_dataset_results}
\end{table*}

\begin{table*}[!htbp]
\centering
\renewcommand{\arraystretch}{1.05}
\setlength{\tabcolsep}{6pt}
\begin{tabular}{l|l|l|ccccc}
\toprule
\bfseries Dataset & \bfseries Anchor construction & \bfseries Intervention Metric $s_{\mathrm{int}}(x)$
& \bfseries Clean & \bfseries $\epsilon{=}1$ & \bfseries $\epsilon{=}4$ & \bfseries $\epsilon{=}8$ & \bfseries $\epsilon{=}16$ \\
\midrule

Caltech101 & \cellcolor{rowgray}Fixed anchor ($\sigma_{\mathrm a}{=}0.1$) & \cellcolor{rowgray}$r(x)$ & \cellcolor{oursbg!35}85.27 & \cellcolor{oursbg!35}72.13 & \cellcolor{oursbg!35}71.56 & \cellcolor{oursbg!35}39.03 & \cellcolor{oursbg!35}17.81 \\
 & \cellcolor{oursbg!35}$\sigma_{\mathrm{corr}}(x)$ & \cellcolor{oursbg!35}$r(x)$ & \cellcolor{rowgray}85.34 & \cellcolor{rowgray}75.07 & \cellcolor{rowgray}81.95 & \cellcolor{rowgray}80.18 & \cellcolor{rowgray}80.44 \\
 & \cellcolor{oursbg!35}$\sigma_{\mathrm{corr}}(x)$ & \cellcolor{oursbg!35}$r(x){+}J(x)$\textcolor{oursaccent}{~$\bigstar$} & \cellcolor{oursbg!35}85.42 & \cellcolor{oursbg!35}81.33 & \cellcolor{oursbg!35}82.14 & \cellcolor{oursbg!35}80.07 & \cellcolor{oursbg!35}80.35 \\
\midrule

Caltech256 & \cellcolor{rowgray}Fixed anchor ($\sigma_{\mathrm a}{=}0.1$) & \cellcolor{rowgray}$r(x)$ & \cellcolor{oursbg!35}79.60 & \cellcolor{oursbg!35}65.51 & \cellcolor{oursbg!35}62.35 & \cellcolor{oursbg!35}51.76 & \cellcolor{oursbg!35}14.01 \\
 & \cellcolor{oursbg!35}$\sigma_{\mathrm{corr}}(x)$ & \cellcolor{oursbg!35}$r(x)$ & \cellcolor{rowgray}79.54 & \cellcolor{rowgray}70.57 & \cellcolor{rowgray}77.25 & \cellcolor{rowgray}76.94 & \cellcolor{rowgray}76.00 \\
 & \cellcolor{oursbg!35}$\sigma_{\mathrm{corr}}(x)$ & \cellcolor{oursbg!35}$r(x){+}J(x)$\textcolor{oursaccent}{~$\bigstar$} & \cellcolor{oursbg!35}80.51 & \cellcolor{oursbg!35}76.14 & \cellcolor{oursbg!35}77.10 & \cellcolor{oursbg!35}76.57 & \cellcolor{oursbg!35}75.65 \\
\midrule

CIFAR10 & \cellcolor{rowgray}Fixed anchor ($\sigma_{\mathrm a}{=}0.1$) & \cellcolor{rowgray}$r(x)$ & \cellcolor{oursbg!35}83.61 & \cellcolor{oursbg!35}49.05 & \cellcolor{oursbg!35}70.13 & \cellcolor{oursbg!35}54.15 & \cellcolor{oursbg!35}27.67 \\
 & \cellcolor{oursbg!35}$\sigma_{\mathrm{corr}}(x)$ & \cellcolor{oursbg!35}$r(x)$ & \cellcolor{rowgray}83.71 & \cellcolor{rowgray}55.17 & \cellcolor{rowgray}78.28 & \cellcolor{rowgray}76.28 & \cellcolor{rowgray}78.00 \\
 & \cellcolor{oursbg!35}$\sigma_{\mathrm{corr}}(x)$ & \cellcolor{oursbg!35}$r(x){+}J(x)$\textcolor{oursaccent}{~$\bigstar$} & \cellcolor{oursbg!35}84.49 & \cellcolor{oursbg!35}71.93 & \cellcolor{oursbg!35}77.67 & \cellcolor{oursbg!35}76.33 & \cellcolor{oursbg!35}78.05 \\
\midrule

CIFAR100 & \cellcolor{rowgray}Fixed anchor ($\sigma_{\mathrm a}{=}0.1$) & \cellcolor{rowgray}$r(x)$ & \cellcolor{oursbg!35}56.13 & \cellcolor{oursbg!35}22.10 & \cellcolor{oursbg!35}34.40 & \cellcolor{oursbg!35}24.31 & \cellcolor{oursbg!35}11.11 \\
 & \cellcolor{oursbg!35}$\sigma_{\mathrm{corr}}(x)$ & \cellcolor{oursbg!35}$r(x)$ & \cellcolor{rowgray}56.23 & \cellcolor{rowgray}30.46 & \cellcolor{rowgray}48.14 & \cellcolor{rowgray}43.78 & \cellcolor{rowgray}44.46 \\
 & \cellcolor{oursbg!35}$\sigma_{\mathrm{corr}}(x)$ & \cellcolor{oursbg!35}$r(x){+}J(x)$\textcolor{oursaccent}{~$\bigstar$} & \cellcolor{oursbg!35}56.06 & \cellcolor{oursbg!35}43.55 & \cellcolor{oursbg!35}48.12 & \cellcolor{oursbg!35}43.89 & \cellcolor{oursbg!35}44.57 \\
\midrule

DTD & \cellcolor{rowgray}Fixed anchor ($\sigma_{\mathrm a}{=}0.1$) & \cellcolor{rowgray}$r(x)$ & \cellcolor{oursbg!35}35.48 & \cellcolor{oursbg!35}26.54 & \cellcolor{oursbg!35}21.86 & \cellcolor{oursbg!35}9.10 & \cellcolor{oursbg!35}3.19 \\
 & \cellcolor{oursbg!35}$\sigma_{\mathrm{corr}}(x)$ & \cellcolor{oursbg!35}$r(x)$ & \cellcolor{rowgray}35.32 & \cellcolor{rowgray}32.82 & \cellcolor{rowgray}34.73 & \cellcolor{rowgray}33.67 & \cellcolor{rowgray}33.24 \\
 & \cellcolor{oursbg!35}$\sigma_{\mathrm{corr}}(x)$ & \cellcolor{oursbg!35}$r(x){+}J(x)$\textcolor{oursaccent}{~$\bigstar$} & \cellcolor{oursbg!35}39.63 & \cellcolor{oursbg!35}31.17 & \cellcolor{oursbg!35}34.10 & \cellcolor{oursbg!35}33.03 & \cellcolor{oursbg!35}32.61 \\
\midrule

EuroSAT & \cellcolor{rowgray}Fixed anchor ($\sigma_{\mathrm a}{=}0.1$) & \cellcolor{rowgray}$r(x)$ & \cellcolor{oursbg!35}37.30 & \cellcolor{oursbg!35}9.23 & \cellcolor{oursbg!35}19.90 & \cellcolor{oursbg!35}20.07 & \cellcolor{oursbg!35}8.37 \\
 & \cellcolor{oursbg!35}$\sigma_{\mathrm{corr}}(x)$ & \cellcolor{oursbg!35}$r(x)$ & \cellcolor{rowgray}37.13 & \cellcolor{rowgray}18.22 & \cellcolor{rowgray}37.14 & \cellcolor{rowgray}38.74 & \cellcolor{rowgray}40.79 \\
 & \cellcolor{oursbg!35}$\sigma_{\mathrm{corr}}(x)$ & \cellcolor{oursbg!35}$r(x){+}J(x)$\textcolor{oursaccent}{~$\bigstar$} & \cellcolor{oursbg!35}41.04 & \cellcolor{oursbg!35}25.95 & \cellcolor{oursbg!35}36.75 & \cellcolor{oursbg!35}38.49 & \cellcolor{oursbg!35}40.54 \\
\midrule

FGVC-Aircraft & \cellcolor{rowgray}Fixed anchor ($\sigma_{\mathrm a}{=}0.1$) & \cellcolor{rowgray}$r(x)$ & \cellcolor{oursbg!35}15.57 & \cellcolor{oursbg!35}10.80 & \cellcolor{oursbg!35}9.24 & \cellcolor{oursbg!35}2.97 & \cellcolor{oursbg!35}2.01 \\
 & \cellcolor{oursbg!35}$\sigma_{\mathrm{corr}}(x)$ & \cellcolor{oursbg!35}$r(x)$ & \cellcolor{rowgray}14.52 & \cellcolor{rowgray}13.89 & \cellcolor{rowgray}16.20 & \cellcolor{rowgray}16.50 & \cellcolor{rowgray}16.59 \\
 & \cellcolor{oursbg!35}$\sigma_{\mathrm{corr}}(x)$ & \cellcolor{oursbg!35}$r(x){+}J(x)$\textcolor{oursaccent}{~$\bigstar$} & \cellcolor{oursbg!35}19.26 & \cellcolor{oursbg!35}13.95 & \cellcolor{oursbg!35}14.94 & \cellcolor{oursbg!35}15.24 & \cellcolor{oursbg!35}15.33 \\
\midrule

Flowers102 & \cellcolor{rowgray}Fixed anchor ($\sigma_{\mathrm a}{=}0.1$) & \cellcolor{rowgray}$r(x)$ & \cellcolor{oursbg!35}61.72 & \cellcolor{oursbg!35}49.37 & \cellcolor{oursbg!35}40.66 & \cellcolor{oursbg!35}10.99 & \cellcolor{oursbg!35}2.50 \\
 & \cellcolor{oursbg!35}$\sigma_{\mathrm{corr}}(x)$ & \cellcolor{oursbg!35}$r(x)$ & \cellcolor{rowgray}61.05 & \cellcolor{rowgray}58.17 & \cellcolor{rowgray}62.53 & \cellcolor{rowgray}60.95 & \cellcolor{rowgray}60.74 \\
 & \cellcolor{oursbg!35}$\sigma_{\mathrm{corr}}(x)$ & \cellcolor{oursbg!35}$r(x){+}J(x)$\textcolor{oursaccent}{~$\bigstar$} & \cellcolor{oursbg!35}64.92 & \cellcolor{oursbg!35}58.97 & \cellcolor{oursbg!35}61.73 & \cellcolor{oursbg!35}60.16 & \cellcolor{oursbg!35}59.94 \\
\midrule

Food101 & \cellcolor{rowgray}Fixed anchor ($\sigma_{\mathrm a}{=}0.1$) & \cellcolor{rowgray}$r(x)$ & \cellcolor{oursbg!35}74.50 & \cellcolor{oursbg!35}62.88 & \cellcolor{oursbg!35}48.30 & \cellcolor{oursbg!35}33.91 & \cellcolor{oursbg!35}3.09 \\
 & \cellcolor{oursbg!35}$\sigma_{\mathrm{corr}}(x)$ & \cellcolor{oursbg!35}$r(x)$ & \cellcolor{rowgray}76.62 & \cellcolor{rowgray}75.62 & \cellcolor{rowgray}76.90 & \cellcolor{rowgray}76.17 & \cellcolor{rowgray}73.20 \\
 & \cellcolor{oursbg!35}$\sigma_{\mathrm{corr}}(x)$ & \cellcolor{oursbg!35}$r(x){+}J(x)$\textcolor{oursaccent}{~$\bigstar$} & \cellcolor{oursbg!35}81.72 & \cellcolor{oursbg!35}76.27 & \cellcolor{oursbg!35}76.09 & \cellcolor{oursbg!35}75.38 & \cellcolor{oursbg!35}72.40 \\
\midrule

OxfordPets & \cellcolor{rowgray}Fixed anchor ($\sigma_{\mathrm a}{=}0.1$) & \cellcolor{rowgray}$r(x)$ & \cellcolor{oursbg!35}80.89 & \cellcolor{oursbg!35}77.32 & \cellcolor{oursbg!35}73.29 & \cellcolor{oursbg!35}29.52 & \cellcolor{oursbg!35}7.71 \\
 & \cellcolor{oursbg!35}$\sigma_{\mathrm{corr}}(x)$ & \cellcolor{oursbg!35}$r(x)$ & \cellcolor{rowgray}80.38 & \cellcolor{rowgray}82.17 & \cellcolor{rowgray}86.10 & \cellcolor{rowgray}83.59 & \cellcolor{rowgray}82.09 \\
 & \cellcolor{oursbg!35}$\sigma_{\mathrm{corr}}(x)$ & \cellcolor{oursbg!35}$r(x){+}J(x)$\textcolor{oursaccent}{~$\bigstar$} & \cellcolor{oursbg!35}85.36 & \cellcolor{oursbg!35}83.70 & \cellcolor{oursbg!35}84.55 & \cellcolor{oursbg!35}81.98 & \cellcolor{oursbg!35}80.49 \\
\midrule

StanfordCars & \cellcolor{rowgray}Fixed anchor ($\sigma_{\mathrm a}{=}0.1$) & \cellcolor{rowgray}$r(x)$ & \cellcolor{oursbg!35}47.27 & \cellcolor{oursbg!35}40.59 & \cellcolor{oursbg!35}24.98 & \cellcolor{oursbg!35}4.91 & \cellcolor{oursbg!35}3.49 \\
 & \cellcolor{oursbg!35}$\sigma_{\mathrm{corr}}(x)$ & \cellcolor{oursbg!35}$r(x)$ & \cellcolor{rowgray}45.18 & \cellcolor{rowgray}47.32 & \cellcolor{rowgray}47.99 & \cellcolor{rowgray}47.16 & \cellcolor{rowgray}47.25 \\
 & \cellcolor{oursbg!35}$\sigma_{\mathrm{corr}}(x)$ & \cellcolor{oursbg!35}$r(x){+}J(x)$\textcolor{oursaccent}{~$\bigstar$} & \cellcolor{oursbg!35}49.84 & \cellcolor{oursbg!35}45.50 & \cellcolor{oursbg!35}45.53 & \cellcolor{oursbg!35}44.70 & \cellcolor{oursbg!35}44.78 \\
\midrule

STL10 & \cellcolor{rowgray}Fixed anchor ($\sigma_{\mathrm a}{=}0.1$) & \cellcolor{rowgray}$r(x)$ & \cellcolor{oursbg!35}95.07 & \cellcolor{oursbg!35}83.55 & \cellcolor{oursbg!35}85.55 & \cellcolor{oursbg!35}57.01 & \cellcolor{oursbg!35}26.76 \\
 & \cellcolor{oursbg!35}$\sigma_{\mathrm{corr}}(x)$ & \cellcolor{oursbg!35}$r(x)$ & \cellcolor{rowgray}95.28 & \cellcolor{rowgray}86.33 & \cellcolor{rowgray}93.85 & \cellcolor{rowgray}93.19 & \cellcolor{rowgray}93.21 \\
 & \cellcolor{oursbg!35}$\sigma_{\mathrm{corr}}(x)$ & \cellcolor{oursbg!35}$r(x){+}J(x)$\textcolor{oursaccent}{~$\bigstar$} & \cellcolor{oursbg!35}96.32 & \cellcolor{oursbg!35}93.11 & \cellcolor{oursbg!35}93.77 & \cellcolor{oursbg!35}93.19 & \cellcolor{oursbg!35}93.21 \\
\toprule

\textbf{Average} & \cellcolor{rowgray}Fixed anchor ($\sigma_{\mathrm a}{=}0.1$) & \cellcolor{rowgray}$r(x)$
& \cellcolor{rowgray}\underline{62.70} & \cellcolor{rowgray}47.42 & \cellcolor{rowgray}46.85 & \cellcolor{rowgray}28.14 & \cellcolor{rowgray}10.64 \\
 & \cellcolor{oursbg!35}$\sigma_{\mathrm{corr}}(x)$ & \cellcolor{oursbg!35}$r(x)$
& \cellcolor{oursbg!35}62.52 & \cellcolor{oursbg!35}\underline{53.82} & \cellcolor{bestcell}\textbf{61.76} & \cellcolor{bestcell}\textbf{60.60} & \cellcolor{bestcell}\textbf{60.50} \\
 & \cellcolor{oursbg!35}$\sigma_{\mathrm{corr}}(x)$ & \cellcolor{oursbg!35}$r(x){+}J(x)$\textcolor{oursaccent}{~$\bigstar$}
& \cellcolor{bestcell}\textbf{65.38} & \cellcolor{bestcell}\textbf{58.47} & \cellcolor{oursbg!35}\underline{61.04} & \cellcolor{oursbg!35}\underline{59.92} & \cellcolor{oursbg!35}\underline{59.83} \\
\midrule
\multicolumn{3}{l|}{\itshape\textcolor{gray}{$\Delta$ Response-conditioned anchor construction ($\sigma_{corr}(x)$ and $s_{int}(x)=r(x)$)}}
& \textcolor{deltaneg}{$-0.18$} & \textcolor{deltapos}{$+6.40$} & \textcolor{deltapos}{$+14.91$} & \textcolor{deltapos}{$+32.46$} & \textcolor{deltapos}{$+49.86$} \\
\multicolumn{3}{l|}{\itshape\textcolor{gray}{$+\Delta$ Defensive intervention score ($\sigma_{corr}(x)$ and $s_{int}(x)=r(x) + J(x)$)}}
& \textcolor{deltapos}{$+2.86$} & \textcolor{deltapos}{$+4.65$} & \textcolor{deltaneg}{$-0.72$} & \textcolor{deltaneg}{$-0.68$} & \textcolor{deltaneg}{$-0.67$} \\
\bottomrule
\end{tabular}
\caption{\textbf{Per-dataset breakdown of Table~2~\emph{(main paper)}.} Defended classification
accuracy (\%) under  PGD-10 attacks at $\epsilon\in\{1,4,8,16\}/255$,
alongside clean accuracy, under the three configurations of Table~2 — a
fixed anchor ($\sigma_{\mathrm{a}}=0.1$) with the drift-only score $r(x)$;
the response-conditioned anchor $\sigma_{\mathrm{corr}}(x)$ with $r(x)$;
and $\sigma_{\mathrm{corr}}(x)$ with the adopted fused score $r(x)+J(x)$ —
evaluated separately on each of the 12 downstream datasets. The Average
row and the two $\Delta$ rows beneath it reproduce Table~2 exactly,
isolating the gain from response-conditioned anchor construction from the
gain attributable to the fused defensive intervention score at the
per-dataset level.}
\label{tab:adaptive_correction_per_dataset}
\end{table*}

\begin{table*}[!htbp]
\centering
\renewcommand{\arraystretch}{1.15}
\setlength{\tabcolsep}{5pt}
\resizebox{\textwidth}{!}{
\begin{tabular}{l|
>{\columncolor{white}}c>{\columncolor{oursbg!35}}c|
>{\columncolor{white}}c>{\columncolor{oursbg!35}}c|
>{\columncolor{white}}c>{\columncolor{oursbg!35}}c|
>{\columncolor{white}}c>{\columncolor{oursbg!35}}c|
>{\columncolor{white}}c>{\columncolor{oursbg!35}}c}
\toprule
\multirow{2}{*}{\bfseries Dataset}
& \multicolumn{2}{c|}{\bfseries Clean}
& \multicolumn{2}{c|}{\bfseries $\epsilon{=}1$}
& \multicolumn{2}{c|}{\bfseries $\epsilon{=}4$}
& \multicolumn{2}{c|}{\bfseries $\epsilon{=}8$}
& \multicolumn{2}{c}{\bfseries $\epsilon{=}16$} \\
& \cellcolor{white}$r(x)$ & \cellcolor{oursbg!35}$r(x){+}J(x)$\textcolor{oursaccent}{~$\bigstar$}
& \cellcolor{white}$r(x)$ & \cellcolor{oursbg!35}$r(x){+}J(x)$\textcolor{oursaccent}{~$\bigstar$}
& \cellcolor{white}$r(x)$ & \cellcolor{oursbg!35}$r(x){+}J(x)$\textcolor{oursaccent}{~$\bigstar$}
& \cellcolor{white}$r(x)$ & \cellcolor{oursbg!35}$r(x){+}J(x)$\textcolor{oursaccent}{~$\bigstar$}
& \cellcolor{white}$r(x)$ & \cellcolor{oursbg!35}$r(x){+}J(x)$\textcolor{oursaccent}{~$\bigstar$} \\
\midrule
Caltech101      & 94.89 & 98.95 & 85.76 & 88.74 & 99.42 & 99.77 & 100    & 99.93 & 100 & 100  \\
Caltech256      & 83.04 & 96.07 & 87.97 & 93.62 & 99.58 & 99.84 & 99.96  & 99.94 & 100 & 100  \\
CIFAR10         & 97.24 & 98.95 & 70.34 & 90.50 & 99.85 & 98.95 & 100    & 99.96 & 100  &  100 \\
CIFAR100        & 97.43 & 96.92 & 66.43 & 93.83 & 99.90 & 99.56 & 100    & 100   & 100 & 100  \\
STL10           & 87.64 & 99.49 & 90.00 & 93.75 & 99.97 & 99.84 & 100    & 100   & 100 & 100  \\
OxfordPets      & 32.50 & 87.37 & 94.45 & 97.94 & 99.94 & 100   & 100    & 100   & 100 & 100  \\
Flowers102      & 78.50 & 98.14 & 95.00 & 97.22 & 100   & 100   & 100    & 100   & 100 & 100  \\
Food101         & 66.02 & 95.41 & 97.30 & 99.09 & 99.99 & 99.94 & 100    & 99.99 & 100  & 100  \\
StanfordCars    & 46.60 & 89.85 & 96.37 & 97.97 & 100   & 100   & 100    & 100   & 100 & 100  \\
FGVC-Aircraft   & 47.91 & 92.54 & 84.48 & 94.18 & 100   & 100   & 100    & 100   & 100 & 100  \\
DTD             & 72.91 & 95.42 & 92.67 & 88.35 & 99.87 & 99.74 & 100    & 100   & 100 & 100  \\
EuroSAT         & 85.91 & 96.22 & 61.02 & 83.74 & 99.87 & 99.46 & 99.98  & 99.96 & 100 & 100 \\
\midrule
\textbf{Average} & 74.22 & \textbf{95.44}~\textcolor{deltapos}{\scriptsize(+21.22)} & 85.15 & \textbf{93.24}~\textcolor{deltapos}{\scriptsize(+8.09)} & \textbf{99.87} & 99.76~\textcolor{deltaneg}{\scriptsize(-0.11)} & \textbf{99.99} & 99.98~\textcolor{deltaneg}{\scriptsize(-0.01)} & 100 & 100~\textcolor{grouplabel}{\scriptsize(0.00)} \\
\bottomrule
\end{tabular}}
\caption{\textbf{Per-dataset breakdown of Table~4~\emph{(main paper)}.} Percentage of clean
inputs and PGD-10 adversarial inputs, at $\epsilon\in\{1,4,8,16\}/255$,
correctly assigned to the intervention gate under the drift-only score
$s_{\mathrm{int}}(x)=r(x)$ and the adopted fused score
$s_{\mathrm{int}}(x)=r(x)+J(x)$, evaluated separately on each of the 12
downstream datasets. The Average row reproduces the aggregate values from
Table~4 of the main paper for cross-checking.}
\label{tab:detection_accuracy_per_dataset}
\end{table*}

\begin{table}[!htbp]
\centering
\scriptsize
\renewcommand{\arraystretch}{1.15}
\setlength{\tabcolsep}{3pt}
\resizebox{\columnwidth}{!}{%
\begin{tabular}{
    l |
    >{\color{gray}\columncolor{gray!10}}c |
    c |
    cc |
    >{\columncolor{oursbg}}c
}
\toprule
\multirow{3}{*}{\bfseries Setting}
& \multicolumn{1}{c|}{\itshape\textcolor{gray}{Zero-shot}}
& \multicolumn{1}{c|}{\itshape\textcolor{grouplabel}{Adv. Finetuning}}
& \multicolumn{3}{c}{\itshape\textcolor{grouplabel}{Test-time Defense}} \\
\cmidrule(lr){2-6}

& \textcolor{gray}{\bfseries CLIP}
& \bfseries CLIP-FT
& \bfseries TTE
& \bfseries HD
& \bfseries \method{} \\

& {\scriptsize\itshape\textcolor{gray}{ICML '21}}
& {}
& {}
& {}
& \textcolor{oursaccent}{$\bigstar$} \\

\midrule
Clean
& 66.52
& {60.31}
& {66.98}
& {61.34}
& 65.38 \\

$\epsilon=1/255$
& 3.40
& {3.62}
& {38.09}
& {13.56}
& 58.47 \\

$\epsilon=4/255$
& 0.07
& {1.27}
& {9.33}
& {0.58}
& 61.04 \\

$\epsilon=8/255$
& 0.03
& {0.42}
& {4.12}
& {0.0}
& 59.92 \\

$\epsilon=16/255$
& 0.02
& {0.11}
& {1.13}
& {0.0}
& 59.83 \\

\bottomrule
\end{tabular}%
}
\vspace{-0.5em}
\caption{\textbf{Comparison with additional baselines.} Clean and PGD-10
robust accuracy (\%) at $\epsilon\in\{1,4,8,16\}/255$ for zero-shot CLIP,
CLIP-FT, TTE, HD, and \method{}, averaged over the 12 downstream datasets
under the same evaluation protocol as Table~1 of the main paper.}
\label{tab:main_table1_results_with_other_baselines}
\vspace{-1em}
\end{table}

\begin{table*}[!htbp]
\centering
\renewcommand{\arraystretch}{1.20}
\setlength{\tabcolsep}{6pt}
\resizebox{\textwidth}{!}{
\begin{tabular}{l| >{\color{gray}\columncolor{gray!10}}c|ccc|ccccccc|>{\columncolor{oursbg}}c}
\toprule
\multirow{3}{*}{\bfseries Setting} & \multicolumn{1}{c|}{\scriptsize\itshape\textcolor{gray}{Zero-shot}} & \multicolumn{3}{c|}{\itshape\textcolor{grouplabel}{Adversarial Finetuning}} & \multicolumn{8}{c}{\itshape\textcolor{grouplabel}{Test-time Defense}}  \\
\cmidrule(lr){2-13}
 & \textcolor{gray}{\bfseries CLIP}  & \bfseries TeCoA & \bfseries PMG-AFT & \bfseries FARE & \bfseries Anti-Adv & \bfseries TTC & \bfseries DOC & \bfseries SS-TPT & \bfseries AOM & \bfseries MAC & \bfseries Defend-CLIP & \bfseries \method{} \\
 & {\scriptsize\itshape\textcolor{gray}{(ICML '21)}} &  {\scriptsize\itshape\textcolor{iclrcol}{(ICLR '23)}} & {\scriptsize\itshape\textcolor{cvpr25col}{(CVPR '24)}} & {\scriptsize\itshape\textcolor{icmlcol}{(ICML '24)}} & {\scriptsize\itshape\textcolor{aaaicol}{(AAAI '22)}} & {\scriptsize\itshape\textcolor{cvpr25col}{(CVPR '25)}} & {\scriptsize\itshape\textcolor{aaaicol}{(AAAI '26)}} & {\scriptsize\itshape\textcolor{icmlcol}{(ICML '26)}} & {\scriptsize\itshape\textcolor{cvpr25col}{(CVPR '25)}} & {\scriptsize\itshape\textcolor{cvpr26col}{(CVPR '26)}} & {\scriptsize\itshape\textcolor{cvpr26col}{(CVPR '26)}} & \textcolor{oursaccent}{~$\bigstar$}\\
\midrule
Clean & 66.52  & 43.15 & 45.66 & 54.41 & 61.93 & 62.57 & 62.87 & 63.27 & 50.60 & \cellcolor{bestcell}\textbf{66.14} & 62.70 & \underline{65.38}~\textcolor{deltaneg}{\scriptsize(-0.76)} \\
$\epsilon=1/255$ & 3.40  & 26.40 & 30.67 & 22.83 & 14.15 & 32.35 & 43.65 & 55.19 & \underline{55.86} & 23.72 & 47.42 & \cellcolor{bestcell}\textbf{58.47}~\textcolor{deltapos}{\scriptsize(+2.61)} \\
$\epsilon=4/255$ & 0.07  & 4.73 & 5.63 & 0.87 & 0.67 & 23.31 & 32.44 & 36.66 & \underline{49.62} & 33.29 & 46.85 & \cellcolor{bestcell}\textbf{61.04}~\textcolor{deltapos}{\scriptsize(+11.42)} \\
$\epsilon=8/255$ & 0.03  & 0.40 & 0.47 & 0.13 & 0.22 & 9.74 & 9.77 & 24.22 & \underline{30.83} & 18.98 & 28.14 & \cellcolor{bestcell}\textbf{59.92}~\textcolor{deltapos}{\scriptsize(+29.09)} \\
$\epsilon=16/255$ & 0.02  & 0.12 & 0.13 & 0.08 & 0.08 & 5.52 & 2.08 & 13.10 & \underline{13.33} & 8.75 & 10.64 & \cellcolor{bestcell}\textbf{59.83}~\textcolor{deltapos}{\scriptsize(+46.50)} \\
$\epsilon=24/255$ & 0.0 & 0.01 & 0.04 & 0.0 & 0.0 & 3.21 & 3.01 & 7.21 & \underline{7.28} & 4.74  & 6.43 & \cellcolor{bestcell}\textbf{58.64}~\textcolor{deltapos}{\scriptsize(+51.36)} \\
$\epsilon=32/255$ & 0.0 & 0.0 & 0.0 & 0.0 & 0.0 & 0.20 & 0.10 & 4.56 & \underline{4.77} &0.90 & 3.21 & \cellcolor{bestcell}\textbf{58.89}~\textcolor{deltapos}{\scriptsize(+54.12)}   \\
\bottomrule
\end{tabular}
}
\vspace{-1em}
\caption{\textbf{Extension of Table~1~\emph{(main paper)} to $\epsilon\in\{24,32\}/255$.}
Clean and PGD-10 robust accuracy (\%) at $\epsilon\in\{1,4,8,16,24,32\}/255$,
averaged over the 12 downstream datasets. The budgets $\epsilon=24/255$
and $\epsilon=32/255$ lie beyond the range used to calibrate the
correction-strength mapping (Eq.~11), evaluating whether \method{}
extrapolates to attack strengths not seen during calibration while
existing defenses continue to degrade.}
\label{tab:main_table1_results_with_eps24_and_eps32}
\vspace{-1em}
\end{table*}

\begin{table*}[!htbp]
\centering
\renewcommand{\arraystretch}{1.20}
\setlength{\tabcolsep}{12pt}
\resizebox{\textwidth}{!}{\tiny
\begin{tabular}{ll| ccccc|>{\columncolor{oursbg}}c}
\toprule
\multirow{3}{*}{\bfseries Dataset} & \multirow{3}{*}{\bfseries Setting} & \multicolumn{6}{c}{\itshape\textcolor{grouplabel}{Test-time Defense}} \\
\cmidrule(lr){3-7} \cmidrule(lr){8-8}
 & & \bfseries TTC & \bfseries DOC & \bfseries MAC & \bfseries AOM & \bfseries Defend-CLIP & \bfseries \method{} \\
 & & {\scriptsize\itshape\textcolor{cvpr25col}{(CVPR '25)}} & {\scriptsize\itshape\textcolor{aaaicol}{(AAAI '26)}} & {\scriptsize\itshape\textcolor{cvpr26col}{(CVPR '26)}} & {\scriptsize\itshape\textcolor{cvpr25col}{(CVPR '25)}} & {\scriptsize\itshape\textcolor{cvpr26col}{(CVPR '26)}} & \textcolor{oursaccent}{~$\bigstar$} \\
\midrule
\multirow{5}{*}{Caltech101}
 & Clean & \underline{86.06} & \cellcolor{bestcell}\textbf{86.19} & 85.81 & 81.85 & 85.27 & 85.42~\textcolor{deltaneg}{\scriptsize(-0.77)} \\
 & $\epsilon=1/255$ & 52.33 & 66.42 & 50.94 & \cellcolor{bestcell}\textbf{83.59} & 67.27 & \underline{81.81}~\textcolor{deltaneg}{\scriptsize(-1.78)} \\
 & $\epsilon=4/255$ & 45.66 & 59.94 & 76.37 & \underline{84.12} & 80.36 & \cellcolor{bestcell}\textbf{84.83}~\textcolor{deltapos}{\scriptsize(+0.71)} \\
 & $\epsilon=8/255$ & 19.08 & 28.54 & 55.32 & \underline{65.89} & 63.21 & \cellcolor{bestcell}\textbf{84.36}~\textcolor{deltapos}{\scriptsize(+18.47)} \\
 & $\epsilon=16/255$ & 4.35 & 4.70 & \underline{25.88} & 16.73 & 14.06 & \cellcolor{bestcell}\textbf{85.49}~\textcolor{deltapos}{\scriptsize(+59.61)} \\
\midrule
\multirow{5}{*}{CIFAR10}
 & Clean & 81.30 & 81.15 & \underline{84.09} & 55.10 & 83.61 & \cellcolor{bestcell}\textbf{84.49}~\textcolor{deltapos}{\scriptsize(+0.40)} \\
 & $\epsilon=1/255$ & 24.44 & 41.38 & 47.76 & \underline{73.65} & 41.43 & \cellcolor{bestcell}\textbf{74.30}~\textcolor{deltapos}{\scriptsize(+0.65)} \\
 & $\epsilon=4/255$ & 34.67 & 43.56 & 47.15 & \cellcolor{bestcell}\textbf{83.38} & 79.19 & \underline{83.00}~\textcolor{deltaneg}{\scriptsize(-0.38)} \\
 & $\epsilon=8/255$ & 14.97 & 18.16 & 22.06 & \underline{76.72} & 72.70 & \cellcolor{bestcell}\textbf{83.67}~\textcolor{deltapos}{\scriptsize(+6.95)} \\
 & $\epsilon=16/255$ & 5.39 & 4.94 & 5.67 & \underline{28.96} & 24.94 & \cellcolor{bestcell}\textbf{84.51}~\textcolor{deltapos}{\scriptsize(+55.55)} \\
\midrule
\multirow{5}{*}{CIFAR100}
 & Clean & 56.37 & \underline{56.43} & \cellcolor{bestcell}\textbf{57.25} & 27.68 & 56.13 & 56.06~\textcolor{deltaneg}{\scriptsize(-1.19)} \\
 & $\epsilon=1/255$ & 16.23 & 25.87 & 28.20 & \underline{37.57} & 18.73 & \cellcolor{bestcell}\textbf{46.59}~\textcolor{deltapos}{\scriptsize(+9.02)} \\
 & $\epsilon=4/255$ & 19.35 & 24.04 & 26.00 & \underline{50.41} & 45.65 & \cellcolor{bestcell}\textbf{55.62}~\textcolor{deltapos}{\scriptsize(+5.21)} \\
 & $\epsilon=8/255$ & 10.62 & 12.05 & 9.26 & \underline{44.33} & 39.65 & \cellcolor{bestcell}\textbf{55.15}~\textcolor{deltapos}{\scriptsize(+10.82)} \\
 & $\epsilon=16/255$ & 8.26 & 8.12 & 1.66 & \underline{14.53} & 9.85 & \cellcolor{bestcell}\textbf{56.83}~\textcolor{deltapos}{\scriptsize(+42.30)} \\
\midrule
\multirow{5}{*}{DTD}
 & Clean & 35.85 & 35.05 & \cellcolor{bestcell}\textbf{40.00} & 26.65 & 35.48 & \underline{39.63}~\textcolor{deltaneg}{\scriptsize(-0.37)} \\
 & $\epsilon=1/255$ & 24.68 & 29.79 & 18.35 & \underline{31.33} & 25.96 & \cellcolor{bestcell}\textbf{32.77}~\textcolor{deltapos}{\scriptsize(+1.44)} \\
 & $\epsilon=4/255$ & 18.13 & 24.26 & 32.13 & \underline{33.30} & 30.16 & \cellcolor{bestcell}\textbf{38.30}~\textcolor{deltapos}{\scriptsize(+5.00)} \\
 & $\epsilon=8/255$ & 6.75 & 8.35 & 17.29 & \underline{21.44} & 18.30 & \cellcolor{bestcell}\textbf{37.24}~\textcolor{deltapos}{\scriptsize(+15.80)} \\
 & $\epsilon=16/255$ & 3.77 & 4.52 & 5.27 & \underline{7.13} & 3.99 & \cellcolor{bestcell}\textbf{39.63}~\textcolor{deltapos}{\scriptsize(+32.50)} \\
\midrule
\multirow{5}{*}{FGVCAircraft}
 & Clean & 15.78 & 15.30 & \cellcolor{bestcell}\textbf{19.71} & 12.60 & 15.57 & \underline{19.26}~\textcolor{deltaneg}{\scriptsize(-0.45)} \\
 & $\epsilon=1/255$ & 8.28 & 11.94 & 3.54 & \cellcolor{bestcell}\textbf{14.76} & 9.84 & \underline{13.98}~\textcolor{deltaneg}{\scriptsize(-0.78)} \\
 & $\epsilon=4/255$ & 8.88 & 12.39 & 15.81 & \underline{16.05} & 13.77 & \cellcolor{bestcell}\textbf{17.61}~\textcolor{deltapos}{\scriptsize(+1.56)} \\
 & $\epsilon=8/255$ & 4.29 & 4.83 & 6.72 & \underline{9.36} & 7.17 & \cellcolor{bestcell}\textbf{17.76}~\textcolor{deltapos}{\scriptsize(+8.40)} \\
 & $\epsilon=16/255$ & 2.76 & 3.30 & 1.14 & \underline{4.38} & 2.19 & \cellcolor{bestcell}\textbf{17.76}~\textcolor{deltapos}{\scriptsize(+13.38)} \\
\midrule
\multirow{5}{*}{Flowers102}
 & Clean & 63.08 & 62.49 & \cellcolor{bestcell}\textbf{64.95} & 50.67 & 61.72 & \underline{64.92}~\textcolor{deltaneg}{\scriptsize(-0.03)} \\
 & $\epsilon=1/255$ & 29.45 & 42.20 & 11.74 & \underline{55.85} & 48.90 & \cellcolor{bestcell}\textbf{58.87}~\textcolor{deltapos}{\scriptsize(+3.02)} \\
 & $\epsilon=4/255$ & 27.53 & 40.97 & 46.43 & \underline{59.18} & 55.33 & \cellcolor{bestcell}\textbf{64.14}~\textcolor{deltapos}{\scriptsize(+4.96)} \\
 & $\epsilon=8/255$ & 6.79 & 8.52 & 25.00 & \underline{36.09} & 32.28 & \cellcolor{bestcell}\textbf{63.69}~\textcolor{deltapos}{\scriptsize(+27.60)} \\
 & $\epsilon=16/255$ & 2.78 & 2.86 & 4.28 & \underline{6.62} & 2.81 & \cellcolor{bestcell}\textbf{64.35}~\textcolor{deltapos}{\scriptsize(+57.73)} \\
\midrule
\multirow{5}{*}{OxfordPet}
 & Clean & 65.49 & 71.98 & \cellcolor{bestcell}\textbf{87.35} & 78.60 & 80.89 & \underline{85.36}~\textcolor{deltaneg}{\scriptsize(-1.99)} \\
 & $\epsilon=1/255$ & 46.71 & 64.02 & 18.53 & \underline{83.29} & 74.93 & \cellcolor{bestcell}\textbf{83.84}~\textcolor{deltapos}{\scriptsize(+0.55)} \\
 & $\epsilon=4/255$ & 43.82 & 71.98 & 69.86 & \cellcolor{bestcell}\textbf{86.48} & 85.25 & \underline{85.39}~\textcolor{deltaneg}{\scriptsize(-1.09)} \\
 & $\epsilon=8/255$ & 11.71 & 18.10 & 39.71 & \underline{70.02} & 68.82 & \cellcolor{bestcell}\textbf{83.40}~\textcolor{deltapos}{\scriptsize(+13.38)} \\
 & $\epsilon=16/255$ & 1.98 & 2.56 & 7.63 & \underline{11.23} & 10.03 & \cellcolor{bestcell}\textbf{83.16}~\textcolor{deltapos}{\scriptsize(+71.93)} \\
\midrule
\multirow{5}{*}{StanfordCars}
 & Clean & 41.20 & 41.82 & \cellcolor{bestcell}\textbf{51.65} & 43.45 & 47.27 & \underline{49.84}~\textcolor{deltaneg}{\scriptsize(-1.81)} \\
 & $\epsilon=1/255$ & 15.19 & 26.13 & 6.53 & \cellcolor{bestcell}\textbf{45.80} & 40.43 & \underline{45.24}~\textcolor{deltaneg}{\scriptsize(-0.56)} \\
 & $\epsilon=4/255$ & 19.73 & 32.10 & 36.41 & \underline{44.15} & 40.72 & \cellcolor{bestcell}\textbf{48.79}~\textcolor{deltapos}{\scriptsize(+4.64)} \\
 & $\epsilon=8/255$ & 5.27 & 7.71 & 15.55 & \underline{17.61} & 14.25 & \cellcolor{bestcell}\textbf{49.00}~\textcolor{deltapos}{\scriptsize(+31.39)} \\
 & $\epsilon=16/255$ & 3.33 & 4.48 & 2.20 & \underline{6.83} & 3.47 & \cellcolor{bestcell}\textbf{50.06}~\textcolor{deltapos}{\scriptsize(+43.23)} \\
\midrule
\multirow{5}{*}{STL10}
 & Clean & 95.96 & 95.90 & \underline{96.28} & 90.09 & 95.07 & \cellcolor{bestcell}\textbf{96.32}~\textcolor{deltapos}{\scriptsize(+0.04)} \\
 & $\epsilon=1/255$ & 65.73 & 80.49 & 65.83 & \underline{92.54} & 74.99 & \cellcolor{bestcell}\textbf{93.54}~\textcolor{deltapos}{\scriptsize(+1.00)} \\
 & $\epsilon=4/255$ & 64.22 & 77.90 & 87.62 & \underline{92.54} & 91.75 & \cellcolor{bestcell}\textbf{95.76}~\textcolor{deltapos}{\scriptsize(+3.22)} \\
 & $\epsilon=8/255$ & 23.47 & 31.24 & 65.45 & \underline{79.14} & 78.43 & \cellcolor{bestcell}\textbf{96.00}~\textcolor{deltapos}{\scriptsize(+16.86)} \\
 & $\epsilon=16/255$ & 4.03 & 3.08 & \underline{26.82} & 17.27 & 16.56 & \cellcolor{bestcell}\textbf{96.43}~\textcolor{deltapos}{\scriptsize(+69.61)} \\
\midrule
\multirow{5}{*}{\bfseries Average}
 & Clean & 60.12 & 60.70 & \cellcolor{bestcell}\textbf{65.23} & 51.85 & 62.33 & \underline{64.59}~\textcolor{deltaneg}{\scriptsize(-0.64)} \\
 & $\epsilon=1/255$ & 31.45 & 43.14 & 27.94 & \underline{57.60} & 44.72 & \cellcolor{bestcell}\textbf{58.99}~\textcolor{deltapos}{\scriptsize(+1.39)} \\
 & $\epsilon=4/255$ & 31.33 & 43.02 & 48.64 & \underline{61.07} & 58.02 & \cellcolor{bestcell}\textbf{63.67}~\textcolor{deltapos}{\scriptsize(+2.60)} \\
 & $\epsilon=8/255$ & 11.44 & 15.28 & 28.48 & \underline{46.73} & 43.87 & \cellcolor{bestcell}\textbf{63.37}~\textcolor{deltapos}{\scriptsize(+16.64)} \\
 & $\epsilon=16/255$ & 4.07 & 4.28 & 8.95 & \underline{12.63} & 9.77 & \cellcolor{bestcell}\textbf{64.25}~\textcolor{deltapos}{\scriptsize(+51.62)} \\
\bottomrule
\end{tabular}
}
\caption{\textbf{Per-dataset breakdown of Table~5~\emph{(main paper)}.} Clean and PGD-100
robust accuracy (\%) at $\epsilon\in\{1,4,8,16\}/255$ for each of the 9
downstream datasets used in this evaluation (excluding EuroSAT,
Caltech256, and Food101), alongside the 9-dataset average reported in
Table~5 of the main paper. Values in parentheses denote the difference
between \method{} and the strongest baseline in each row.}
\label{tab:main_table5_results_per_dataset}
\end{table*}

\begin{table*}[!htbp]
\centering
\renewcommand{\arraystretch}{1.25}
\setlength{\tabcolsep}{8pt}
\resizebox{\textwidth}{!}{\tiny
\begin{tabular}{ll| cccccc|>{\columncolor{oursbg}}c}
\toprule
\multirow{3}{*}{\bfseries Dataset} & \multirow{3}{*}{\bfseries Setting} & \multicolumn{7}{c}{\itshape\textcolor{grouplabel}{Test-time Defense}} \\
\cmidrule(lr){3-8} \cmidrule(lr){9-9}
 & & \bfseries TTC & \bfseries DOC & \bfseries SS-TPT & \bfseries AOM & \bfseries MAC & \bfseries Defend-CLIP & \bfseries \method{} \\
 & & {\scriptsize\itshape\textcolor{cvpr25col}{(CVPR '25)}} & {\scriptsize\itshape\textcolor{aaaicol}{(AAAI '26)}} & {\scriptsize\itshape\textcolor{icmlcol}{(ICML '26)}} & {\scriptsize\itshape\textcolor{cvpr25col}{(CVPR '25)}} & {\scriptsize\itshape\textcolor{cvpr26col}{(CVPR '26)}} & {\scriptsize\itshape\textcolor{cvpr26col}{(CVPR '26)}} & \textcolor{oursaccent}{~$\bigstar$} \\
\midrule
\multirow{5}{*}{\bfseries ImageNet-A}
 & Clean & 19.50 & 18.79 & \cellcolor{bestcell}\textbf{27.13} & 10.95 & \underline{19.60} & 16.08 & 18.79~\textcolor{deltaneg}{\scriptsize(-8.34)} \\
 & $\epsilon=1/255$ & 9.05 & 12.86 & \underline{15.97} & 12.96 & 5.03 & 10.55 & \cellcolor{bestcell}\textbf{15.98}~\textcolor{deltapos}{\scriptsize(+0.01)} \\
 & $\epsilon=4/255$ & 4.72 & 6.63 & 6.63 & \underline{8.94} & 5.93 & 7.54 & \cellcolor{bestcell}\textbf{16.48}~\textcolor{deltapos}{\scriptsize(+7.54)} \\
 & $\epsilon=8/255$ & 2.81 & 3.52 & \underline{3.71} & 3.52 & 1.91 & 2.11 & \cellcolor{bestcell}\textbf{15.48}~\textcolor{deltapos}{\scriptsize(+11.77)} \\
 & $\epsilon=16/255$ & 2.61 & \underline{3.02} & 1.30 & 2.31 & 0.20 & 0.90 & \cellcolor{bestcell}\textbf{16.08}~\textcolor{deltapos}{\scriptsize(+13.06)} \\
\midrule
\multirow{5}{*}{\bfseries ImageNet}
 & Clean & 35.50 & 38.60 & \cellcolor{bestcell}\textbf{63.35} & 50.88 & \underline{61.68} & 54.56 & 57.94~\textcolor{deltaneg}{\scriptsize(-5.41)} \\
 & $\epsilon=1/255$ & 28.26 & 40.40 & 51.91 & \underline{53.72} & 18.50 & 49.48 & \cellcolor{bestcell}\textbf{56.20}~\textcolor{deltapos}{\scriptsize(+2.48)} \\
 & $\epsilon=4/255$ & 14.84 & 25.58 & 34.18 & \underline{40.26} & 30.34 & 38.60 & \cellcolor{bestcell}\textbf{55.84}~\textcolor{deltapos}{\scriptsize(+15.58)} \\
 & $\epsilon=8/255$ & 3.84 & 6.32 & \underline{22.92} & 15.66 & 11.86 & 14.00 & \cellcolor{bestcell}\textbf{53.42}~\textcolor{deltapos}{\scriptsize(+30.50)} \\
 & $\epsilon=16/255$ & 2.66 & 4.38 & \underline{12.09} & 7.86 & 4.72 & 6.20 & \cellcolor{bestcell}\textbf{53.04}~\textcolor{deltapos}{\scriptsize(+40.95)} \\
\midrule
\multirow{5}{*}{\bfseries ImageNet-S}
 & Clean & 31.32 & 31.42 & \cellcolor{bestcell}\textbf{42.52} & 35.98 & \underline{41.46} & 40.64 & 40.86~\textcolor{deltaneg}{\scriptsize(-1.66)} \\
 & $\epsilon=1/255$ & 23.98 & 32.42 & \underline{33.87} & \cellcolor{bestcell}\textbf{36.36} & 15.64 & 26.36 & 31.32~\textcolor{deltaneg}{\scriptsize(-5.04)} \\
 & $\epsilon=4/255$ & 17.92 & 26.44 & 23.79 & \underline{29.68} & 22.30 & 24.18 & \cellcolor{bestcell}\textbf{34.60}~\textcolor{deltapos}{\scriptsize(+4.92)} \\
 & $\epsilon=8/255$ & 7.84 & 12.66 & \underline{17.57} & 14.46 & 15.36 & 10.24 & \cellcolor{bestcell}\textbf{30.06}~\textcolor{deltapos}{\scriptsize(+12.49)} \\
 & $\epsilon=16/255$ & 3.30 & 4.68 & \underline{11.59} & 8.68 & 8.82 & 4.46 & \cellcolor{bestcell}\textbf{31.54}~\textcolor{deltapos}{\scriptsize(+19.95)} \\
\midrule
\multirow{5}{*}{\bfseries ImageNet-R}
 & Clean & 50.10 & 49.10 & \cellcolor{bestcell}\textbf{63.10} & 54.90 & \underline{60.00} & 58.70 & 58.80~\textcolor{deltaneg}{\scriptsize(-4.30)} \\
 & $\epsilon=1/255$ & 35.50 & 45.60 & 51.60 & \cellcolor{bestcell}\textbf{54.30} & 22.50 & 47.50 & \underline{53.20}~\textcolor{deltaneg}{\scriptsize(-1.10)} \\
 & $\epsilon=4/255$ & 19.00 & 31.90 & 36.30 & \underline{43.30} & 31.00 & 40.10 & \cellcolor{bestcell}\textbf{53.90}~\textcolor{deltapos}{\scriptsize(+10.60)} \\
 & $\epsilon=8/255$ & 4.90 & 8.00 & \underline{27.90} & 16.00 & 15.00 & 12.90 & \cellcolor{bestcell}\textbf{52.50}~\textcolor{deltapos}{\scriptsize(+24.60)} \\
 & $\epsilon=16/255$ & 2.90 & 4.70 & \underline{18.00} & 7.60 & 7.60 & 4.50 & \cellcolor{bestcell}\textbf{51.80}~\textcolor{deltapos}{\scriptsize(+33.80)} \\
\midrule
\multirow{5}{*}{\bfseries ImageNet-V2}
 & Clean & 41.60 & 40.38 & \cellcolor{bestcell}\textbf{57.23} & 42.62 & \underline{54.70} & 47.80 & 51.16~\textcolor{deltaneg}{\scriptsize(-6.07)} \\
 & $\epsilon=1/255$ & 23.52 & 34.52 & \underline{45.41} & 45.18 & 16.60 & 39.82 & \cellcolor{bestcell}\textbf{47.76}~\textcolor{deltapos}{\scriptsize(+2.35)} \\
 & $\epsilon=4/255$ & 12.26 & 21.50 & 29.03 & \underline{34.42} & 25.56 & 31.64 & \cellcolor{bestcell}\textbf{47.54}~\textcolor{deltapos}{\scriptsize(+13.12)} \\
 & $\epsilon=8/255$ & 3.76 & 5.56 & \underline{18.92} & 13.76 & 9.50 & 11.00 & \cellcolor{bestcell}\textbf{44.64}~\textcolor{deltapos}{\scriptsize(+25.72)} \\
 & $\epsilon=16/255$ & 2.80 & 3.96 & \underline{8.93} & 7.50 & 3.82 & 5.74 & \cellcolor{bestcell}\textbf{44.40}~\textcolor{deltapos}{\scriptsize(+35.47)} \\
\midrule
\multirow{5}{*}{\bfseries Average}
 & Clean & 35.60 & 35.66 & \cellcolor{bestcell}\textbf{50.67} & 39.07 & \underline{47.49} & 43.56 & 45.78~\textcolor{deltaneg}{\scriptsize(-4.89)} \\
 & $\epsilon=1/255$ & 24.06 & 33.16 & 39.75 & \underline{40.50} & 15.65 & 34.74 & \cellcolor{bestcell}\textbf{41.34}~\textcolor{deltapos}{\scriptsize(+0.84)} \\
 & $\epsilon=4/255$ & 13.75 & 22.41 & 25.99 & \underline{31.32} & 23.03 & 28.41 & \cellcolor{bestcell}\textbf{42.97}~\textcolor{deltapos}{\scriptsize(+11.65)} \\
 & $\epsilon=8/255$ & 4.63 & 7.21 & \underline{18.20} & 12.68 & 10.73 & 10.05 & \cellcolor{bestcell}\textbf{41.50}~\textcolor{deltapos}{\scriptsize(+23.30)} \\
 & $\epsilon=16/255$ & 2.85 & 4.15 & \underline{10.38} & 6.79 & 5.03 & 4.16 & \cellcolor{bestcell}\textbf{41.84}~\textcolor{deltapos}{\scriptsize(+31.46)} \\
\bottomrule
\end{tabular}
}
\caption{\textbf{Per-dataset breakdown of Table~8~\emph{(main paper)}.} Clean and PGD-10
robust accuracy (\%) at $\epsilon\in\{1,4,8,16\}/255$ for ImageNet and its
four distribution-shifted variants (ImageNet-A, ImageNet-S, ImageNet-R,
ImageNet-V2), alongside the 5-dataset average reported in Table~8 of the
main paper. Values in parentheses denote the difference between
\method{} and the strongest baseline in each row.}
\label{tab:main_table8_results_per_dataset}
\end{table*}

\paragraph{Per-Dataset Breakdown of Table~1~\emph{(main paper)}: Consistency of Robustness Gains Across Datasets.}
Table~\ref{tab:main_table1_per_dataset_results} disaggregates the 12-dataset
average reported in Table~1~\emph{(main paper)}  into its per-dataset clean
and PGD-10 robust accuracy, allowing us to verify that the aggregate trend
in Table~1~\emph{(main paper)} is not driven by a small subset of the evaluation suite. At
$\epsilon\geq4/255$, \method{} outperforms every baseline on all 12
datasets, and this margin widens consistently as the attack strengthens,
mirroring the pattern observed in the main aggregate results. At
$\epsilon=1/255$, the picture is more mixed: on a subset of datasets the
strongest baseline remains marginally ahead, consistent with the
explanation given in the main text that distinguishing weak adversarial
examples from clean inputs is the most difficult regime for the defensive
intervention score. The overall average nonetheless favors \method{} even
at this budget, indicating that the occasional shortfall on individual
datasets does not reflect a systematic weakness, and is fully consistent
with the aggregate numbers already reported in Table~1~\emph{(main paper)}.

Notably, the identity of the strongest baseline in Table~1~\emph{(main paper)} is itself not
fixed across datasets or attack budgets once the comparison is broken down
at this level. Different baselines lead on different datasets, and even
within a single dataset the strongest baseline often changes between the
clean setting and each attack budget. This pattern recurs throughout
Table~\ref{tab:main_table1_per_dataset_results} and reinforces the central
observation of Figures~1 and~3~\emph{(main paper)}: a defense configured with a single fixed
correction strength cannot simultaneously serve every dataset and every
attack budget, whereas \method{}'s per-input correction scale adapts
across this same evaluation suite without any per-dataset retuning,
supporting the single-configuration protocol used to obtain Table~1~\emph{(main paper)}.

Clean accuracy is preserved closely across the large majority of the
suite, remaining close to the strongest baseline on most datasets. This
indicates that the response-conditioned correction and the defensive
intervention gate together avoid materially degrading clean performance
even though the correction-strength mapping is calibrated once, on a
single dataset, and held fixed across the entire evaluation suite. Taken
together with the consistent robustness gains at $\epsilon\geq4/255$,
these per-dataset results support the conclusion that the gains reported
in Table~1~\emph{(main paper)} generalize across diverse visual domains, from natural-image
and fine-grained recognition datasets to texture and satellite-imagery
datasets, rather than being an artifact of the datasets used for
calibration or of the 12-dataset averaging in Table~1~\emph{(main paper)}.

\paragraph{Per-Dataset Breakdown of Table~2~\emph{(main paper)}: Consistency of the Two Components.}
Table~\ref{tab:adaptive_correction_per_dataset} disaggregates the ablation
in Table~2~\emph{(main paper)}  across the 12 downstream datasets, isolating
the effect of response-conditioned anchor construction and of the fused
defensive intervention score on each individual dataset rather than only
on their average.

Response-conditioned anchor construction improves robust accuracy at every
attack budget $\epsilon\geq1/255$ on every dataset in the suite, with the
gain growing substantially larger as the attack strengthens, mirroring the
progression already observed in the 12-dataset average. This confirms that
the benefit of constructing the anchor from a sample-specific noise scale,
rather than a single globally fixed scale, is not confined to a subset of
datasets or to the aggregate statistic, but holds uniformly across natural,
fine-grained, and texture or satellite-imagery domains alike. On clean
inputs, however, this component alone has a small and inconsistent effect
across datasets, sometimes marginally positive and sometimes marginally
negative, since adapting only the correction scale does not yet address
whether correction should be applied at all.

Adding the fused defensive intervention score improves clean accuracy on
nearly every dataset in the suite, with only a negligible exception, and
improves accuracy at $\epsilon=1/255$ on most datasets as well, with a
small number of exceptions where the weak-attack accuracy decreases
slightly. This is consistent with the explanation given in the main text:
the fused score trades a small amount of weak-attack correction for a more
reliable separation of clean and weakly attacked inputs, and this
trade-off is occasionally unfavorable on individual datasets even though
it is favorable on average. At $\epsilon\geq4/255$, the fused score
produces a small and fairly uniform reduction in robust accuracy relative
to the drift-only gate across nearly all datasets, occasionally negligible
or absent altogether, mirroring the small average cost already reported in
Table~2~\emph{(main paper)}. Taken together, these per-dataset results confirm that the two
components of \method{} play distinct and complementary roles
consistently across the evaluation suite: response-conditioned anchor
construction is the primary driver of robustness at moderate and strong
attack budgets, while the fused intervention score primarily protects
clean accuracy and the weak-attack regime, at a small and consistent cost
to robustness at higher budgets.

\paragraph{Per-Dataset Breakdown of Table~4~\emph{(main paper)}: Detection Consistency Across Datasets.}
Table~\ref{tab:detection_accuracy_per_dataset} disaggregates the detection
accuracy reported in Table~4~\emph{(main paper)}  across the 12 downstream
datasets, reporting the percentage of clean and adversarial inputs
correctly assigned to the intervention gate under the drift-only score
$r(x)$ and the adopted fused score $r(x)+J(x)$.

The most notable pattern is not only that the fused score raises the
average detection accuracy on clean inputs and at $\epsilon=1/255$, as
already reported in Table~4~\emph{(main paper)}, but that it substantially narrows the spread
of detection accuracy across datasets. Under the drift-only score, clean
and weak-attack detection accuracy varies considerably from one dataset to
another, with some datasets detected reasonably reliably and others far
less so. Under the fused score, this variance is greatly reduced: the
weakest-performing dataset under the fused score remains far better
separated than the weakest-performing dataset under the drift-only score.
This indicates that the prediction-instability signal $J(x)$ does not
merely improve the easiest cases further, but specifically compensates for
datasets where relative cross-noise drift alone is a poor separator of
clean and weakly attacked inputs, yielding a detector whose reliability is
far more uniform across visual domains.

At $\epsilon\geq4/255$, both intervention scores already achieve
near-ceiling detection accuracy on almost every dataset, and the small
decreases introduced by the fused score are scattered across datasets
rather than concentrated in a particular domain, consistent with the
negligible average cost already reported in Table~4~\emph{(main paper)}. Taken together, these
per-dataset results confirm that the benefit of combining relative
cross-noise drift with prediction instability is a general property of the
fused score rather than an effect driven by a small number of
easily-separable datasets, supporting its use as a single, fixed
intervention criterion across the full evaluation suite.

\paragraph{Comparison with Additional Baselines (Table~\ref{tab:main_table1_results_with_other_baselines}).}
Table~\ref{tab:main_table1_results_with_other_baselines} extends the
comparison in Table~1~\emph{(main paper)} to two further baselines,
TTE~\citep{perez2021enhancing} and HD~\citep{wu2021attacking}, alongside
CLIP-FT, an adversarially finetuned version of vanilla CLIP obtained using
a 2-step PGD attack with both step size and attack budget set to
$1/255$. All methods are evaluated under the same protocol as Table~1~\emph{(main paper)}:
PGD-10 attacks at $\epsilon\in\{1,4,8,16\}/255$, averaged over the 12
downstream datasets.

CLIP-FT, finetuned under a weak $1/255$ budget, retains reasonable clean
accuracy but provides only limited robustness beyond the narrow regime it
was finetuned for, degrading substantially as $\epsilon$ increases, in
line with the general limitation of adversarial finetuning discussed in
the main paper. TTE attains the highest clean accuracy among all methods
compared, including \method{}, showing that a fixed test-time intervention
can preserve, or even marginally exceed, zero-shot clean performance.
However, TTE's robust accuracy collapses in the same manner already
observed for the test-time defenses reported in Table~1~\emph{(main paper)}: it degrades
sharply as the attack strengthens, falling to single-digit accuracy by
$\epsilon=4/255$ and to near-zero accuracy at $\epsilon=16/255$. HD behaves
similarly but starts from a substantially weaker operating point,
remaining below every other method under attack and reaching zero accuracy
at $\epsilon\in\{8,16\}/255$.

These results reinforce the central failure mode identified in the main
paper: methods that apply a fixed intervention, whether through
finetuning or a fixed test-time mechanism, are unable to sustain
robustness once the attack strength moves beyond the narrow regime for
which they were configured or evaluated. \method{} exceeds both additional
test-time baselines by a wide margin at every attack budget
$\epsilon\geq1/255$, while remaining within a small margin of TTE's clean
accuracy, without requiring any knowledge of the attack budget at
inference. This extended comparison indicates that the degradation
pattern motivating \method{} is not specific to the baselines originally
selected for Table~1~\emph{(main paper)}, but recurs across additional finetuning- and
test-time-defense approaches evaluated under the same protocol.

\paragraph{Extrapolation Beyond the Calibrated Range (Table~\ref{tab:main_table1_results_with_eps24_and_eps32}).}
Table~\ref{tab:main_table1_results_with_eps24_and_eps32} extends the
comparison in Table~1~\emph{(main paper)}  to substantially stronger PGD-10
attacks at $\epsilon\in\{24,32\}/255$. These budgets lie entirely outside
the range $\epsilon\in\{1,4,8,16\}/255$ used to calibrate the
correction-strength mapping in Eq.~11~\emph{(main paper)}, and therefore test whether the
calibrated mapping continues to provide effective correction once the
attack strength moves beyond the conditions under which it was fit. This
table provides the full numerical results underlying Figure~9~\emph{(main paper)} and the
extended range shown in Figure~1~\emph{(main paper)}.

Every baseline continues the collapse already visible at
$\epsilon=16/255$ in Table~1~\emph{(main paper)}: by $\epsilon=24/255$, all baselines fall
below $10\%$ accuracy, and by $\epsilon=32/255$, most baselines are at or
near $0\%$, with SS-TPT retaining the highest baseline accuracy at both
budgets and still falling to $4.77\%$ at $\epsilon=32/255$. \method{}, in
contrast, remains close to its accuracy at $\epsilon=16/255$ across both
extended budgets, exceeding the strongest baseline by $51.36$ and $54.12$
percentage points at $\epsilon=24/255$ and $\epsilon=32/255$, respectively.

This behavior indicates that the linear mapping in Eq.~11~\emph{(main paper)}, although fit
using only four calibration points spanning $\epsilon\in\{1,4,8,16\}/255$,
continues to assign a sufficiently large correction scale to inputs
exhibiting even larger relative cross-noise drift, since the adopted
mapping has no predefined upper cap on $\sigma_{\mathrm{corr}}(x)$. We
emphasize that this is an empirical observation over the specific attack
budgets evaluated here, rather than a guarantee that the mapping
extrapolates correctly to arbitrarily large or qualitatively different
perturbations; within the tested range, however, it allows \method{} to
sustain robustness well beyond the conditions used for calibration, in
contrast to every baseline considered.

\paragraph{Per-Dataset Breakdown of Table~5~\emph{(main paper)}: Consistency of Robustness Gains Under PGD-100.}
Table~\ref{tab:main_table5_results_per_dataset} disaggregates the
9-dataset average reported in Table~5~\emph{(main paper)}  into its
per-dataset clean and PGD-100 robust accuracy, examining whether the
robustness of \method{} observed under PGD-10 is preserved when the
attack is optimized for a substantially larger number of iterations.

The pattern closely mirrors the per-dataset breakdown of Table~1~\emph{(main paper)}
(Table~\ref{tab:main_table1_per_dataset_results}). At
$\epsilon\geq4/255$, \method{} outperforms every baseline on all 9
datasets, and this margin widens consistently as the attack strengthens,
matching the trend already observed in the aggregate PGD-100 results and
in the PGD-10 breakdown. At $\epsilon=1/255$, results are more mixed: on a
small subset of datasets the strongest baseline remains marginally ahead,
consistent with the same explanation given for the PGD-10 case, namely
that separating weak adversarial examples from clean inputs is the most
difficult regime for the defensive intervention score. As with Table~1~\emph{(main paper)},
the overall average still favors \method{} at this budget.

The identity of the strongest baseline again shifts across datasets and
attack budgets rather than remaining fixed to a single method, reinforcing
the same conclusion drawn from the per-dataset breakdown of Table~1~\emph{(main paper)}: no
globally fixed correction strategy serves every dataset and every attack
budget simultaneously, whereas \method{}'s per-input correction scale
adapts across the suite without retuning. Clean accuracy is preserved
closely relative to the strongest baseline across the large majority of
datasets, with only minor reductions elsewhere.

Taken together, these results indicate that the robustness gains of
\method{}, and their consistency across individual datasets, are not an
artifact of the specific optimizer used to generate adversarial examples:
the same qualitative pattern observed under PGD-10 in
Table~\ref{tab:main_table1_per_dataset_results} continues to hold when the
attack is optimized for 100 steps instead of 10, across this 9-dataset
subset of the full evaluation suite.

\paragraph{Per-Dataset Breakdown of Table~8~\emph{(main paper)}: ImageNet-Scale Robustness and Distribution Shift.}
Table~\ref{tab:main_table8_results_per_dataset} disaggregates the average
reported in Table~8~\emph{(main paper)} across ImageNet and its four
distribution-shifted variants (ImageNet-A, ImageNet-S, ImageNet-R, and
ImageNet-V2), reporting clean and PGD-10 robust accuracy at
$\epsilon\in\{1,4,8,16\}/255$ for each dataset individually.

At $\epsilon\geq4/255$, \method{} outperforms every baseline on all five
datasets, and this margin widens consistently with attack strength,
mirroring the pattern already established on the 12-dataset suite in
Table~\ref{tab:main_table1_per_dataset_results}. This indicates that the
robustness gains of \method{} extend to ImageNet-scale evaluation and are
not disrupted by the distribution shifts present in the four
out-of-distribution variants. At $\epsilon=1/255$, results are more mixed
than on the downstream-dataset suite, with a couple of datasets showing a
small decrease relative to the strongest baseline at this budget,
consistent with the explanation given elsewhere in the paper that the
weak-attack regime is the most difficult setting for the defensive
intervention score to separate from clean inputs.

\method{} maintains competitive clean accuracy across all five datasets
while delivering substantial robustness gains at every attack budget
$\epsilon\geq4/255$, indicating that the trade-off in clean accuracy
remains modest relative to the scale of the robustness improvement.
Overall, these per-dataset results confirm that the robustness advantage
of \method{} transfers from the 12-dataset downstream suite to
ImageNet-scale classification and to natural distribution shifts,
consistent with the aggregate numbers reported in Table~8~\emph{(main paper)}.

\section{Ablation of \method{}'s Hyperparameters}
\label{sec:appendix_hyperparameter_ablation}

The main configuration of \method{} fixes three hyperparameters beyond
the calibrated coefficients $(a,b)$: the intervention threshold $\tau$,
the number of noise-averaged views $M$ used to construct the anchor
(Eq.~\ref{eq:adaptive-anchor}), and the extrapolation strength $\alpha$
(Eq.~\ref{eq:final-feature}). This appendix reports PGD-10 accuracy,
averaged across the 12 downstream datasets, as each of these three
hyperparameters is varied independently while the other two are held at
their main-configuration values ($\tau{=}0.7$, $M{=}10$, $\alpha{=}2.0$),
motivating the specific operating point adopted throughout the main
results.

\subsection{Intervention Threshold}
\label{sec:appendix_hp_threshold}

\begin{figure}[!htbp]
\centering
\includegraphics[width=\linewidth]{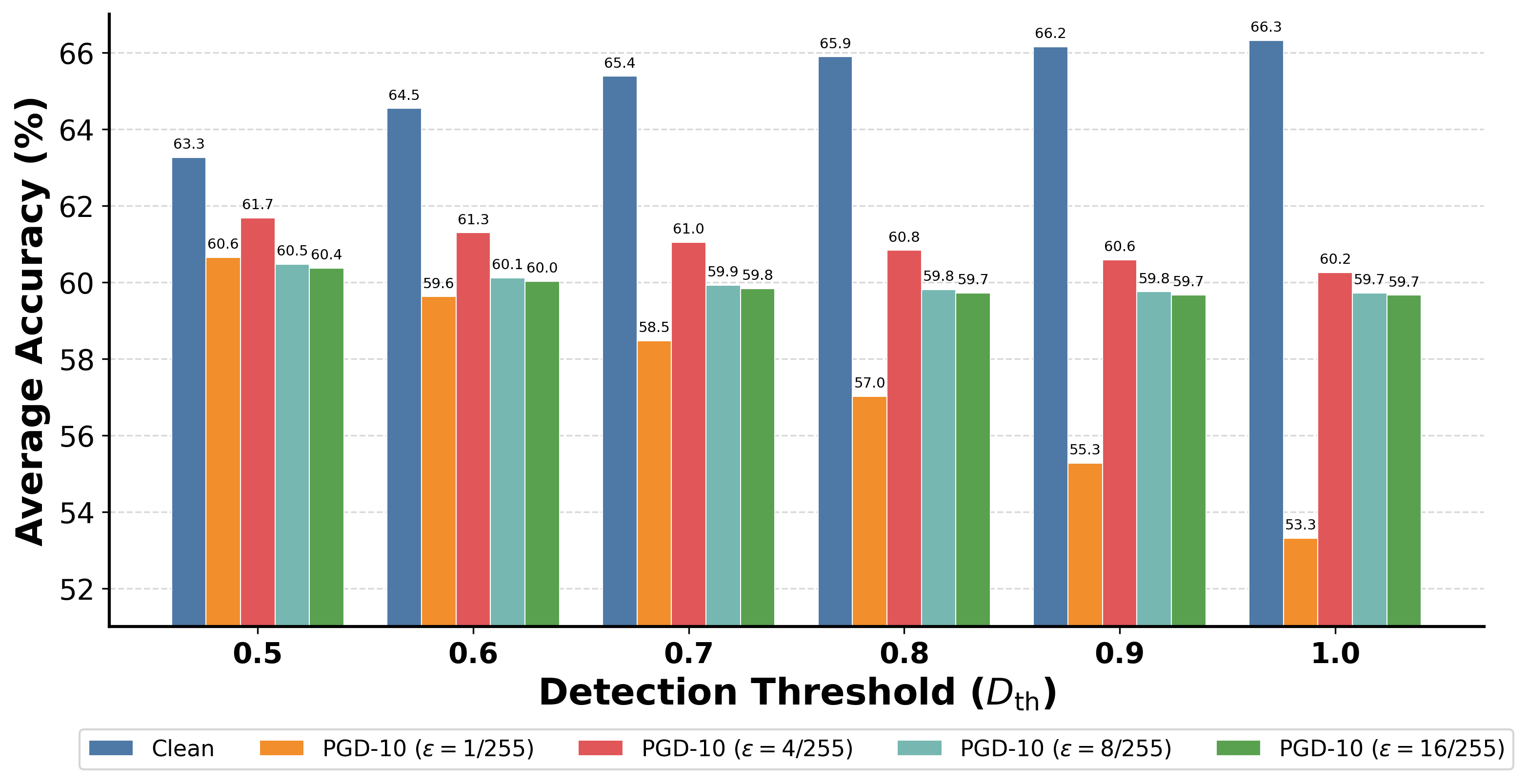}
\caption{\textbf{Sensitivity to the intervention threshold $\tau$
(denoted $D_{\mathrm{th}}$ on the axis).} Clean and PGD-10 robust
accuracy (\%) at $\epsilon\in\{1,4,8,16\}/255$, averaged across 12
downstream datasets, as the threshold is swept from $0.5$ to $1.0$. The
main configuration adopts $\tau{=}0.7$.}
\label{fig:appendix_hp_threshold}
\end{figure}

Figure~\ref{fig:appendix_hp_threshold} extends the threshold sweep
already reported in Table~3 of the main paper (which covers
$\tau\in\{0.6,\ldots,0.9\}$ alongside the drift-only $\tau{=}0.35$
baseline) to the full range $\tau\in\{0.5,\ldots,1.0\}$; the values at
$\tau{=}0.7$ match Table~3's adopted row up to rounding, confirming this
is the same threshold under a different axis label.

Clean accuracy increases monotonically with $\tau$, from $63.3\%$ at
$\tau{=}0.5$ to $66.3\%$ at $\tau{=}1.0$, since a higher threshold
corrects fewer clean inputs unnecessarily. Robust accuracy moves in the
opposite direction, but the size of this effect depends heavily on
attack strength. At $\epsilon{=}1/255$, robust accuracy falls sharply
and consistently as $\tau$ increases, from $60.6\%$ at $\tau{=}0.5$ to
$53.3\%$ at $\tau{=}1.0$, since a stricter threshold increasingly fails
to trigger correction on weakly perturbed inputs whose intervention
score is only marginally above the clean distribution. At
$\epsilon\in\{4,8,16\}/255$, by contrast, robust accuracy declines only
slightly and nearly flattens across the same range, since the
intervention score for inputs at these budgets already clears even the
strictest threshold tested (consistent with the near-$100\%$ detection
rates reported in Table~4 for $\epsilon\geq4/255$). The adopted value
$\tau{=}0.7$ sits at a balanced point that captures most of the
available clean-accuracy gain without yet incurring the steep
weak-attack cost visible at higher thresholds.

\subsection{Number of Anchor Views}
\label{sec:appendix_hp_views}

\begin{figure}[!htbp]
\centering
\includegraphics[width=\linewidth]{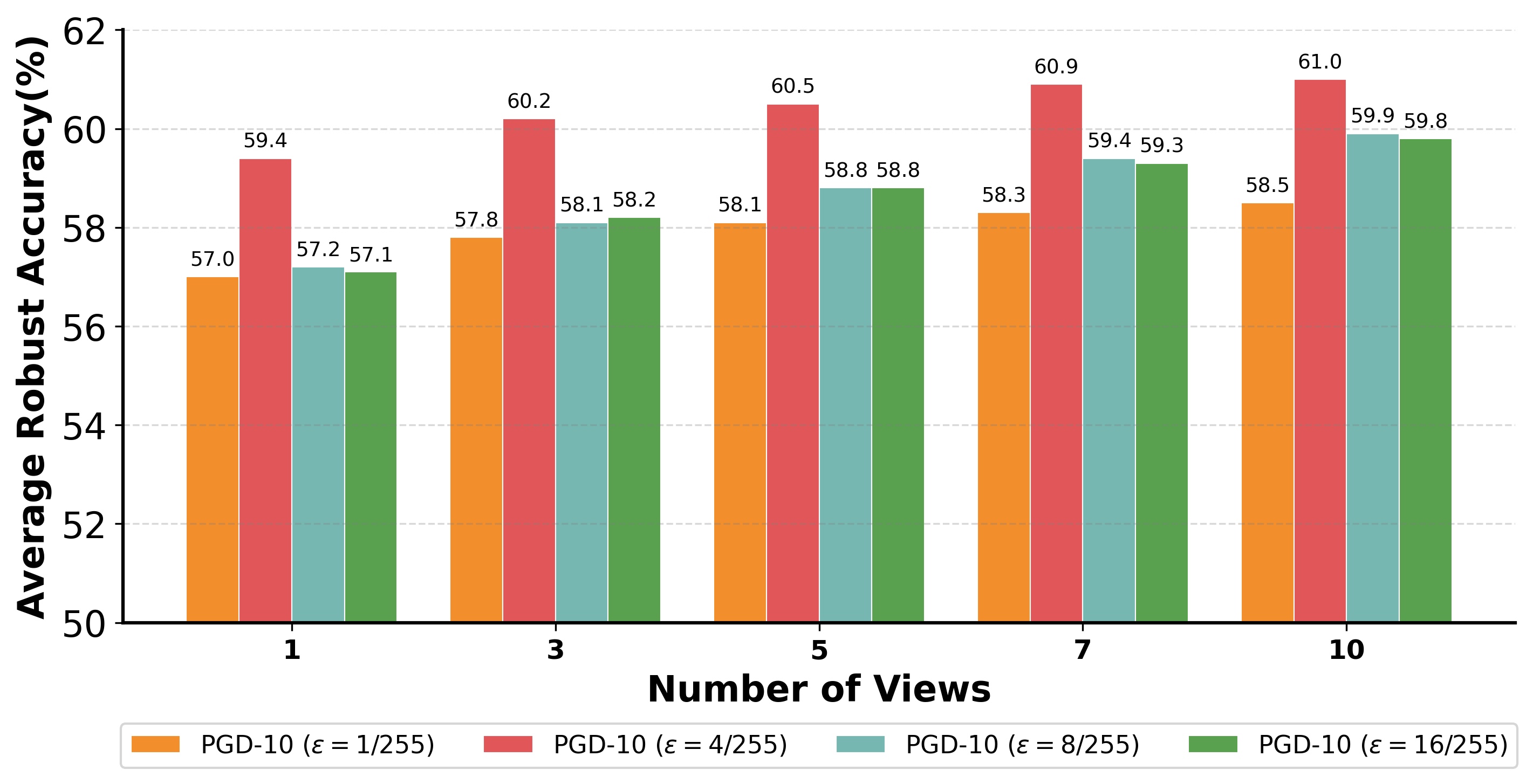}
\caption{\textbf{Sensitivity to the number of anchor views $M$.}
Average PGD-10 robust accuracy (\%) at $\epsilon\in\{1,4,8,16\}/255$,
averaged across 12 downstream datasets, as the number of noise-averaged
views used to construct the anchor (Eq.~\ref{eq:adaptive-anchor}) is
increased from $1$ to $10$. The main configuration adopts $M{=}10$.}
\label{fig:appendix_hp_views}
\end{figure}

Figure~\ref{fig:appendix_hp_views} sweeps $M$ from $1$ to $10$. Robust
accuracy increases monotonically with $M$ at every attack budget, which
is the expected behavior of averaging over more Gaussian-noised views:
a larger $M$ reduces the variance of the anchor estimate
$f_{\mathrm{anc}}(x)$ around its expectation, yielding a more reliable
correction target. The gains are not uniform across this range, however:
moving from $M{=}1$ to $M{=}3$ already recovers most of the improvement
at every budget, while moving from $M{=}7$ to $M{=}10$ adds only a small
further gain. Since each additional view costs one additional forward
pass through the frozen visual encoder
(Appendix~\ref{sec:appendix-algorithm-cost}), this diminishing-returns
pattern motivates $M{=}10$ as a reasonable point that captures nearly
all of the achievable benefit of anchor averaging without incurring
substantially more inference cost for a vanishing additional gain.

\subsection{Extrapolation Strength}
\label{sec:appendix_hp_alpha}

\begin{figure}[!htbp]
\centering
\includegraphics[width=\linewidth]{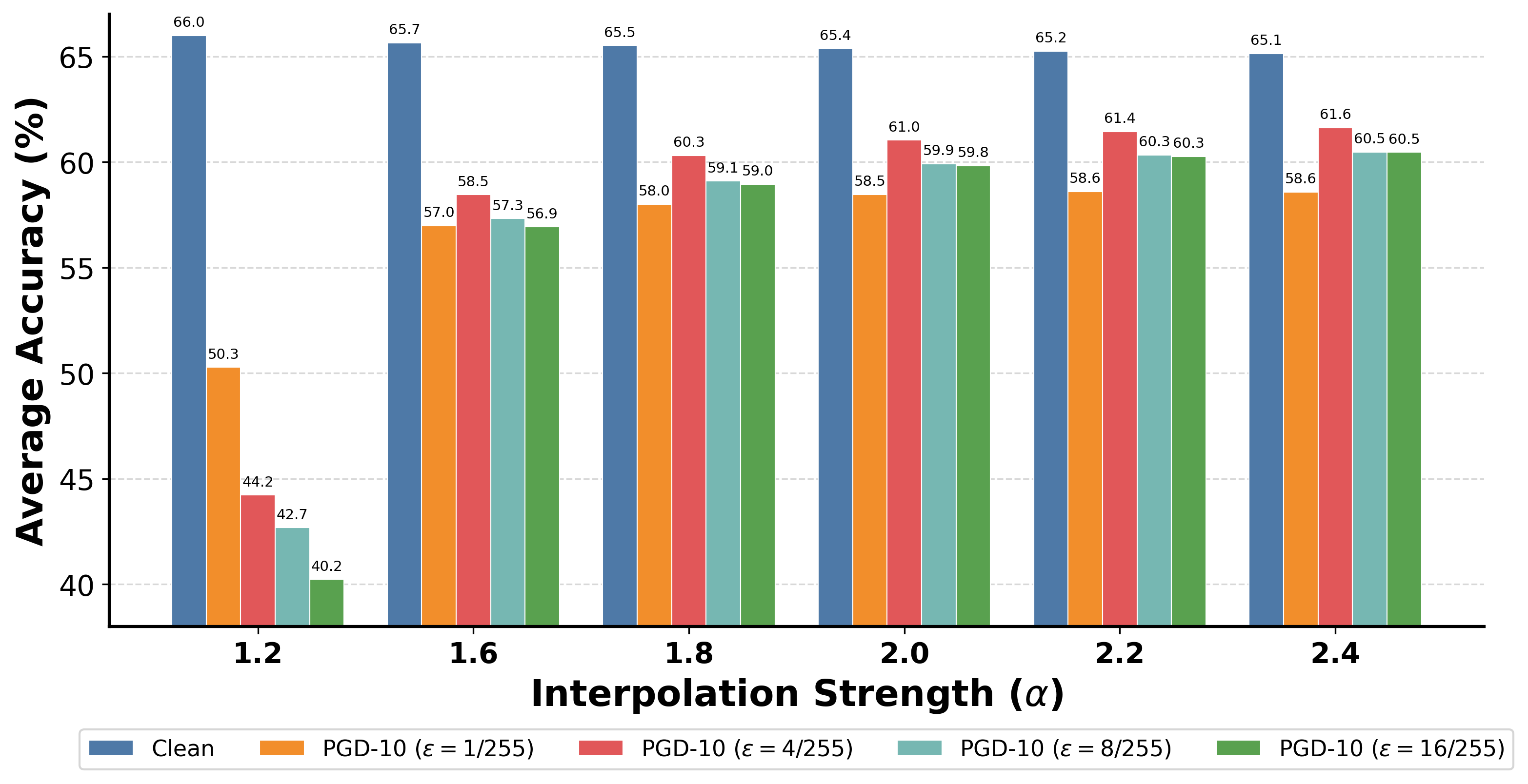}
\caption{\textbf{Sensitivity to the extrapolation strength $\alpha$.}
Clean and PGD-10 robust accuracy (\%) at $\epsilon\in\{1,4,8,16\}/255$,
averaged across 12 downstream datasets, as $\alpha$ is increased from
$1.2$ — the fixed value used by AOM and Defend-CLIP — to $2.4$. The main
configuration adopts $\alpha{=}2.0$.}
\label{fig:appendix_hp_alpha}
\end{figure}

Figure~\ref{fig:appendix_hp_alpha} sweeps $\alpha$ from $1.2$, the exact
extrapolation strength used by AOM and Defend-CLIP in their published
configurations, up to $2.4$. At $\alpha{=}1.2$, robust accuracy is
severely limited, and the shortfall grows with attack strength: from
$50.3\%$ at $\epsilon{=}1/255$ down to just $40.2\%$ at
$\epsilon{=}16/255$, nearly $20$ percentage points below the accuracy
\method{} achieves at this budget under its adopted configuration. This
directly illustrates the limitation motivating \method{}: a fixed, weak
extrapolation strength cannot reach far enough to counteract strongly
displaced inputs. As $\alpha$ increases, robust accuracy rises sharply
across all budgets, with the largest gains at higher $\epsilon$, and
begins to plateau from around $\alpha\approx2.0$ onward; clean accuracy
declines only gradually over the same range, from $66.0\%$ at
$\alpha{=}1.2$ to $65.1\%$ at $\alpha{=}2.4$. The adopted value
$\alpha{=}2.0$ sits at this plateau, capturing nearly all of the
available robustness gain while keeping the clean-accuracy cost small.
Critically, \method{} can operate reliably at this substantially larger
$\alpha$ specifically because $\sigma_{\mathrm{corr}}(x)$ constructs an
anchor matched to each input's own correction demand: extrapolating this
far with a single globally fixed anchor, as in AOM and Defend-CLIP,
would risk overshooting clean and weakly attacked inputs, whereas the
response-conditioned anchor keeps this risk contained even at
$\alpha{=}2.0$.

\section{Analysis of Relative Cross-Noise Drift}
\label{sec:appendix_cross_noise_drift}

Section~\ref{sec:response-policy}~\emph{(main paper)} introduces the relative cross-noise
drift $r(x)$ (Eq.~\ref{eq:relative-drift}) as a graded, sample-specific
signal of correction demand, and Figure~\ref{fig:drift-signal}~\emph{(main paper)} illustrates
its behavior on a single representative dataset. This appendix extends
that analysis across the full evaluation suite, examining whether $r(x)$
consistently separates clean inputs from weakly and strongly attacked
inputs on every individual dataset, rather than only in aggregate, and
whether this behavior is preserved when the attack is optimized for
substantially more iterations and evaluated beyond the calibration range.

\begin{figure*}[!htbp]
\centering
\includegraphics[width=\textwidth]{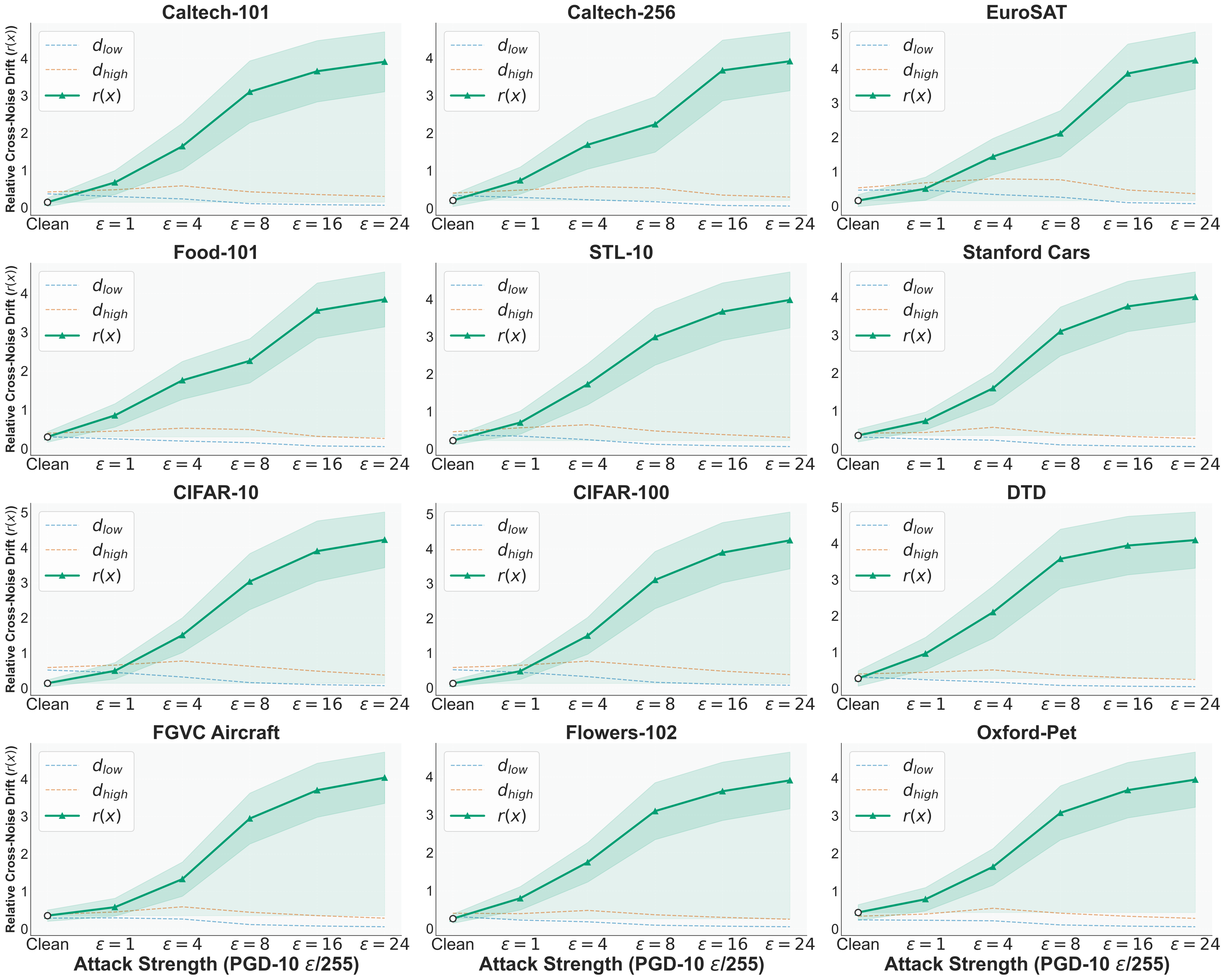}
\caption{\textbf{Relative cross-noise drift across the 12 downstream
datasets under PGD-10.}}
\label{fig:appendix_drift_pgd10_12datasets}
\end{figure*}

\begin{figure*}[!htbp]
\centering
\includegraphics[width=\textwidth]{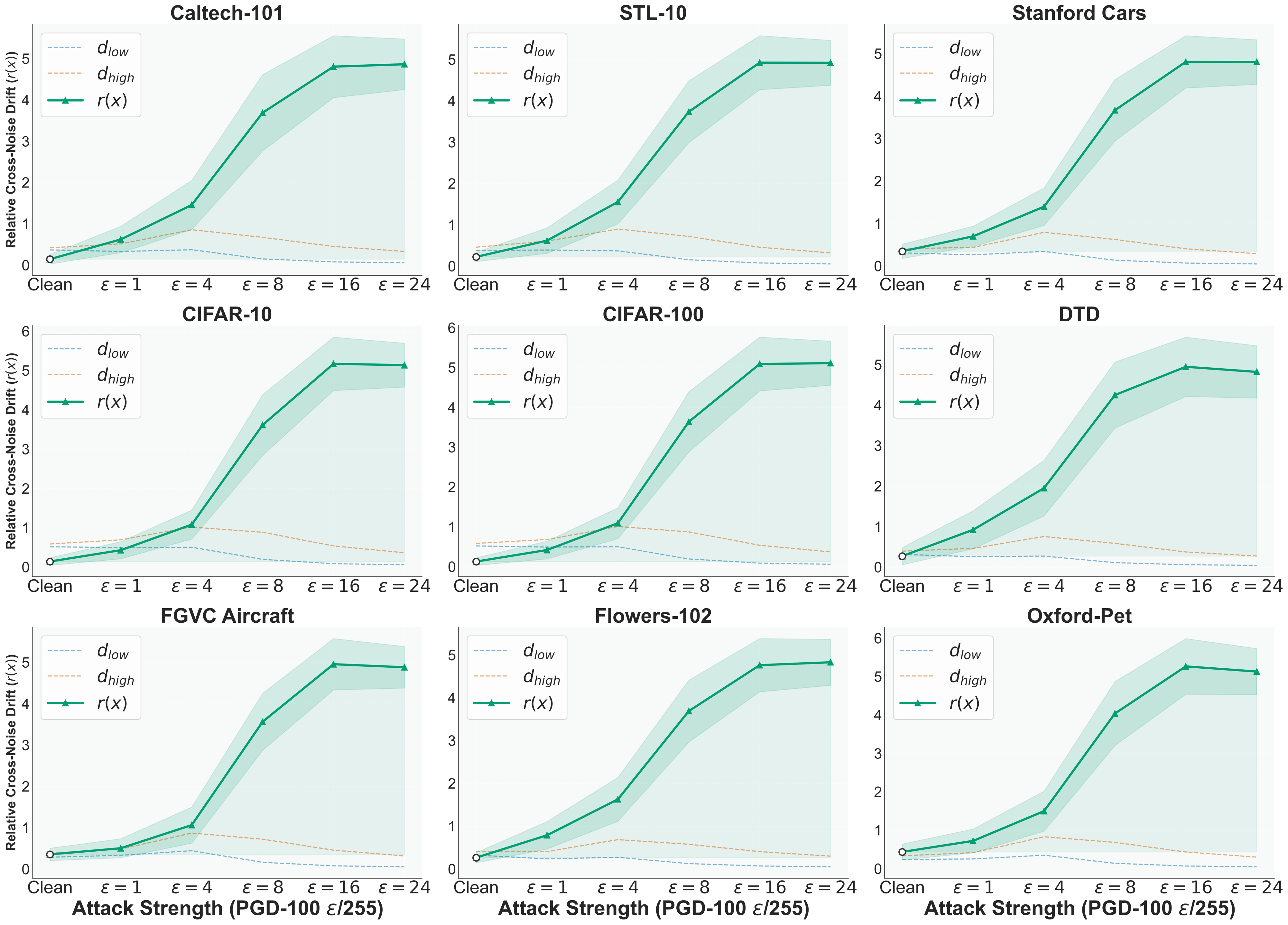}
\caption{\textbf{Relative cross-noise drift across 9 downstream datasets
under PGD-100.}}
\label{fig:appendix_drift_pgd100_9datasets}
\end{figure*}

Figure~\ref{fig:appendix_drift_pgd10_12datasets} reports $d_{\mathrm{low}}(x)$,
$d_{\mathrm{high}}(x)$, and $r(x)$ separately for each of the 12
downstream datasets under PGD-10. The pattern already established for a
single dataset in Figure~\ref{fig:drift-signal}~\emph{(main paper)} holds, without exception,
across every dataset in the suite: the relative cross-noise drift $r(x)$
increases monotonically from clean inputs through $\epsilon=24/255$,
whereas the absolute drifts $d_{\mathrm{low}}(x)$ and $d_{\mathrm{high}}(x)$
individually remain comparatively flat, plateau, or even decline as the
attack strengthens. In most datasets, $d_{\mathrm{low}}(x)$ decreases as
$\epsilon$ increases beyond the weakest budgets, consistent with the
false-stability behavior discussed in the main text: under weak noise
probing, strongly attacked inputs can drift no more, or even less, than
clean inputs. Because $r(x)$ measures the change relative to this
weak-noise baseline, it amplifies exactly this effect rather than being
obscured by it, which is why $r(x)$ orders attack strength cleanly on
every dataset even where $d_{\mathrm{low}}(x)$ and $d_{\mathrm{high}}(x)$
individually do not.

The shaded band around $r(x)$, indicating variability across samples
generated at the same nominal budget, widens consistently with $\epsilon$
in every panel. This confirms at the per-dataset level the same behavior
underlying the design of \method{}: a fixed perturbation budget does not
displace every sample by the same amount in feature space, so a
sample-specific correction scale, rather than one tied only to the nominal
$\epsilon$, is needed even within a single attack budget. The datasets
differ somewhat in how quickly $r(x)$ saturates, with some domains (e.g.,
DTD, EuroSAT) rising sharply by $\epsilon=8/255$ and others (e.g.,
Caltech-101, STL-10) continuing to increase through $\epsilon=16/255$
before flattening, but the qualitative monotone trend and the growth in
variance are shared across all 12 datasets regardless of this difference
in saturation point. Notably, $r(x)$ continues to increase, or at least
does not collapse, from $\epsilon=16/255$ to $\epsilon=24/255$ on every
dataset, which is consistent with the sustained robust accuracy of
\method{} at this extended budget reported in
Table~\ref{tab:main_table1_results_with_eps24_and_eps32}: because the
linear mapping (Eq.~\ref{eq:linear-map}~\emph{(main paper)}) has no predefined upper cap, a
correction-demand signal that keeps rising past the calibration range
continues to be translated into a correspondingly larger correction scale.

Figure~\ref{fig:appendix_drift_pgd100_9datasets} repeats this analysis
under PGD-100 on the 9-dataset subset used in Table~5~\emph{(main paper)}, examining whether
the same behavior holds when the attack is optimized for substantially
more iterations. The qualitative pattern is unchanged: $r(x)$ increases
monotonically with $\epsilon$ on every dataset, while $d_{\mathrm{low}}(x)$
and $d_{\mathrm{high}}(x)$ remain comparatively uninformative individually,
and the variance of $r(x)$ again grows with attack strength. Comparing the
two figures at matched nominal budgets, $r(x)$ reaches somewhat larger
values under PGD-100 than under PGD-10 on the datasets common to both
grids. This indicates that $r(x)$ reflects the effective strength of the
attack achieved by the optimizer, rather than only the nominal
perturbation budget $\epsilon$, which is consistent with $r(x)$ being
used in the main text as a proxy for correction demand rather than as a
direct estimate of $\epsilon$ itself.

Taken together, these per-dataset results support the design choice of
calibrating the linear correction-strength mapping (Eq.~\ref{eq:linear-map}~\emph{(main paper)})
once, on a single dataset, and holding it fixed thereafter: although the
absolute scale and saturation point of $r(x)$ vary somewhat across
domains, the qualitative relationship between $r(x)$ and attack severity
is consistent across every dataset in the suite, across two different
adversarial optimizers, and beyond the calibration range itself,
providing a stable basis for a single, dataset-agnostic mapping.

\section{Analysis of Correction-Scale Distributions}
\label{sec:appendix_correction_scale_distributions}

Section~\ref{sec:response-policy}~\emph{(main paper)} converts the relative cross-noise drift
$r(x)$ into a sample-specific correction scale $\sigma_{\mathrm{corr}}(x)$
using a calibrated linear mapping (Eq.~\ref{eq:linear-map}), and
Figure~\ref{fig:drift-calibration}~\emph{(main paper)} additionally shows logistic and
power-law alternatives fit to the same four calibration points. This
appendix examines the resulting correction-scale distributions in detail.
We first compare the three mappings averaged across datasets, under
PGD-10 on the 12 downstream datasets, under PGD-100 on the 9-dataset
subset used in Table~5~\emph{(main paper)}, and under PGD-10 on the ImageNet-scale suite, to
examine whether the aggregate mapping behavior already summarized in
Figure~\ref{fig:sigma_mapping_ablation}~\emph{(main paper)} is consistent across attack
optimizers and evaluation scale. We then break this comparison down to the
level of individual datasets, to verify that this behavior is not an
artifact of dataset averaging.

\begin{figure*}[!htbp]
\centering
\includegraphics[width=0.8\textwidth]{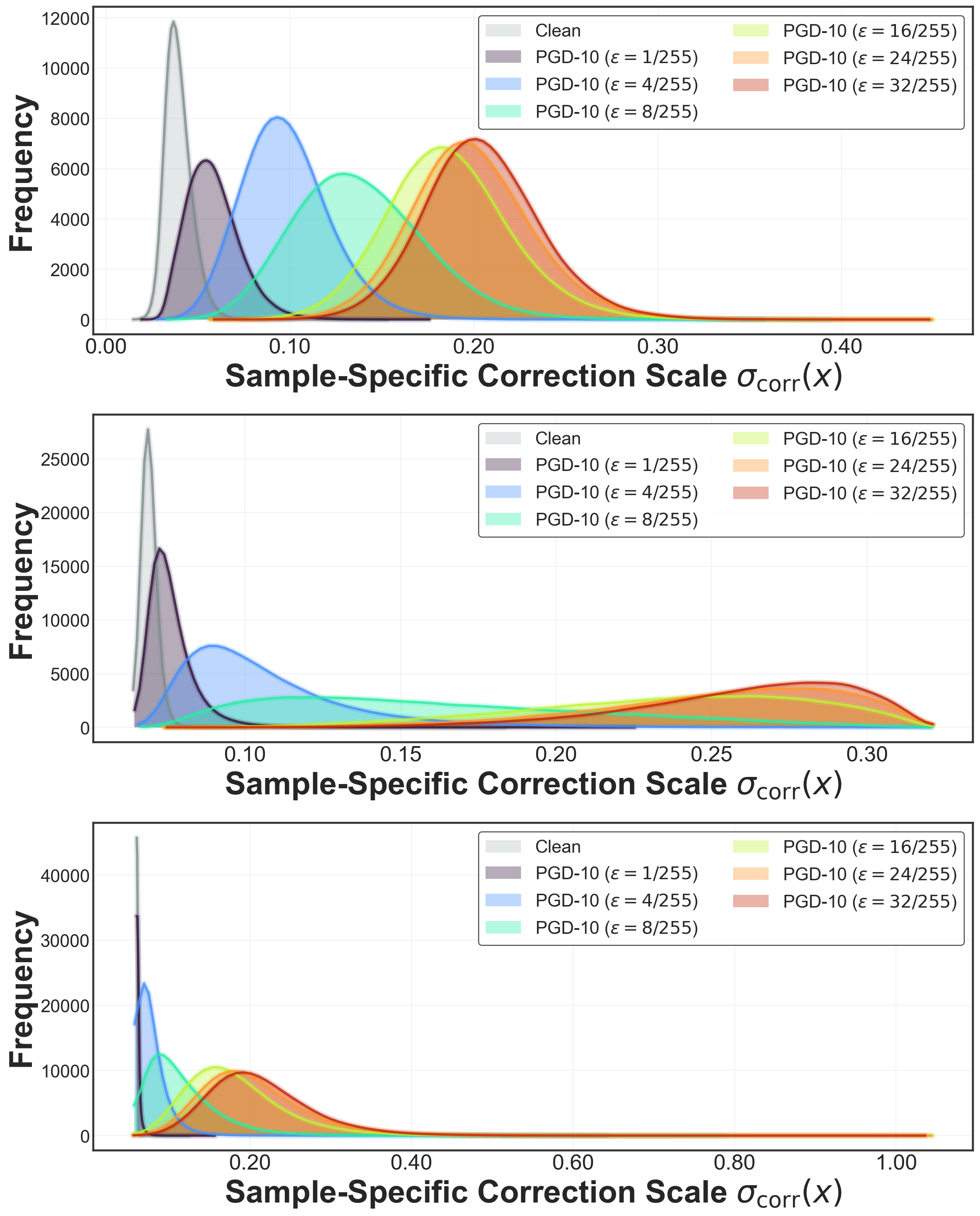}
\caption{\textbf{Comparison of correction-strength mappings under PGD-10,
averaged across 12 datasets.} Correction-scale distributions produced by
the linear (top), logistic (middle), and power-law (bottom) mappings, at
$\epsilon\in\{1,4,8,16,24,32\}/255$, averaged across the 12 downstream
datasets.}
\label{fig:appendix_correction_combined_pgd10_12datasets}
\end{figure*}

\begin{figure*}[!htbp]
\centering
\includegraphics[width=0.9\textwidth]{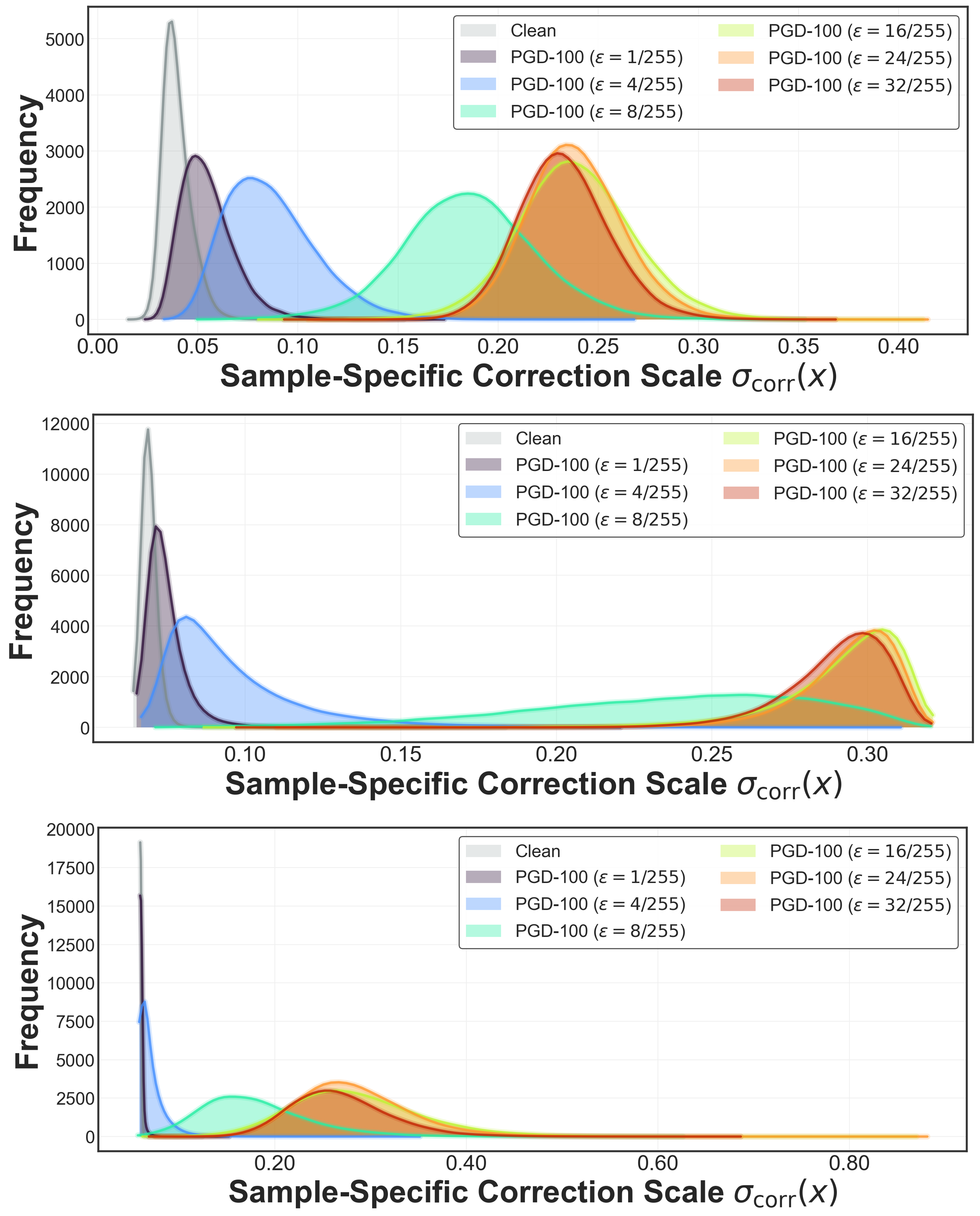}
\caption{\textbf{Comparison of correction-strength mappings under
PGD-100, averaged across 9 datasets.} Correction-scale distributions
produced by the linear (top), logistic (middle), and power-law (bottom)
mappings, at $\epsilon\in\{1,4,8,16,24,32\}/255$, averaged across the same
9-dataset subset used in Table~5 of the main paper.}
\label{fig:appendix_correction_combined_pgd100_9datasets}
\end{figure*}

\begin{figure*}[!htbp]
\centering
\includegraphics[width=\textwidth]{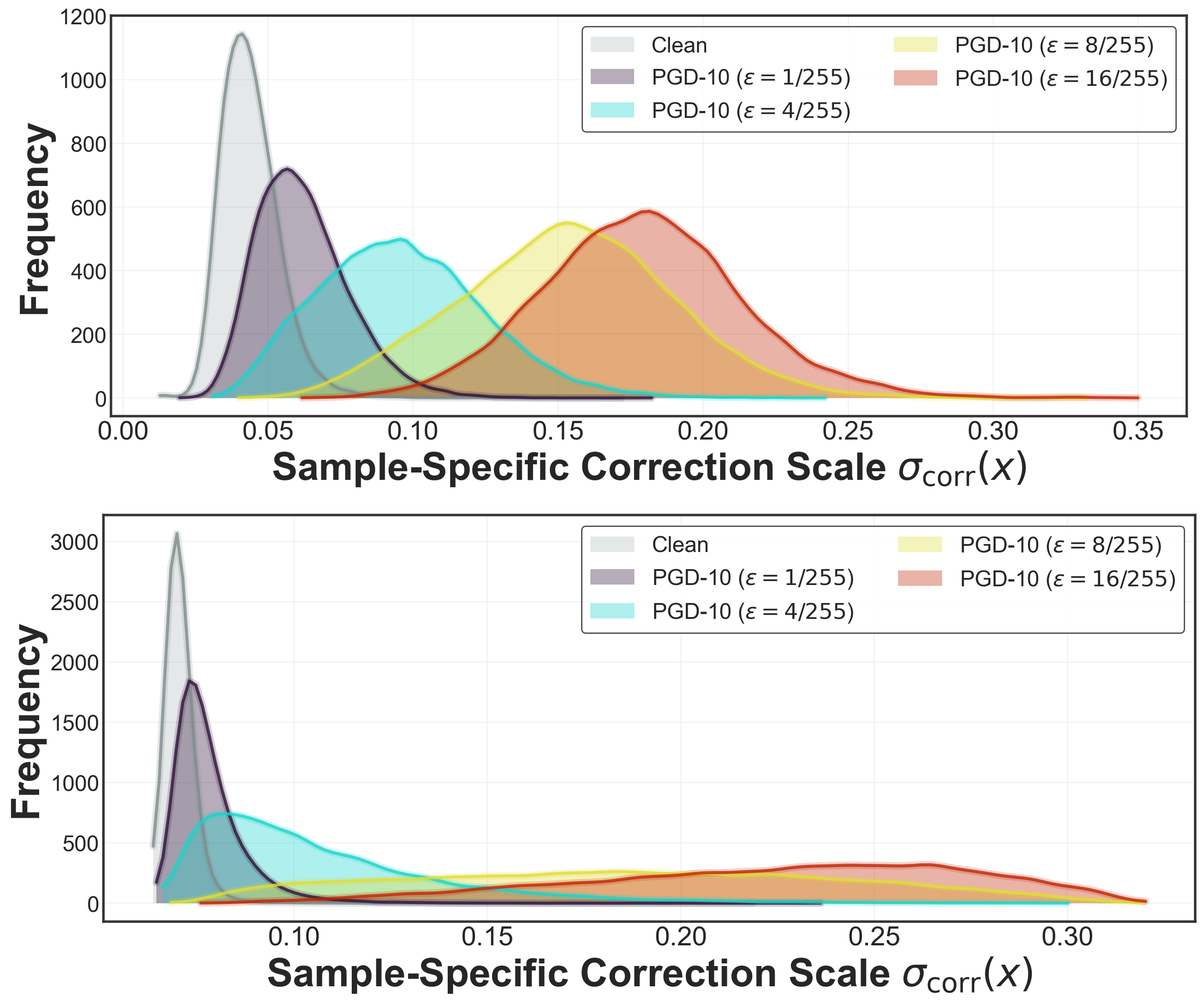}
\caption{\textbf{Comparison of correction-strength mappings on
ImageNet-scale data under PGD-10.} Correction-scale distributions
produced by the linear (top) and logistic (bottom) mappings, at
$\epsilon\in\{1,4,8,16\}/255$, averaged across ImageNet and its four
distribution-shifted variants (ImageNet-A, ImageNet-V2, ImageNet-R,
ImageNet-Sketch).}
\label{fig:appendix_correction_combined_imagenet}
\end{figure*}

Figure~\ref{fig:appendix_correction_combined_pgd10_12datasets} compares
the correction-scale distributions produced by the linear, logistic, and
power-law mappings under PGD-10, averaged across the 12 downstream
datasets. Under all three mappings, the distribution of
$\sigma_{\mathrm{corr}}(x)$ shifts toward larger values as $\epsilon$
increases, with clean inputs consistently receiving the smallest scales.
Each distribution also has non-trivial width, showing that samples
generated under the same nominal budget are assigned a range of different
correction scales rather than a single value, consistent with the
sample-to-sample variation in $r(x)$ noted in
Appendix~\ref{sec:appendix_cross_noise_drift}. The three mappings differ
mainly in how sharply they separate budgets: the linear mapping (top)
shifts smoothly and monotonically across the full range with no
predefined upper cap, so separation continues gracefully even at
$\epsilon\in\{24,32\}/255$; the logistic mapping (middle) compresses
clean and weakly attacked inputs into much sharper peaks but saturates
toward its upper asymptote, causing $\epsilon\in\{8,16,24,32\}/255$ to
collapse into heavily overlapping distributions; and the power-law
mapping (bottom) shows the same pattern in a more extreme form, with the
sharpest low-end compression and the widest high-end spread of the three.
Despite these differences, all three mappings achieve similarly strong
robust accuracy on average (Figure~\ref{fig:sigma_mapping_ablation}~\emph{(main paper)}). The
linear mapping is adopted for the main results because its unbounded,
graduated growth avoids the saturation among moderate-to-strong attacks
seen under the logistic mapping and the very heavy tail of the power-law
mapping.

Figure~\ref{fig:appendix_correction_combined_pgd100_9datasets} repeats
this comparison under PGD-100 on the 9-dataset subset used in Table~5~\emph{(main paper)}.
The same adaptive behavior holds: correction scales shift with attack
strength and vary within each budget under all three mappings, and the
linear mapping again separates budgets most gracefully. Two differences
from PGD-10 stand out. First, at matched nominal $\epsilon$, correction
scales are shifted higher under PGD-100 than PGD-10 across all three
mappings, mirroring the larger $r(x)$ values already observed under
PGD-100 in Appendix~\ref{sec:appendix_cross_noise_drift} and confirming
that this directly propagates into a larger assigned correction scale via
Eq.~\ref{eq:linear-map}. Second, the clustering among
$\epsilon\in\{16,24,32\}/255$ is more pronounced, with these three budgets
nearly overlapping under every mapping; this does not come at a cost to
robustness, since \method{} sustains strong accuracy across this range
under PGD-100 (Table~\ref{tab:main_table5_results_per_dataset}) — a
saturated but still large correction scale remains sufficient to counter
these strongly displaced inputs.

Figure~\ref{fig:appendix_correction_combined_imagenet} extends the
comparison to ImageNet-scale data under PGD-10, using the linear and
logistic mappings at $\epsilon\in\{1,4,8,16\}/255$. The same qualitative
behavior is preserved: correction scales shift rightward with attack
strength, vary within each budget, and separate more gracefully under the
linear mapping than under the logistic mapping, which again compresses
clean and weak attacks into sharp peaks and saturates at higher budgets.

Across all three settings, these results confirm that \method{}'s
correction strength adapts consistently at two levels: across attack
budgets, as reflected in the progressive rightward shift of each
distribution, and within a single budget, as reflected in each
distribution's spread — and that this adaptive behavior, along with the
rationale for adopting the unbounded linear mapping, holds across
attack optimizers and evaluation scale.

\begin{figure*}[!htbp]
\centering
\includegraphics[width=\textwidth]{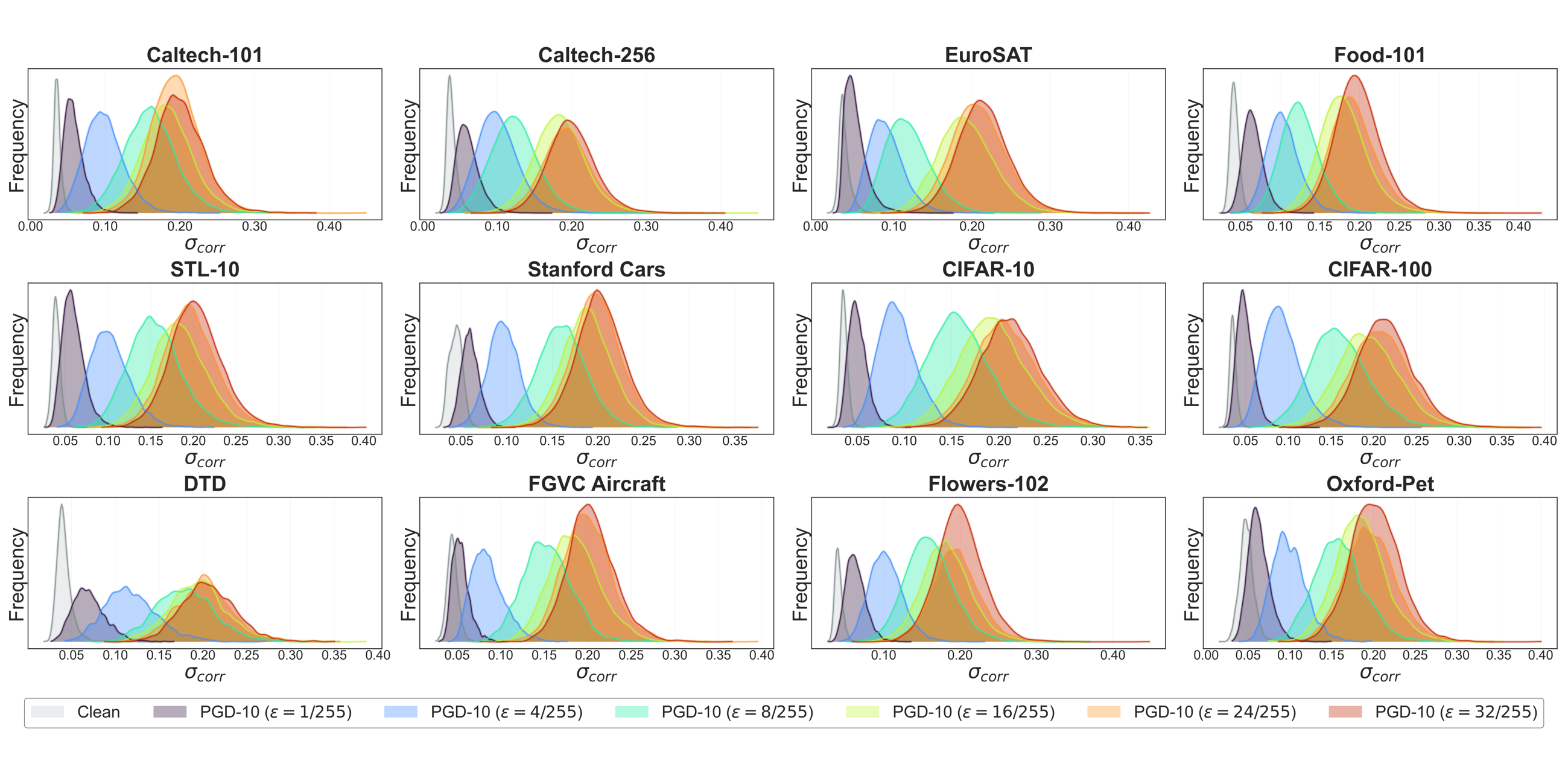}
\caption{\textbf{Per-dataset correction-scale distributions under the
linear mapping.} Distribution of the sample-specific correction scale
$\sigma_{\mathrm{corr}}(x)$ produced by the adopted linear mapping
(Eq.~\ref{eq:linear-map}), for each of the 12 downstream datasets under
PGD-10 attacks at $\epsilon\in\{1,4,8,16,24,32\}/255$.}
\label{fig:appendix_correction_linear_pgd10}
\end{figure*}

\begin{figure*}[!htbp]
\centering
\includegraphics[width=\textwidth]{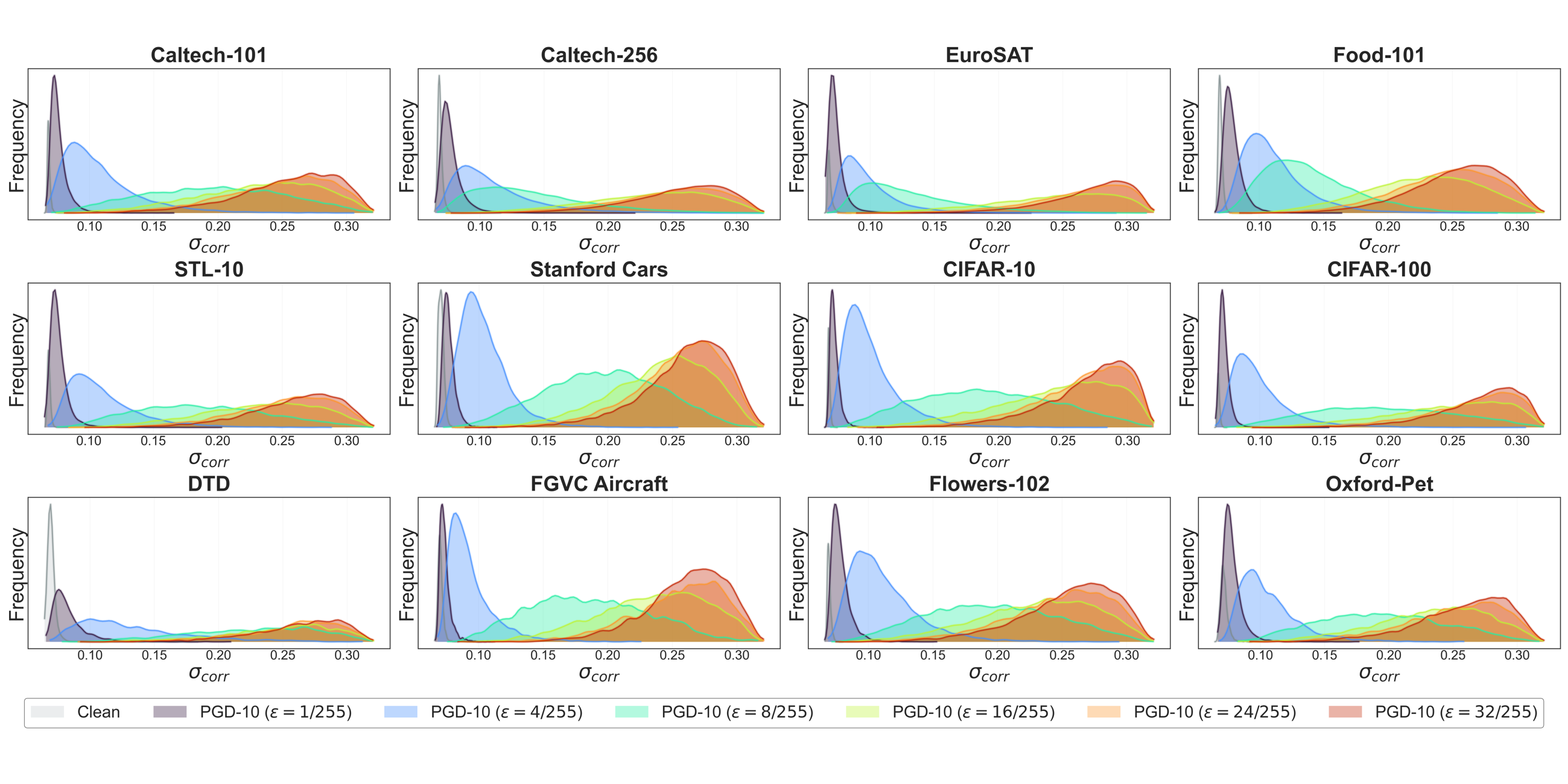}
\caption{\textbf{Per-dataset correction-scale distributions under the
logistic mapping.} Distribution of $\sigma_{\mathrm{corr}}(x)$ produced by
the logistic alternative mapping, for each of the 12 downstream datasets
under PGD-10 attacks at $\epsilon\in\{1,4,8,16,24,32\}/255$.}
\label{fig:appendix_correction_logistic_pgd10}
\end{figure*}

\begin{figure*}[!htbp]
\centering
\includegraphics[width=\textwidth]{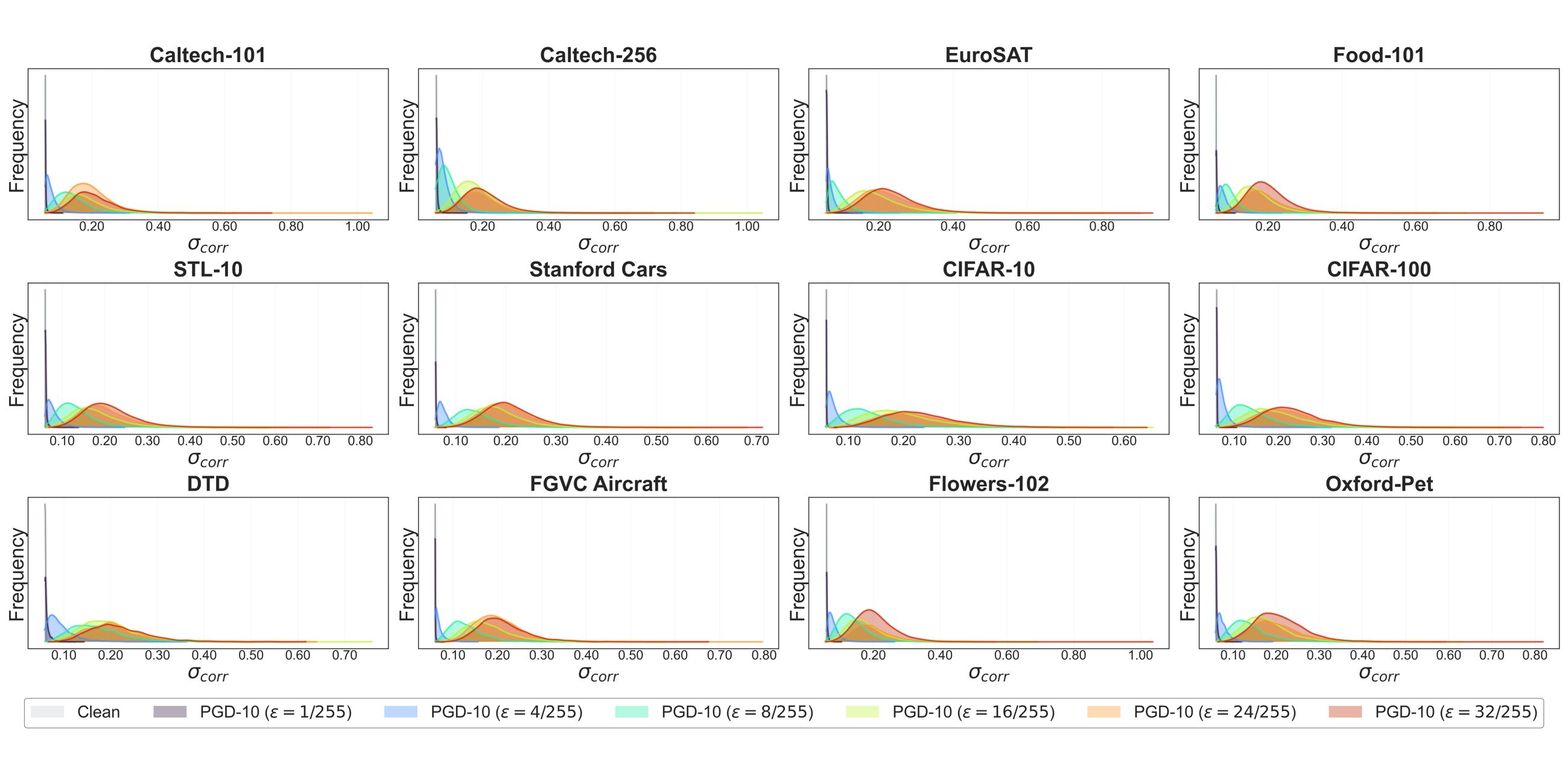}
\caption{\textbf{Per-dataset correction-scale distributions under the
power-law mapping.} Distribution of $\sigma_{\mathrm{corr}}(x)$ produced by
the power-law alternative mapping, for each of the 12 downstream datasets
under PGD-10 attacks at $\epsilon\in\{1,4,8,16,24,32\}/255$.}
\label{fig:appendix_correction_power_pgd10}
\end{figure*}

Figures~\ref{fig:appendix_correction_linear_pgd10},
\ref{fig:appendix_correction_logistic_pgd10}, and
\ref{fig:appendix_correction_power_pgd10} report the per-dataset
correction-scale distributions underlying the PGD-10, 12-dataset
comparison in Figure~\ref{fig:appendix_correction_combined_pgd10_12datasets},
for the linear, logistic, and power-law mappings respectively, for readers
interested in the dataset-level detail.

\section{Analysis of Defensive Intervention}
\label{sec:appendix_defensive_intervention}

Section~\ref{sec:response-policy}~\emph{(main paper)} introduces the defensive intervention
score $s_{\mathrm{int}}(x)=r(x)+J(x)$ (Eq.~\ref{eq:intervention-score}~\emph{(main paper)}),
which fuses the relative cross-noise drift $r(x)$ with the
prediction-instability score $J(x)$ to decide whether correction should be
applied to a given input, and Figure~\ref{fig:fused-gate}~\emph{(main paper)} illustrates the
complementary roles of the two signals on an aggregate basis. This
appendix examines the same three quantities, $r(x)$, $J(x)$, and
$s_{\mathrm{int}}(x)$, at the level of individual datasets, to verify that
this complementary behavior, and the resulting improvement in separating
clean from weakly attacked inputs, holds consistently across the 12
downstream datasets rather than only in aggregate.

\begin{figure*}[!htbp]
\centering
\includegraphics[width=\textwidth]{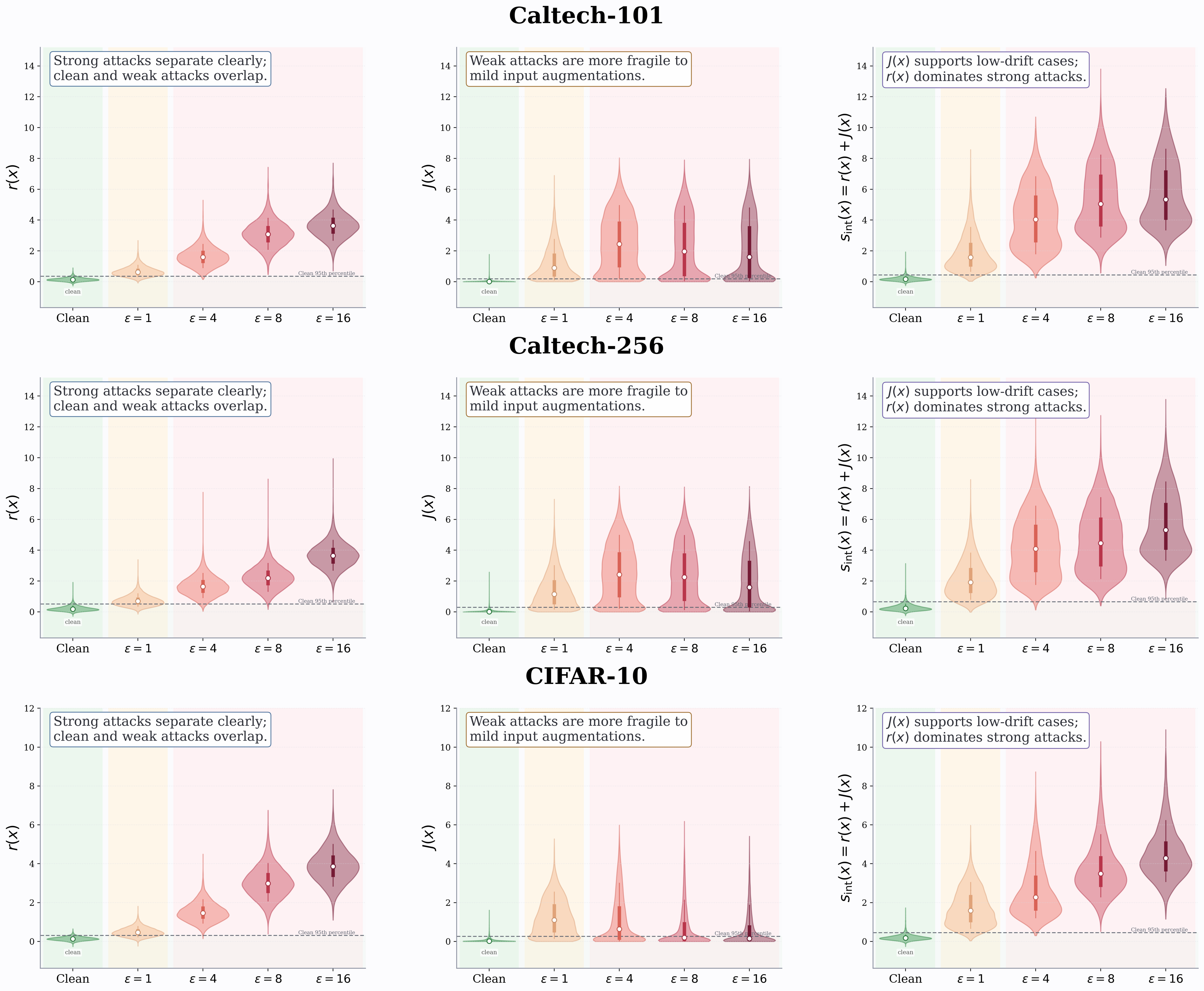}
\caption{\textbf{Per-dataset intervention-metric distributions (Part 1 of
4): Caltech101, Caltech256, CIFAR10.} Violin plots of $r(x)$ (left),
$J(x)$ (middle), and $s_{\mathrm{int}}(x)=r(x)+J(x)$ (right) for clean
inputs and PGD-10 adversarial inputs at $\epsilon\in\{1,4,8,16\}/255$.}
\label{fig:appendix_violin_part1}
\end{figure*}

\begin{figure*}[!htbp]
\centering
\includegraphics[width=\textwidth]{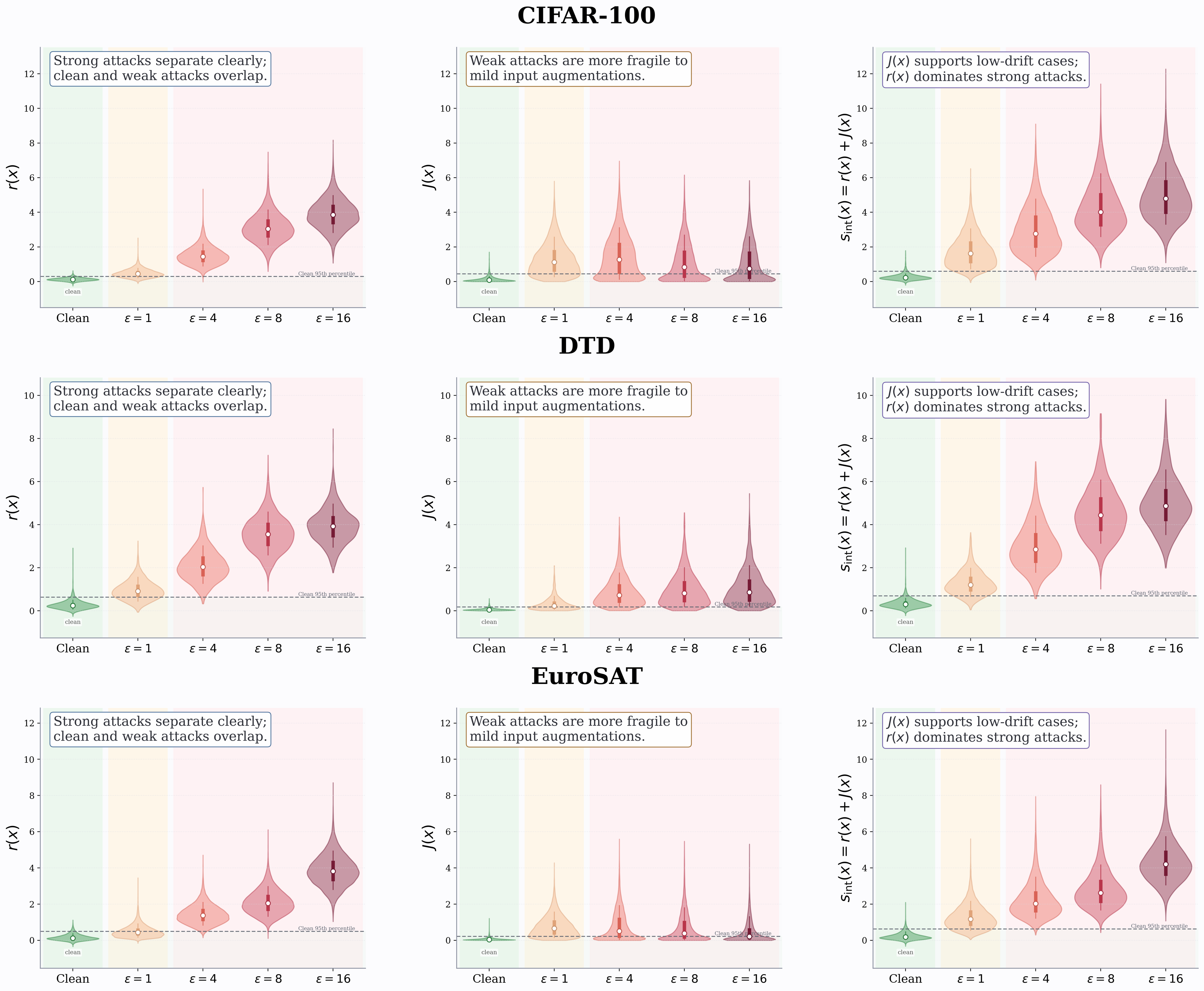}
\caption{\textbf{Per-dataset intervention-metric distributions (Part 2 of
4): CIFAR100, DTD, EuroSAT.} Violin plots of $r(x)$ (left), $J(x)$
(middle), and $s_{\mathrm{int}}(x)=r(x)+J(x)$ (right) for clean inputs and
PGD-10 adversarial inputs at $\epsilon\in\{1,4,8,16\}/255$.}
\label{fig:appendix_violin_part2}
\end{figure*}

\begin{figure*}[!htbp]
\centering
\includegraphics[width=\textwidth]{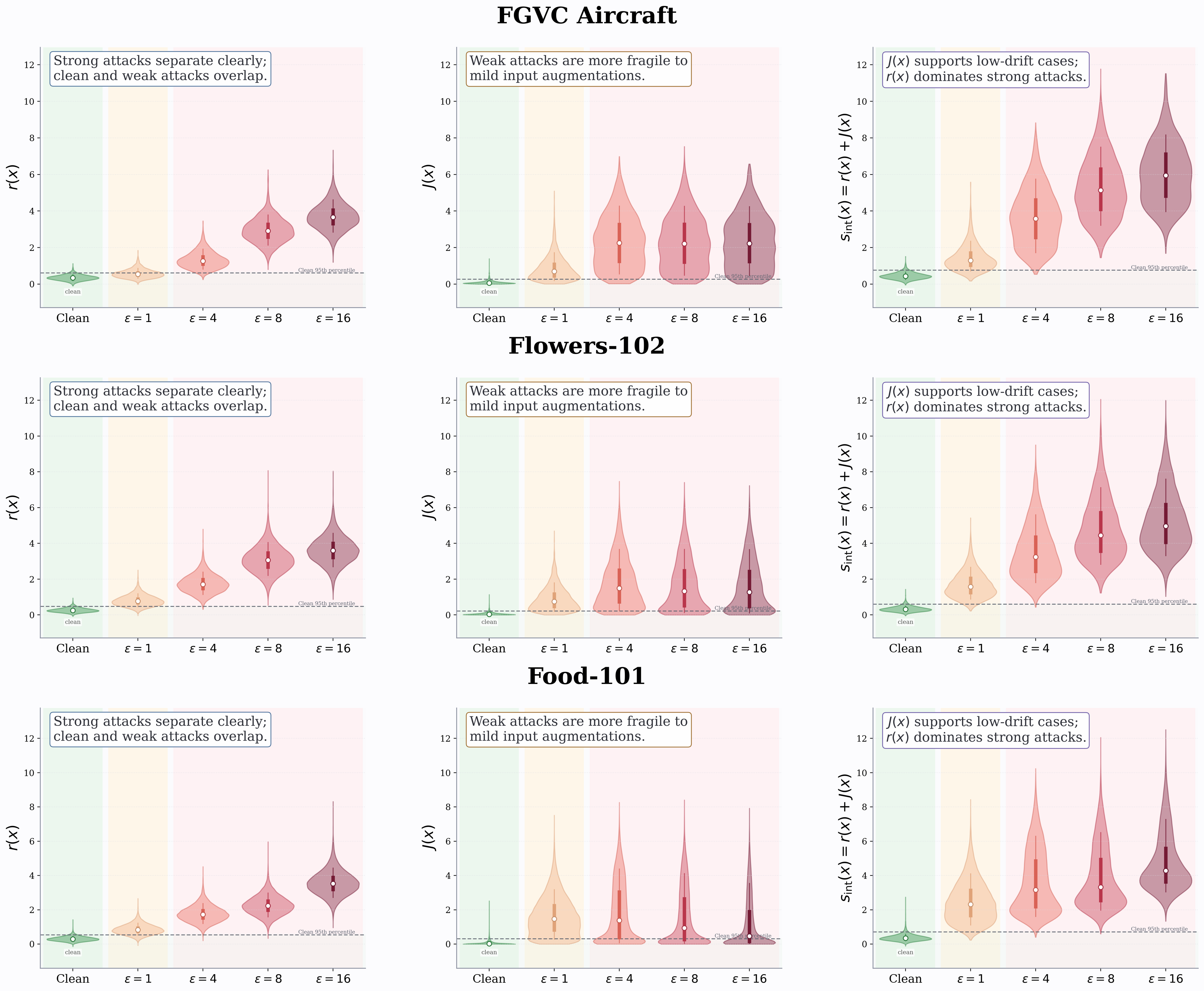}
\caption{\textbf{Per-dataset intervention-metric distributions (Part 3 of
4): FGVC-Aircraft, Flowers102, Food101.} Violin plots of $r(x)$ (left),
$J(x)$ (middle), and $s_{\mathrm{int}}(x)=r(x)+J(x)$ (right) for clean
inputs and PGD-10 adversarial inputs at $\epsilon\in\{1,4,8,16\}/255$.}
\label{fig:appendix_violin_part3}
\end{figure*}

\begin{figure*}[!htbp]
\centering
\includegraphics[width=\textwidth]{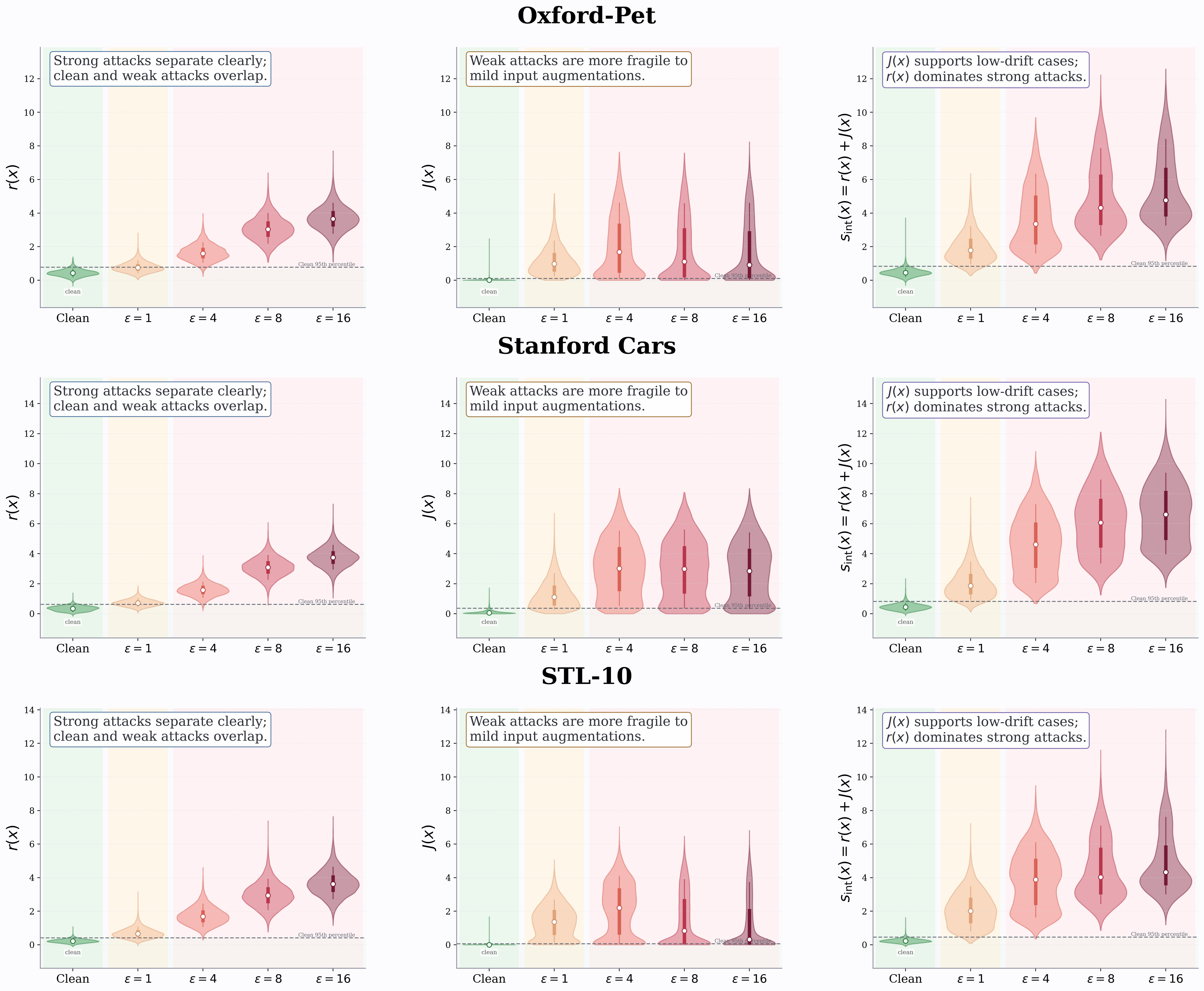}
\caption{\textbf{Per-dataset intervention-metric distributions (Part 4 of
4): OxfordPets, StanfordCars, STL10.} Violin plots of $r(x)$ (left),
$J(x)$ (middle), and $s_{\mathrm{int}}(x)=r(x)+J(x)$ (right) for clean
inputs and PGD-10 adversarial inputs at $\epsilon\in\{1,4,8,16\}/255$.}
\label{fig:appendix_violin_part4}
\end{figure*}

Figures~\ref{fig:appendix_violin_part1}--\ref{fig:appendix_violin_part4}
report violin plots of the three metrics for clean inputs and PGD-10
adversarial inputs at $\epsilon\in\{1,4,8,16\}/255$, split across four
figures of three datasets each. In each panel, the violin width reflects
sample density, the internal markers summarize central tendency and
spread, and a reference line marks the 95th percentile of the clean
distribution to visualize separation from adversarial inputs.

Figures~\ref{fig:appendix_violin_part1}--\ref{fig:appendix_violin_part4}
report the per-dataset distributions of $r(x)$, $J(x)$, and
$s_{\mathrm{int}}(x)$ for all 12 downstream datasets. The pattern
underlying Figure~\ref{fig:fused-gate} in the main text holds
consistently across every dataset. For $r(x)$ (left column), the clean
distribution sits tightly near zero, but the $\epsilon=1/255$ distribution
overlaps substantially with the clean 95th-percentile reference in every
dataset; separation only becomes clear from $\epsilon=4/255$ onward, with
both the median and spread of $r(x)$ growing steadily through
$\epsilon=16/255$. For $J(x)$ (middle column), the clean distribution is
sharply concentrated at zero, and even $\epsilon=1/255$ already separates
visibly above the clean reference in every dataset, exactly where $r(x)$
still overlaps. Notably, $J(x)$ is not monotone in attack strength: on
most datasets its median peaks around $\epsilon\in\{1,4\}/255$ and then
plateaus or decreases at $\epsilon\in\{8,16\}/255$, most visibly on DTD
and EuroSAT. This non-monotonicity is precisely why $J(x)$ is used to
complement rather than replace $r(x)$: it is most informative in the
weak-attack regime that $r(x)$ struggles to separate, while $r(x)$ takes
over as the dominant signal as the attack strengthens. The combined score
$s_{\mathrm{int}}(x)$ (right column) inherits both strengths on every
dataset: it separates from the clean reference already at
$\epsilon=1/255$, where $J(x)$ contributes most, and continues to
separate further through $\epsilon=16/255$, where the growth of $r(x)$
dominates. This consistent, per-dataset complementarity confirms that the
aggregate improvement from fusing $r(x)$ and $J(x)$ (Tables~2--4~\emph{(main paper)}) reflects
a general property of the two signals rather than behavior specific to a
subset of datasets.

\section{Adversarial Threat Scope}
\label{sec:appendix-threat-scope}

Section~\ref{sec:experiments}~\emph{(main paper)} states the adversarial threat setting in
brief. This appendix makes that setting fully explicit, along the three
axes that define any adversarial evaluation — the attacker's \emph{goal},
\emph{knowledge}, and \emph{capability} — and clarifies exactly which
quantities are, and are not, assumed known to the attacker when
\method{} is evaluated.

\subsection{Attacker Goal, Knowledge, and Capability}
\label{sec:appendix-threat-formal}

The attacker's goal is untargeted misclassification of the frozen
zero-shot CLIP classifier: given a clean image $x$ with label $y$, the
attacker seeks an $\ell_p$-bounded perturbation $\delta^\star$ that
maximizes the classification loss (Eq.~\ref{eq:attack}~\emph{(main paper)}), producing
$x_{\mathrm{adv}} = x + \delta^\star$. Throughout this paper, $p=\infty$
and the objective is optimized with PGD~\citep{madry2017towards}.

The attacker is \textbf{white-box with respect to the frozen encoder}:
it has full access to the weights and gradients of the visual and text
encoders $F_v, F_t$ used for zero-shot classification (Eqs.~1--2), and
can backpropagate through the similarity and softmax computation to
construct $\delta^\star$. This is the established threat model used
throughout the training-free test-time defense literature for CLIP that
we compare against — TTC~\citep{xing2025clip},
AOM~\citep{tong2025zero}, Defend-CLIP~\citep{brahma2026defending},
DOC~\citep{jiang2026diversifying}, MAC~\citep{kim2026mac}, and
R-TPT~\citep{sheng2025r} all adopt the same white-box-against-the-frozen-encoder
setting in their evaluations, and our protocol is designed to be directly
comparable to theirs rather than introducing a new one.

Within this shared setting, the attacker is \textbf{oblivious to the
test-time defense}: $\delta^\star$ is optimized against the undefended
classifier alone, before \method{} is applied, and does not
backpropagate through any part of Algorithm~\ref{alg:react-clip} — not
the noise probes, the response-conditioned anchor construction, the
augmentation $\mathcal T$, nor the intervention gate. Concretely, the
attacker has access to the frozen encoder weights, the prompt templates
and class names, and the zero-shot classification pipeline of
Eqs.~1--2, but has no knowledge of whether a test-time defense will be
applied at all, nor of any of its internal quantities: the probe noise
scales $\sigma_{\mathrm{low}}, \sigma_{\mathrm{high}}$, the calibrated
mapping coefficients $(a,b)$, the anchor sample count $M$, the
extrapolation strength $\alpha$, the augmentation operator $\mathcal T$,
or the intervention threshold $\tau$. This is again the same convention
used by TTC, AOM, Defend-CLIP, DOC, MAC, and R-TPT in their reported
results.

A direct consequence, already noted in Sections~\ref{sec:introduction}
and~\ref{sec:response-policy}, is that the defender likewise does not
assume or require knowledge of the attacker's perturbation budget
$\epsilon$: the same fixed hyperparameters
$(\sigma_{\mathrm{low}},\sigma_{\mathrm{high}},a,b,M,\alpha,\tau)$ are
used across every budget and dataset in our evaluation. This symmetry
— neither side observes the other's configuration — is what motivates
treating correction strength as something to be estimated per input
rather than fixed in advance.

\subsection{Attack Suite and Perturbation Budgets}
\label{sec:appendix-threat-suite}

Within this scope, we evaluate four attack objectives, all optimized
against the frozen, undefended CLIP model:

\begin{itemize}
\item \textbf{PGD-10} ($\ell_\infty$, 10 steps): the primary attack used
throughout the main results (Table~\ref{tab:main_results_avg}~\emph{(main paper)}), at
budgets $\epsilon \in \{1,4,8,16\}/255$. This is also the attack used to
generate the four calibration points for the correction-strength
mapping (Section~\ref{sec:response-policy}).
\item \textbf{PGD-100} ($\ell_\infty$, 100 steps): a stronger-optimizer
check at the same four budgets (Table~\ref{tab:pgd100}~\emph{(main paper)}), verifying that
gains are not an artifact of a weak 10-step attacker.
\item \textbf{C\&W-10}~\citep{carlini2017towards}: an attack with a
different loss surface than PGD's cross-entropy objective
(Table~\ref{tab:cw}~\emph{(main paper)}), checking that the correction-demand signal
$r(x)$ generalizes beyond one attack family.
\item \textbf{AutoAttack}: an ensemble attack combining multiple attack
strategies (Table~\ref{tab:autoattack}~\emph{(main paper)}), used as a further check
against a stronger, parameter-free adversary.
\end{itemize}

Beyond the four calibration budgets, Figure~\ref{fig:extended_eps}~\emph{(main paper)}
additionally evaluates PGD-10 at $\epsilon \in \{24,32\}/255$ — outside
the range used to fit $\sigma_{\mathrm{corr}}(x)$ — to check whether the
linear mapping continues to behave reasonably when extrapolated beyond
its calibration range, rather than only within it.

All four attack objectives, at every budget, are optimized under the
same white-box, defense-oblivious scope described in
Section~\ref{sec:appendix-threat-formal}.

\subsection{Evaluation Protocol}
\label{sec:appendix-threat-protocol}

Following standard practice in this evaluation setting, adversarial
examples are crafted only for clean test images that the undefended,
zero-shot CLIP model classifies correctly. Images the undefended model
already misclassifies are left unperturbed and passed through the
defense as-is.

This choice is not a minor bookkeeping detail. The alternative
convention — attacking every test image regardless of whether the
frozen model already misclassified it — produces a reporting artifact
we term \emph{label leakage} (Appendix~\ref{app:label_leakage}), and we
find this artifact to be specific to, and especially severe for,
feature-correction defenses of the kind studied in this paper. Because
an untargeted PGD attack is always optimized against the \emph{true}
label regardless of the model's starting prediction, the resulting
perturbation embeds the true-class direction even on a sample the
frozen model never classified correctly to begin with. Anchor-based
correction, which is built precisely to reverse attack-like
displacements, can then carry such a sample into the correct class for
the first time and be credited with a "restoration" that never
occurred. Neither AOM~\citep{tong2025zero} nor
Defend-CLIP~\citep{brahma2026defending} — the two feature-correction
baselines this paper builds on most directly — corrects for this effect
in their reported results. Appendix~\ref{app:label_leakage} quantifies
the resulting inflation directly on both methods and shows that it can
raise reported robust accuracy above the frozen model's own clean
accuracy, an outcome that is impossible under a correctly-scored
evaluation. We therefore adopt the leakage-free protocol described above
throughout this paper, and re-evaluate every baseline in
Table~\ref{tab:main_results_avg}~\emph{(main paper)} — including AOM and Defend-CLIP —
under this same corrected protocol, so that all comparisons reported in
the main text already reflect genuine robustness rather than this
artifact.

\section{Label Leakage}
\label{app:label_leakage}

Appendix~\ref{sec:appendix-threat-protocol} introduces the leakage-free
evaluation protocol used throughout this paper and names the artifact it
is designed to avoid: \emph{label leakage}, in which a correction-based
defense is credited with "correcting" a sample the frozen model never
classified correctly to begin with. This appendix substantiates that
claim in full. We show that the alternative convention used elsewhere in
this literature — attacking every test sample regardless of whether it
was already misclassified — inflates the reported robustness of
feature-correction defenses specifically, and we quantify this effect on
two representative baselines, AOM~\citep{tong2025zero} and
Defend-CLIP~\citep{brahma2026defending}, neither of which corrects for
it in their own reported evaluations.

\begin{figure}[t]
    \centering
    \includegraphics[width=\linewidth]{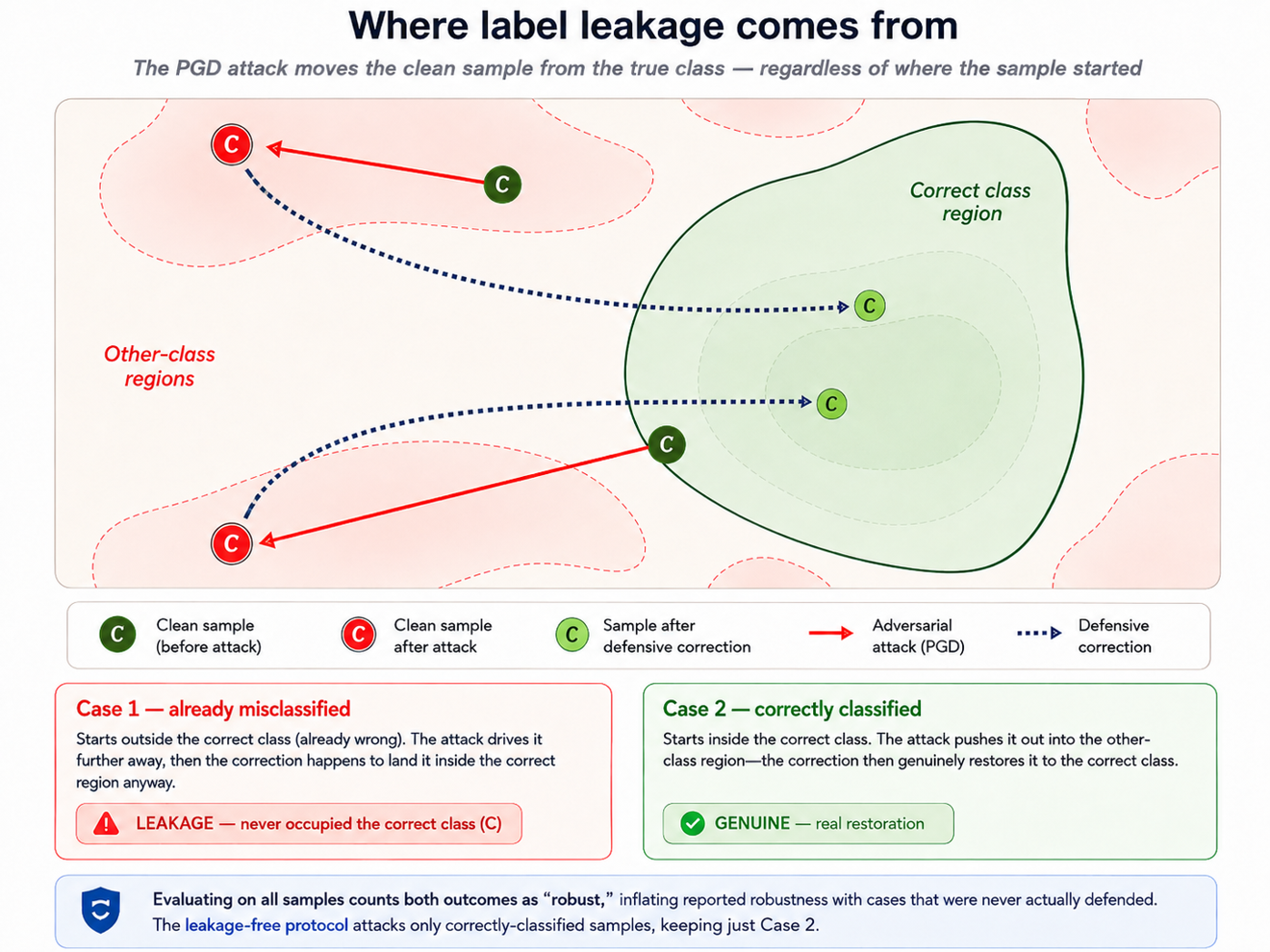}
\caption{\textbf{Where label leakage comes from.} As the attack
is optimized against the \emph{true} label regardless of the sample's
starting point, it displaces the clean feature away from the correct
class in both cases shown, and the correction pulls the attacked
feature back toward it in both cases as well. \textbf{Case 1 (top),
already misclassified:} the clean sample starts outside the correct-class
region. The attack pushes it further away, but the correction still
happens to land it inside the correct-class region — a result that
looks identical to a successful defense, yet the sample never
legitimately occupied that region to begin with. This is
\textbf{label leakage}. \textbf{Case 2 (bottom), correctly classified:}
the clean sample starts inside the correct-class region. The attack
pushes it out into an other-class region, and the correction restores
it to the region it legitimately occupied before attack — a
\textbf{genuine restoration}. Scoring robustness on all samples counts
both cases identically as correct, inflating reported robustness with
cases that were never actually defended; the leakage-free protocol
attacks only samples in Case 2, so only genuine restorations can be
counted as robust.}
\label{fig:app_leakage_concept}
\end{figure}

\paragraph{Setup and attack.}
Let $f_v(x)$ and $f_t^k$ denote the $\ell_2$-normalized visual and text
features (Eq.~\ref{eq:clip-features}), and let $p(y{=}k\mid x)$ and
$\hat y(x)=\arg\max_k p(y{=}k\mid x)$ denote the resulting class
posterior and prediction (Eq.~\ref{eq:clip-prediction}). The attacker
performs untargeted PGD: given the \emph{true} label $y$, it seeks the
$\ell_\infty$-bounded perturbation that maximizes the cross-entropy
loss with respect to $y$ (Eq.~\ref{eq:attack}), approximated by
iterating
\begin{equation}
\label{eq:app_pgd_iterate}
    x^{(t+1)}
    = \Pi_{\mathcal{B}_\infty(x,\epsilon)}\!\left(
    x^{(t)}
    + \eta\,\mathrm{sign}\!\left[
    \nabla_{x^{(t)}}
    \mathcal{L}_{\mathrm{ce}}\big(p(\cdot\mid x^{(t)}),\,y\big)
    \right]\right),
\end{equation}
\noindent where $\Pi_{\mathcal{B}_\infty(x,\epsilon)}$ projects back
onto the $\epsilon$-ball around $x$. The essential point for what
follows: every PGD step is computed from the gradient with respect to
the \emph{true} class $y$, so the perturbation moves the visual feature
$f_v(x)$ away from $y$ along a direction constructed from $y$'s own
loss surface, regardless of whether the model classified $x$ correctly
to begin with.

\paragraph{Noise-anchored feature correction.}
Noise-response defenses~\citep{tong2025zero,brahma2026defending} build
a noise-averaged anchor from $M$ draws $\eta_i\sim\mathcal N(0,I)$ at a
fixed scale $\sigma_k$, and extrapolate the test
feature toward it with strength $\alpha$:
\begin{equation}
\label{eq:app_anchor}
\begin{aligned}
    f_{\mathrm{anc}}(x;\sigma_k) &=\frac{1}{M}\sum_{i=1}^{M} f_v(x+\sigma_k\eta_i), \\
    \tilde f_v(x;\sigma_k,\alpha) &=f_v(x)+\alpha\big[f_{\mathrm{anc}}(x;\sigma_k)-f_v(x)\big].
\end{aligned}
\end{equation}
\noindent The defended prediction $\tilde y(x)=\arg\max_k \tilde
p(y{=}k\mid x)$ scores the normalized corrected feature
$\tilde f_v(x;\sigma_k,\alpha)/\|\tilde f_v(x;\sigma_k,\alpha)\|_2$
against the same text features via Eq.~\ref{eq:clip-prediction}. AOM
applies this operator to every input; Defend-CLIP applies it only to
inputs whose relative-drift statistic crosses a threshold.

\paragraph{Definition of label leakage.}
Let $\mathcal{D}$ be the test set and partition it by the \emph{clean}
prediction of the frozen model:
$\mathcal{D}_{\mathrm{correct}}=\{(x,y):\hat{y}(x)=y\}$ and
$\mathcal{D}_{\mathrm{wrong}}=\mathcal{D}\setminus\mathcal{D}_{\mathrm{correct}}$.
The protocol used by prior work in this family attacks \emph{every}
sample in $\mathcal{D}$ and reports
\begin{equation}
\label{eq:app_reported}
    \mathrm{RobustAcc}_{\mathrm{all}}
    = \frac{1}{|\mathcal{D}|}
    \sum_{(x,y)\in\mathcal{D}}
    \mathbb{1}\big[\tilde{y}(x_{\mathrm{adv}})=y\big].
\end{equation}
\noindent This sum splits into two qualitatively different events:
\begin{equation}
\label{eq:app_split}
\begin{array}{ll}
    \text{True Robustness:} &
    (x,y)\in\mathcal{D}_{\mathrm{correct}}
    \ \wedge\ \tilde{y}(x_{\mathrm{adv}})=y, \\[0.35em]
    \text{Label Leakage:} &
    (x,y)\in\mathcal{D}_{\mathrm{wrong}}
    \ \wedge\ \tilde{y}(x_{\mathrm{adv}})=y.
\end{array}
\end{equation}

\noindent Label leakage credits the defense with a sample the
undefended model never classified correctly. An adversary has no
reason to attack an already-misclassified input, and standard
adversarial-robustness practice evaluates on correctly-classified
samples only~\citep{croce2020reliable},
yet the second event is far from rare here, for a reason specific to
correction-based defenses. Because the untargeted attack in
Eq.~\ref{eq:app_pgd_iterate} ascends the loss with respect to the true
class $y$, the perturbation it produces on a misclassified sample still
embeds the true-class direction: it moves the feature further from $y$
along $y$'s gradient. The interpolation in Eq.~\ref{eq:app_anchor} is
built precisely to reverse attack-like displacements, so applying it
can carry the feature into a class region it never occupied, producing
a "correct" prediction that reflects the geometry of the attack, not
any genuine defense. Figure~\ref{fig:app_leakage_concept} illustrates
both events: Label Leakage corresponds to Case~1 in the figure, where
the sample never occupied the correct class before attack, and True
Robustness corresponds to Case~2, where the correction restores a
sample to a class it legitimately occupied prior to attack. We now
quantify the effect, measuring everything at the same sweep: anchor
scales $\sigma_k\in[0.03,0.28]$, interpolation strengths
$\alpha\in[0,2.4]$, and budgets $\epsilon\in\{1,4,8,16\}/255$.

\begin{figure}[!t]
    \centering
    \includegraphics[width=\linewidth]{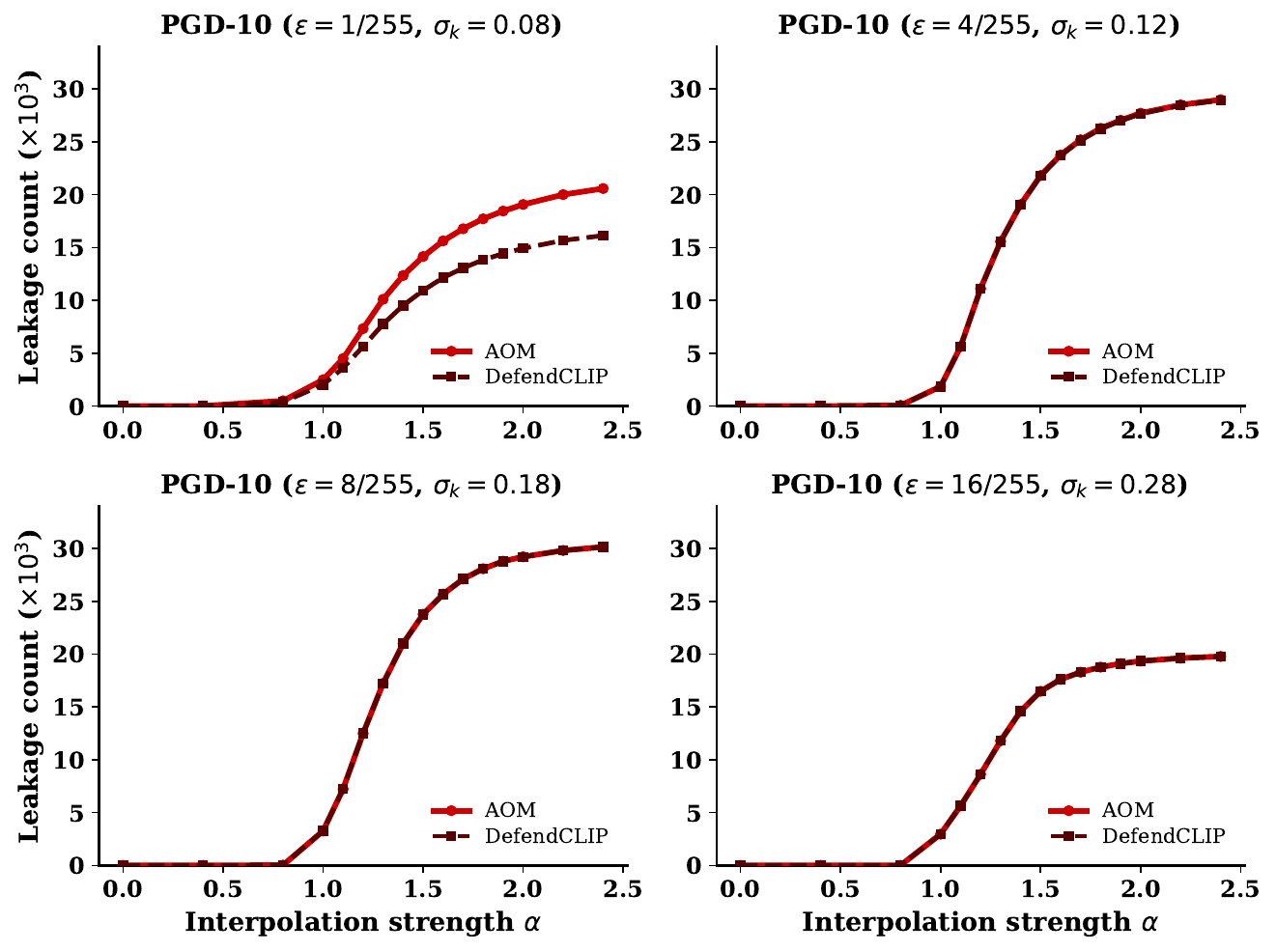}
\caption{\textbf{Leakage counts against interpolation strength $\alpha$},
at each budget's optimal $\sigma_k$, pooled across 12 downstream
datasets.}
\label{fig:leakage_counts_alpha}
\end{figure}

\begin{figure}[!t]
    \centering
    \includegraphics[width=\linewidth]{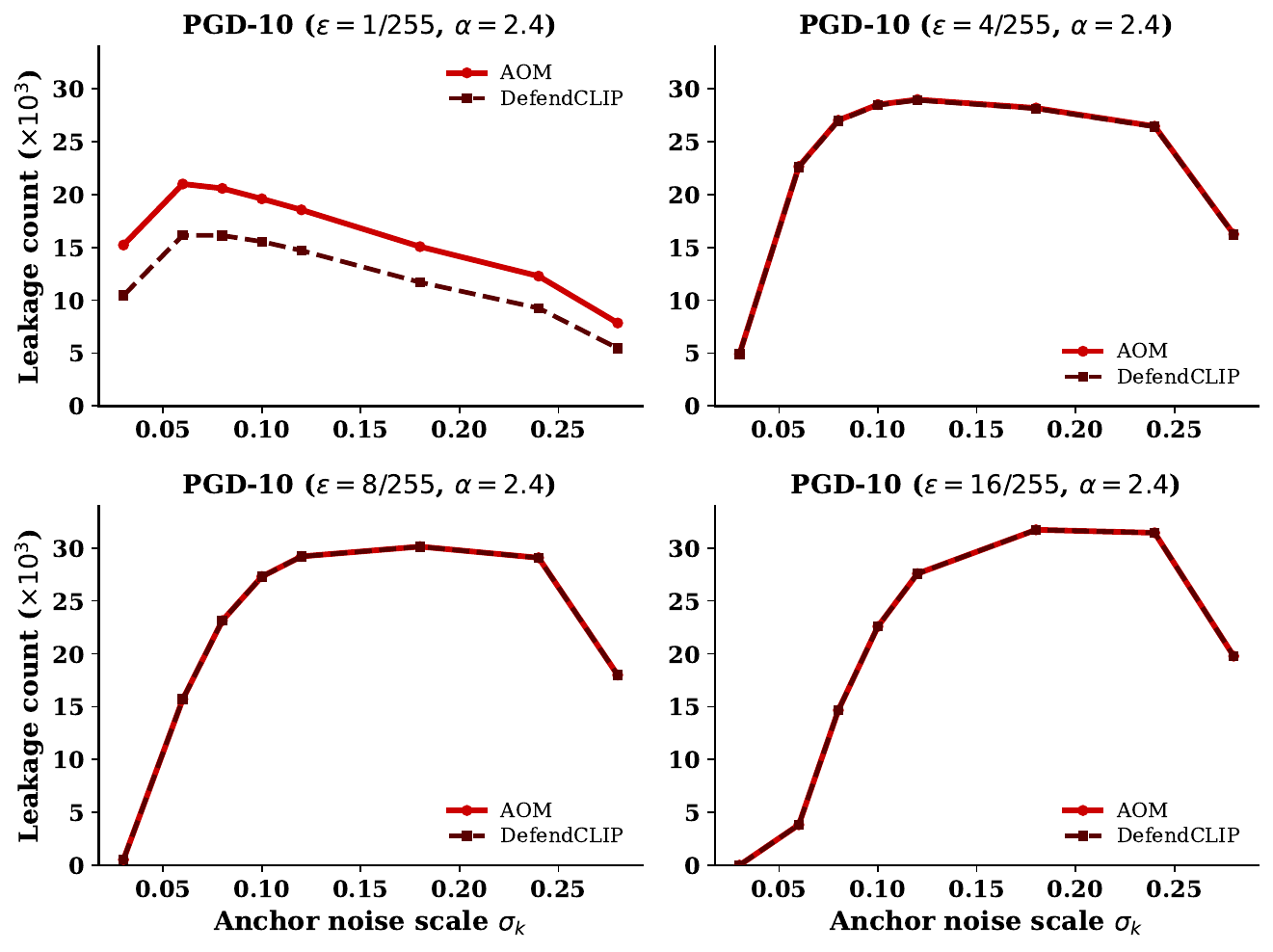}
\caption{\textbf{Leakage counts against the anchor scale $\sigma_k$}, at
$\alpha{=}2.4$, pooled across 12 downstream datasets. }
\label{fig:leakage_counts_sigma}
\end{figure}

\begin{figure}[!t]
    \centering
    \includegraphics[width=\linewidth]{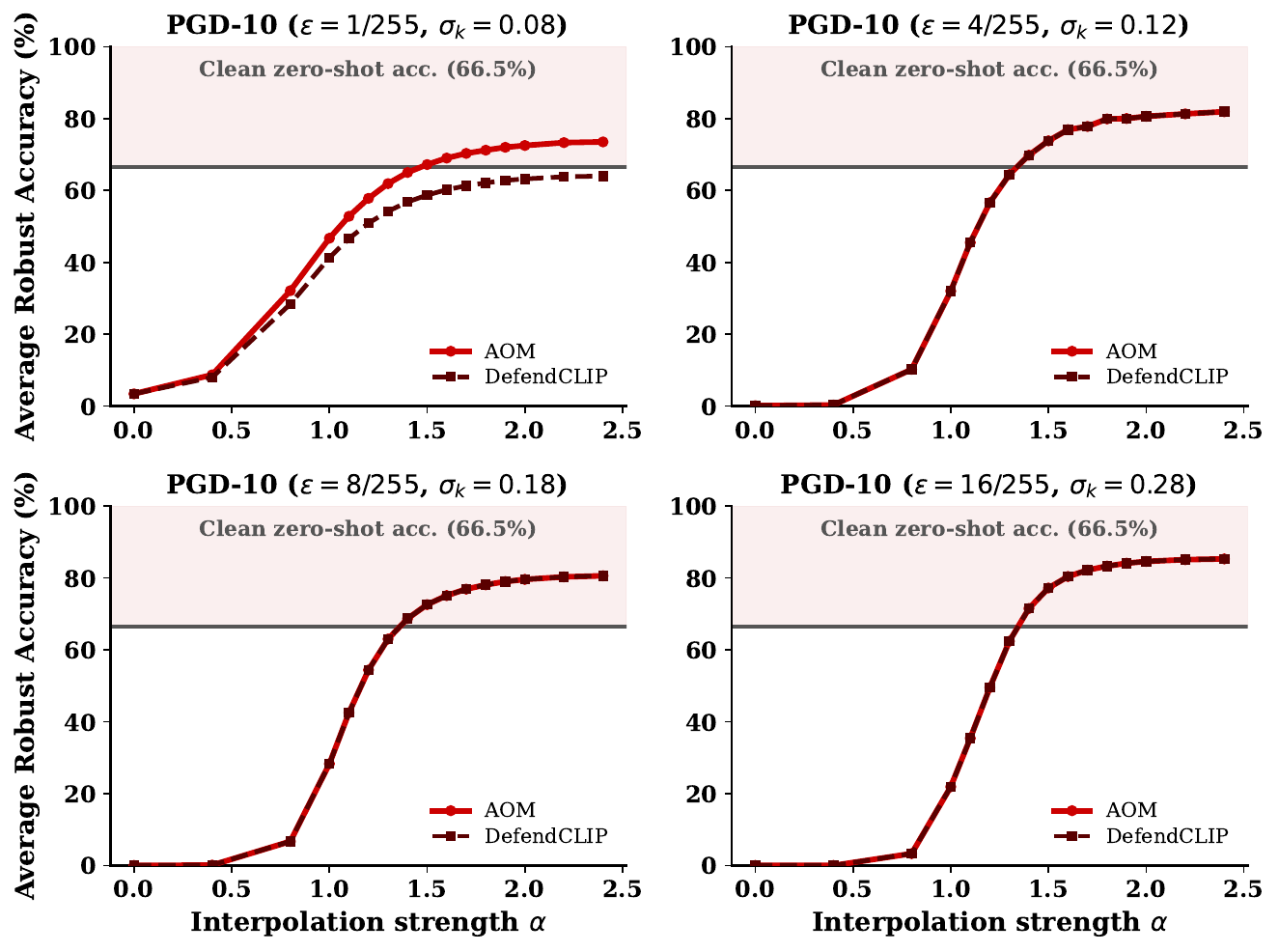}
    \caption{\textbf{Robust accuracy under leakage against $\alpha$} averaged across 12 downstream datasets, at each budget's optimal $\sigma_k$.
    Accuracy rises monotonically through the frozen model's own clean
    accuracy of $66.5\%$ (shaded), crossing it near
    $\alpha\approx1.4$--$1.5$ at every budget, so a sweep under this
    protocol always selects the largest $\alpha$ tested.}
    \label{fig:leakage_reported_alpha}
\end{figure}

\begin{figure}[!t]
    \centering
    \includegraphics[width=\linewidth]{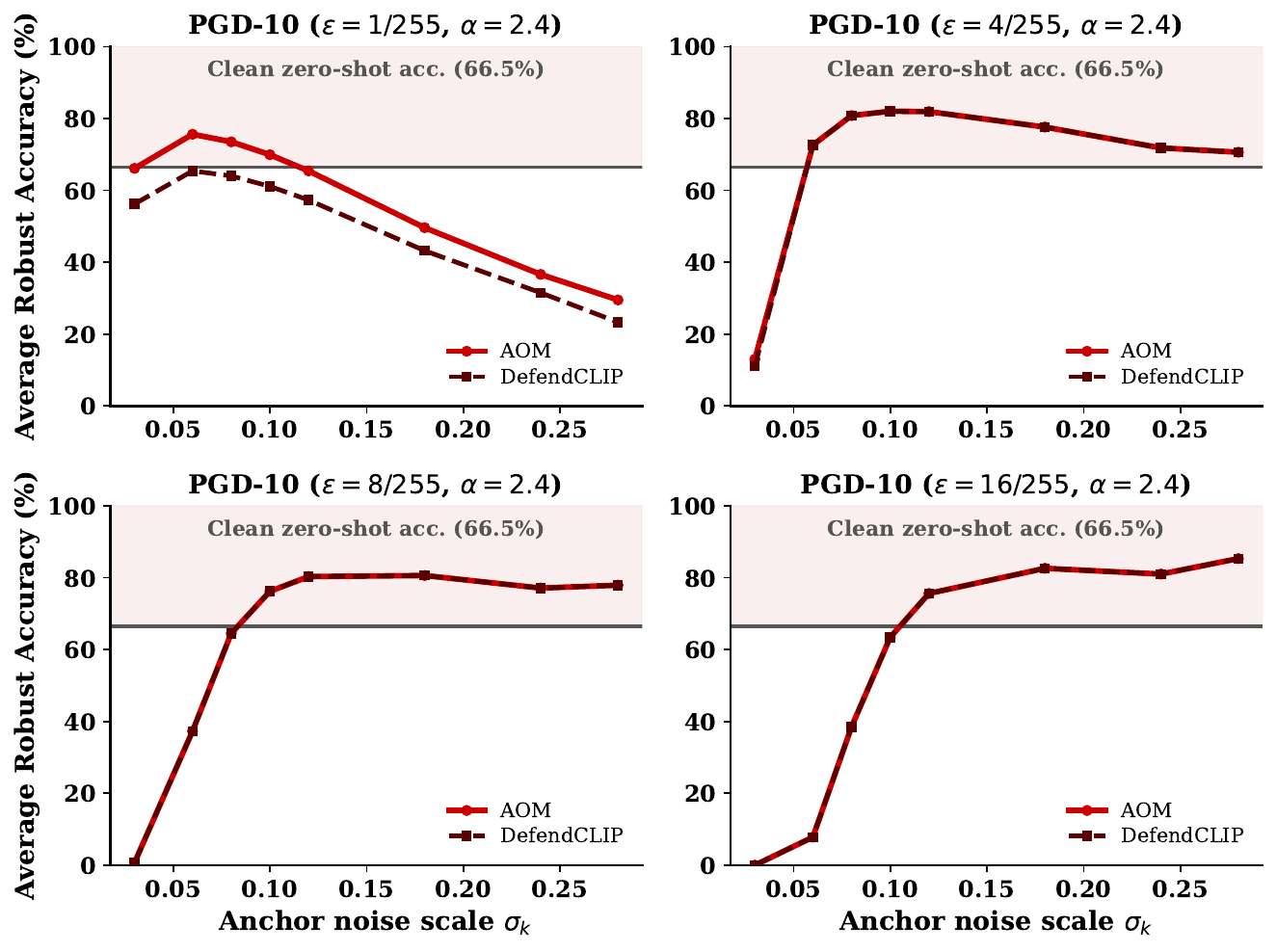}
    \caption{\textbf{Robust accuracy under leakage against $\sigma_k$}
    averaged across 12 downstream datasets, at $\alpha{=}2.4$. Reported
    accuracy reaches $73.5$, $81.9$, $80.6$, and $85.3\%$ at
    $\epsilon{=}1$, $4$, $8$, and $16/255$, above the $66.5\%$ clean
    ceiling and highest under the strongest attack, which is
    impossible for a genuine defense of a $66.5\%$ classifier.}
    \label{fig:leakage_reported_sigma}
\end{figure}

\begin{figure}[!t]
    \centering
    \includegraphics[width=\linewidth]{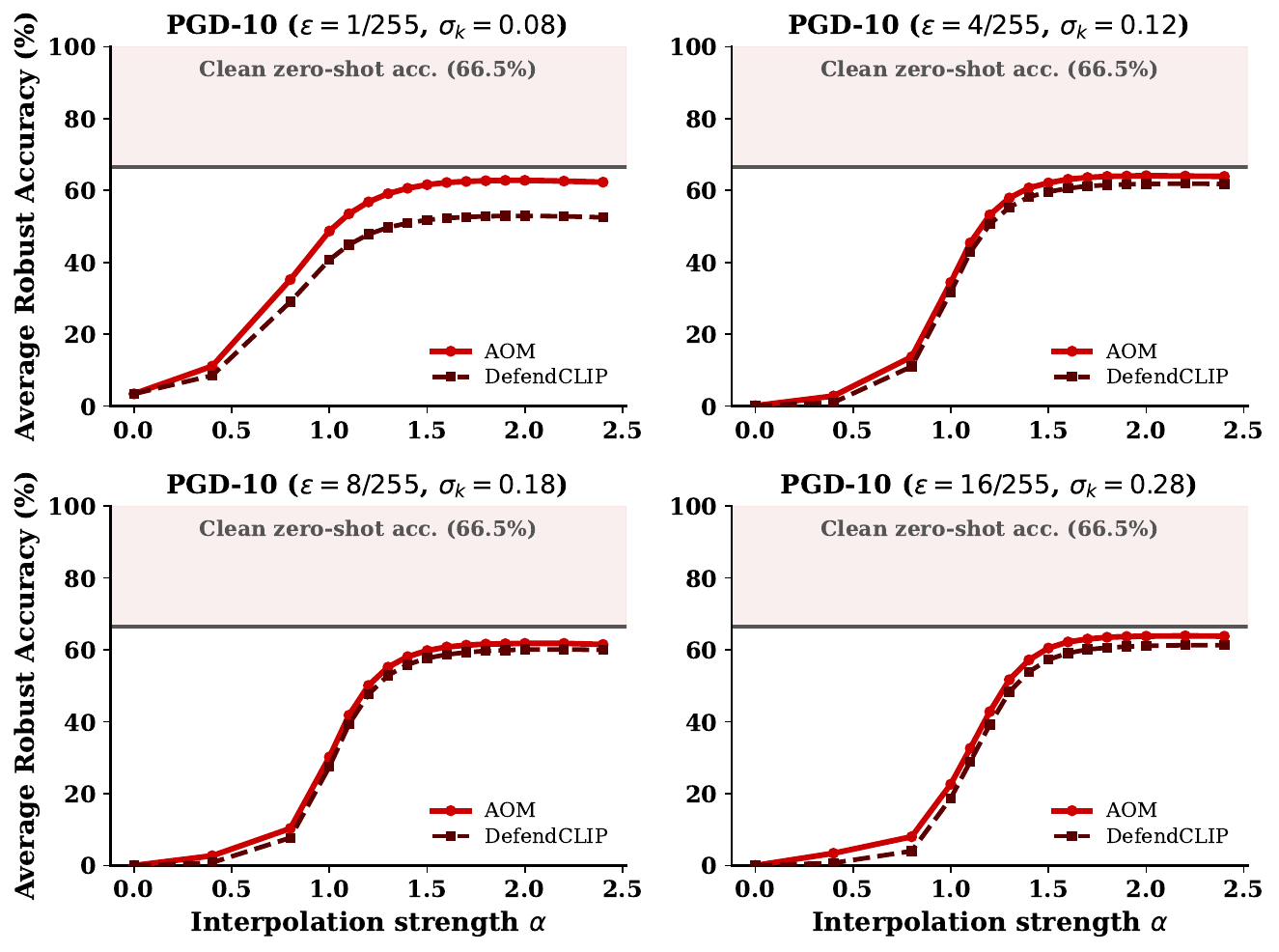}
    \caption{\textbf{Robust accuracy against $\alpha$ without
    leakage}, averaged across 12 downstream datasets.}
    \label{fig:leakage_corrected_alpha}
\end{figure}

\begin{figure}[!t]
    \centering
    \includegraphics[width=\linewidth]{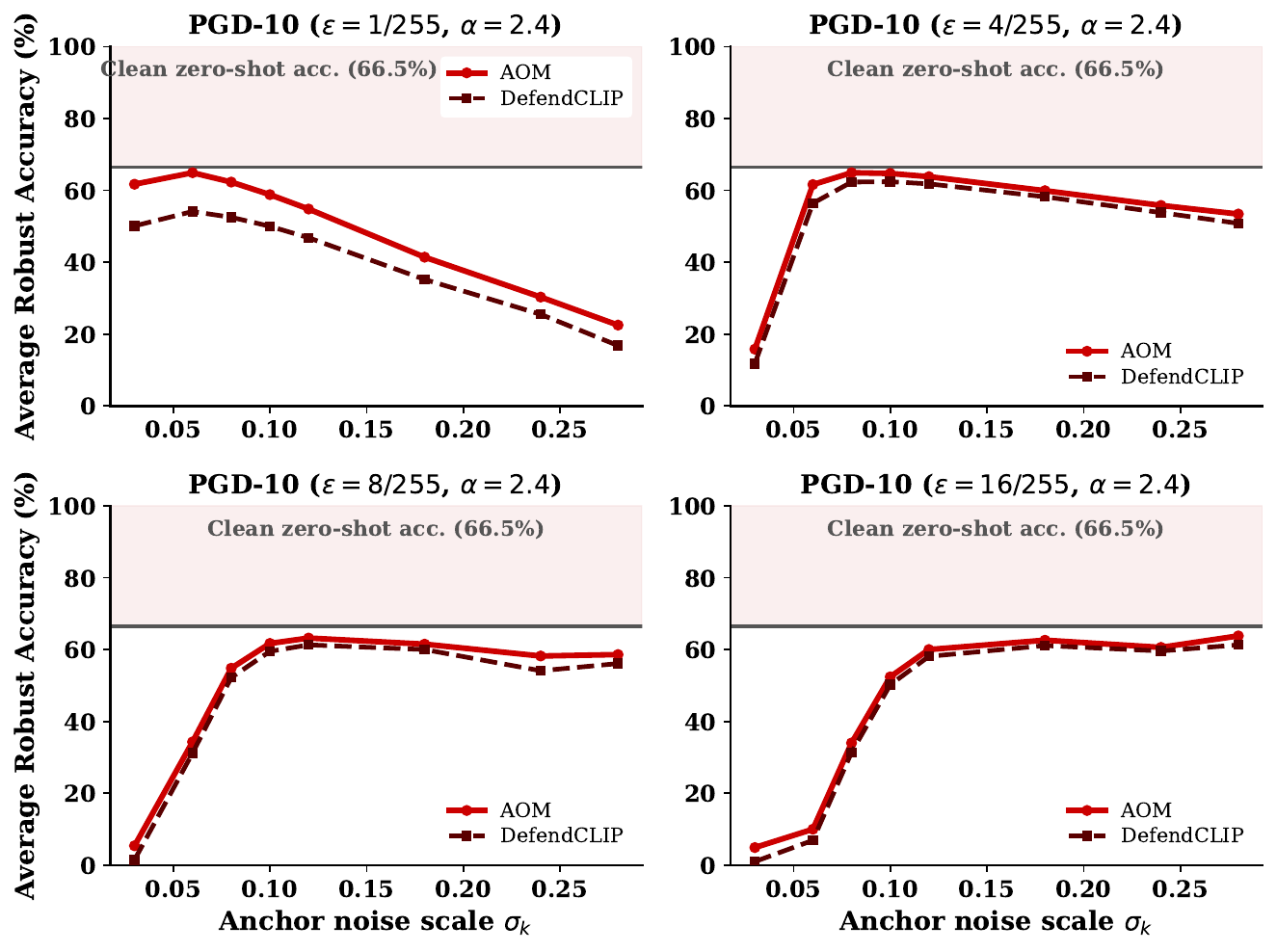}
    \caption{\textbf{Robust accuracy against $\sigma_k$ without
    leakage}, averaged across 12 downstream datasets.}
    \label{fig:leakage_corrected_sigma}
\end{figure}

\begin{figure}[!t]
    \centering
    \includegraphics[width=\linewidth]{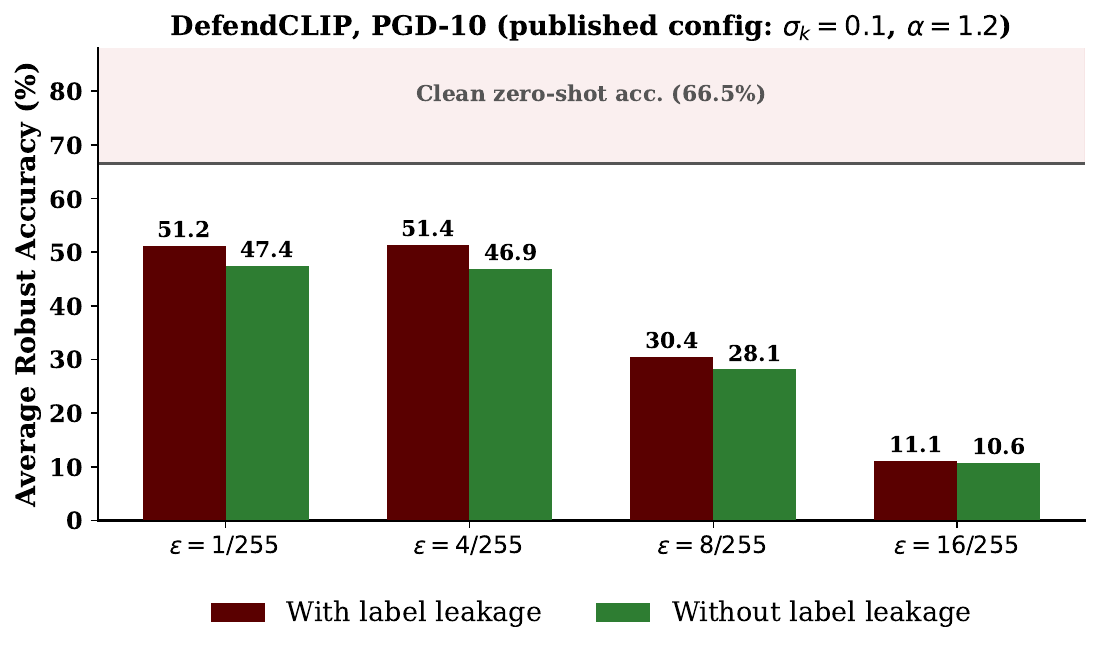}
    \caption{\textbf{Robust accuracy at the published configuration,
    with and without label leakage.} DefendCLIP at its published
    configuration ($\sigma_k{=}0.1$, $\alpha{=}1.2$), pooled across 12 downstream datasets (dark red) and on correctly-classified-only
    samples (green), across all four budgets. Reported robustness is
    inflated by $3.7$ and $4.5\%$ at $\epsilon{=}1$ and $4/255$, the
    regime the configuration was tuned for, and by $2.3$ and $0.5\%$
    at $\epsilon{=}8$ and $16/255$, where the fixed anchor scale is
    too weak for leakage to relocate to: the artifact is present at
    the deployed configuration itself, not only at aggressive
    settings.}
    \label{fig:leakage_accuracy_contrast}
\end{figure}

\begin{figure}[!t]
    \centering
    \includegraphics[width=\linewidth]{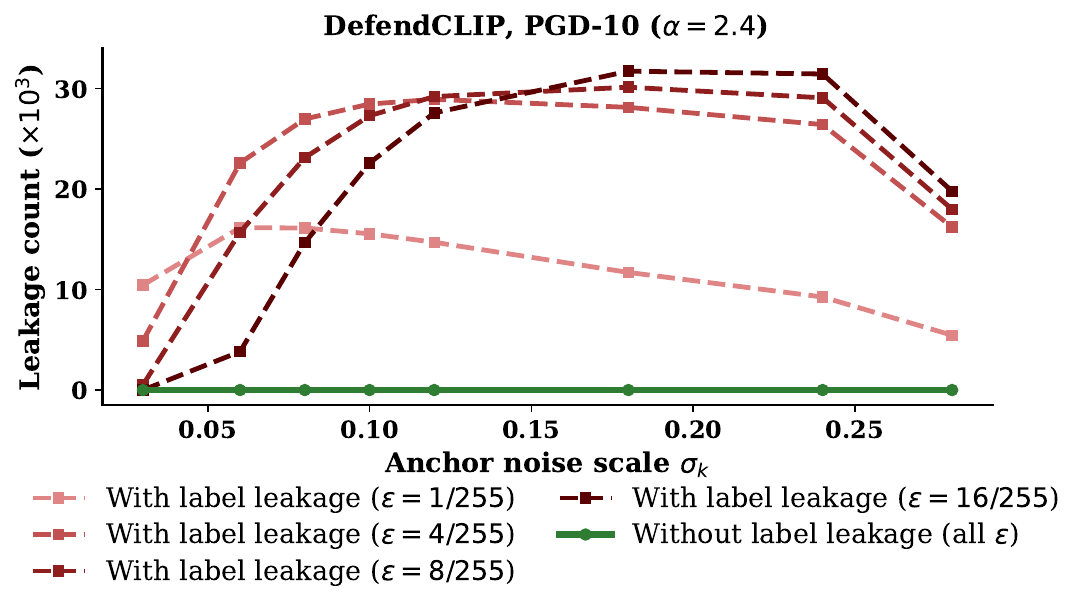}
    \caption{\textbf{Leakage count with and without the attack.}
    DefendCLIP's leakage count under attack (dashed, per budget
    $\epsilon$), pooled across 12 downstream datasets, against the count
    observed when the same clean, misclassified samples are passed
    directly through the correction operator with no attack applied
    (solid green). Because no adversarial perturbation is involved, the
    green curve is attack-agnostic and identical across all $\epsilon$.
    It remains close to zero at every anchor scale, with only a small
    handful of samples flipped correct, consistent with the incidental
    effect of averaging over noise-perturbed views rather than any
    systematic correction. This near-zero baseline stands in sharp
    contrast to the tens of thousands of samples leaked once the same
    misclassified samples are attacked before correction is applied.}
    \label{fig:leakage_contrast}
\end{figure}

\paragraph{Leakage counts.}
Figures~\ref{fig:leakage_counts_alpha} and~\ref{fig:leakage_counts_sigma}
measure the artifact directly. For each configuration, every sample in
the 12 downstream datasets is attacked, regardless of whether the
undefended model classified it correctly beforehand, and the defensive
correction is then applied. The leakage count is simply the number of
samples that were already misclassified before attack but come out
correct after the attack and defensive correction are both applied.

Why this happens follows directly from how the attack is constructed.
Untargeted PGD (Eq.~\ref{eq:app_pgd_iterate}) ascends the loss with
respect to the \emph{true} label $y$, regardless of what the model
predicted on the clean input. This means the perturbation always moves
the feature away from the true class, and in doing so implicitly encodes
the direction toward the true class in the perturbation itself, whether
or not the clean sample started out correctly classified. Applying the
correction operator directly to a clean, already-misclassified sample,
with no attack involved, does not produce this effect: the correction
has no attack-induced direction to reverse, so it does not systematically
relocate such samples into the correct class. It is only once these same
misclassified samples are attacked, and thereby acquire this embedded
true-class direction, that applying the correction operator afterward
starts moving a large fraction of them into the correct class. This
asymmetry is the direct evidence for our hypothesis: the leakage
observed in Figures~\ref{fig:leakage_counts_alpha}
and~\ref{fig:leakage_counts_sigma} is not an incidental side effect of a
strong correction, but a consequence of the correction operator reversing
a true-class direction that the attack itself planted, on samples that
were never displaced from the correct class to begin with.

Figure~\ref{fig:leakage_counts_alpha} sweeps the interpolation strength
$\alpha$ at each attack budget's optimal $\sigma_k$. Leakage is zero when
no correction is applied ($\alpha{=}0$), as expected, since the attack
alone only pushes the feature further from the correct class and cannot
by itself produce a correct prediction. As $\alpha$ increases, leakage
grows sharply, since a stronger correction moves the feature further
along the embedded true-class direction described above. AOM and
Defend-CLIP show largely the same behavior, differing only slightly at
the weakest budget, where Defend-CLIP's intervention gate leaves a small
number of inputs uncorrected; at every other budget the two methods leak
essentially the same number of samples.

Figure~\ref{fig:leakage_counts_sigma} instead fixes $\alpha$ and sweeps
the anchor-noise scale $\sigma_k$. Leakage again follows an inverted U: a
very small scale barely moves the feature, and a very large scale washes
out the true-class direction the attack embedded, so leakage peaks at an
intermediate scale. This peak shifts to larger $\sigma_k$ as the attack
strengthens, mirroring the same budget-dependent scale already seen for
genuine correction in Figure~\ref{fig:sigma_tradeoff}.

What matters most in both figures is simply the scale of the effect. The
12-dataset pool contains over $140{,}000$ test images
(Table~\ref{tab:app_datasets}), and roughly a third of them are
misclassified by frozen, undefended CLIP to begin with
(Table~\ref{tab:main_results_avg}). At its peak, the correction step
alone flips roughly two-thirds of this originally-misclassified
population into "correct" predictions — samples the defense never
genuinely corrected, since they were never displaced from the correct
class by the attack in the first place. This is the core problem with
attacking every sample regardless of its clean prediction: a large pool
of already-wrong samples sits available to be swept into the "robust"
count by the correction operator alone, for both defenses studied here.
The leakage-free protocol used throughout this paper avoids this
entirely by never attacking these samples, so they have no opportunity
to leak.

\paragraph{Robust accuracy with leakage.}
Figures~\ref{fig:leakage_reported_alpha}
and~\ref{fig:leakage_reported_sigma} translate the leakage counts above
into their effect on the reported metric itself. In both sweeps, once
correction is applied strongly enough, reported robust accuracy rises
past the frozen model's own clean accuracy of $66.5\%$ and continues to
climb well above it, reaching as high as $85.3\%$, and this peak occurs
under the \emph{strongest} attack rather than the weakest. Both symptoms
are diagnostic on their own: a defended model cannot be more accurate
under attack than the same model is on clean data, and a defense cannot
become more accurate as the attack grows stronger. Their presence here
confirms that the reported gains include the leaked wins quantified
above rather than genuine robustness alone.

Figure~\ref{fig:leakage_accuracy_contrast} evaluates Defend-CLIP
at its own published configuration under both scorings, and reported
robustness is still inflated at the exact deployed setting, not only at
the extremes explored above. This inflation tracks the fraction of
clean-misclassified samples in each dataset, largest on datasets where
zero-shot CLIP already performs poorly and the most already-wrong
samples are available to leak — precisely the pattern expected if the
spurious gains originate from $\mathcal{D}_{\mathrm{wrong}}$ rather than
from any genuine corrective behavior. All evaluations in this paper
therefore use the leakage-free protocol throughout, attacking only
samples the frozen model classifies correctly before attack; we
re-evaluate every baseline in Table~\ref{tab:main_results_avg}, as well
as \method{} itself, under this same protocol.

\paragraph{Robust accuracy without leakage.}
Figures~\ref{fig:leakage_corrected_alpha}
and~\ref{fig:leakage_corrected_sigma} rescore the identical runs so that
only genuinely defended samples can count: robustness is credited only
on inputs the frozen model already classified correctly before attack,
so the leaked wins quantified in
Figures~\ref{fig:leakage_counts_alpha}--\ref{fig:leakage_counts_sigma}
are excluded by construction. Every symptom of leakage disappears at
once. No configuration, at any $\alpha$ or $\sigma_k$, exceeds the clean
ceiling of $66.5\%$, restoring the basic requirement that a defended
model cannot outperform the same model on clean data.

Figure~\ref{fig:leakage_contrast} isolates the role of the attack
directly, and gives the clearest evidence for the hypothesis stated
above. The green curve applies the correction operator to the same pool
of clean, misclassified samples used throughout this appendix, but with
no adversarial perturbation applied first: these samples are simply
corrected as they are. Because no attack is involved, this curve does
not depend on $\epsilon$ at all, which is why it appears as a single
line rather than one per budget. The resulting count stays close to
zero across the full range of $\sigma_k$: a small number of samples do
flip to correct, which is expected, since averaging over $M$
noise-perturbed views (Eq.~\ref{eq:app_anchor}) is itself a form of
test-time ensembling that can occasionally nudge a borderline sample
across the decision boundary on its own, independent of any attack. What
it rules out is any systematic tendency of the correction operator to
repair misclassified inputs by itself.

The contrast with the dashed curves is therefore the key result. These
use the exact same pool of clean, misclassified samples, but attack them
first, before applying the identical correction operator, and the count
jumps from near-zero to the tens of thousands reported earlier in this
section. Since the only difference between the two settings is whether
the attack is applied before correction, this comparison directly
confirms that leakage is not a property of the correction operator in
isolation: it requires the true-class direction that the attack itself
embeds into the perturbation (as argued in the "Leakage counts"
paragraph above), and only appears once that direction is present for
the correction to reverse.

\section{Algorithmic Details of \method{}}
\label{sec:appendix-algorithm}

Section~\ref{sec:response-policy} introduces two per-input decisions that
together define \method{}: a correction-demand signal that determines
\emph{how strongly} an input should be corrected, and a fused intervention
score that determines \emph{whether} correction should be applied at all.
This appendix makes both decisions precise and reproducible. We first give
the  procedure that calibrates the correction-strength mapping
underlying the first decision (Algorithm~\ref{alg:calibration}), and then
give the test-time procedure that uses this calibrated mapping, together
with the intervention score, to classify each input (Algorithm~\ref{alg:react-clip}).

\subsection{Notation}
\label{sec:appendix-algorithm-notation}

Table~\ref{tab:appendix-notation} summarizes the symbols used throughout
this appendix. All symbols are defined identically to their first use in
Section~\ref{sec:response-policy}.

\begin{table}[!htbp]
\centering
\small
\renewcommand{\arraystretch}{1.15}
\begin{tabular}{l p{5.7cm}}
\toprule
\bfseries Symbol & \bfseries Meaning \\
\midrule
$x$ & Test input image; clean/adversarial status and, if adversarial, its
perturbation budget are unknown to the defense \\
$F_v(\cdot),F_t(\cdot)$ & Frozen CLIP visual and text encoders \\
$f_v(x)$ & $\ell_2$-normalized visual feature of $x$ (Eq.~\ref{eq:clip-features}) \\
$f_t^k$ & $\ell_2$-normalized text feature for class $k$, precomputed once \\
$\sigma_{\mathrm{low}}{<}\sigma_{\mathrm{high}}$ & Fixed weak Gaussian probe scales (main config: $0.02$, $0.05$) \\
$d_{\mathrm{low}}(x), d_{\mathrm{high}}(x)$ & Feature drift induced by each probe scale (Eq.~\ref{eq:probe-drift}) \\
$r(x)$ & Relative cross-noise drift: the correction-demand signal (Eq.~\ref{eq:relative-drift}) \\
$a,b$ & Calibrated linear-mapping coefficients (main config: $0.03$, $0.042$) \\
$\sigma_{\mathrm{corr}}(x)$ & Sample-specific anchor-noise scale, $a+b\,r(x)$ (Eq.~\ref{eq:linear-map}) \\
$M$ & Number of noise samples used to build the anchor (main config: $10$) \\
$f_{\mathrm{anc}}(x)$ & Response-conditioned, noise-averaged feature anchor (Eq.~\ref{eq:adaptive-anchor}) \\
$\alpha$ & Extrapolation strength (main config: $2.0$; $\alpha{>}1$ moves the feature past the anchor) \\
$\mathcal T(\cdot)$ & Weak stochastic augmentation (affine + Gaussian blur + additive noise + color jitter) \\
$J(x)$ & Prediction-instability score, JS divergence between $p(\cdot\mid x)$ and $p(\cdot\mid\mathcal T(x))$ (Eq.~\ref{eq:js-score}) \\
$s_{\mathrm{int}}(x)$ & Defensive intervention score, $r(x){+}J(x)$ (Eq.~\ref{eq:intervention-score}) \\
$\tau$ & Intervention threshold (main config: $0.7$) \\
$g(x)$ & Correction gate, $\mathbb 1[s_{\mathrm{int}}(x){\ge}\tau]$ \\
$f_{\mathrm{out}}(x)$ & Final defended feature used for classification (Eq.~\ref{eq:final-feature}) \\
\bottomrule
\end{tabular}
\caption{Notation used in Algorithms~\ref{alg:calibration} and \ref{alg:react-clip}.}
\label{tab:appendix-notation}
\end{table}

\subsection{Calibration}
\label{sec:appendix-algorithm-calibration}

Before \method{} can be run at test time, the coefficients $(a,b)$ of the
linear correction-strength mapping $\sigma_{\mathrm{corr}}(x)=a+b\,r(x)$
must be fit once. Algorithm~\ref{alg:calibration} details this procedure,
which underlies Figure~\ref{fig:drift-calibration}~\emph{(main paper)} in the main text: for
each of four calibration budgets, adversarial examples are crafted against
frozen CLIP, the average relative drift $\bar r_\epsilon$ is measured, and
an anchor-noise scale $\sigma_\epsilon^\star$ is swept to find the value
that maximizes defended accuracy at that budget. The four resulting
$(\bar r_\epsilon,\sigma_\epsilon^\star)$ pairs are then fit with a linear
regression to obtain $a$ and $b$. This procedure is run once, offline, on
a single calibration dataset (Caltech256), and the resulting $(a,b)$ are
then frozen and reused unchanged by Algorithm~\ref{alg:react-clip} for
every subsequent dataset and attack budget.

\begin{algorithm*}[!htbp]
\DontPrintSemicolon
\LinesNumbered
\SetKwBlock{CalPhaseOne}{\textcolor{oursaccent}{\bfseries Phase 1 --- Adversarial Calibration Set Construction}}{}
\SetKwBlock{CalPhaseTwo}{\textcolor{oursaccent}{\bfseries Phase 2 --- Drift Measurement per Budget}}{}
\SetKwBlock{CalPhaseThree}{\textcolor{oursaccent}{\bfseries Phase 3 --- Anchor-Scale Sweep}}{}
\SetKwBlock{CalPhaseFour}{\textcolor{oursaccent}{\bfseries Phase 4 --- Linear Fit}}{}
\caption{Offline Calibration of the Correction-Strength Mapping (run once)}
\label{alg:calibration}
\KwIn{Calibration set $\mathcal D$ (Caltech256); budgets
$\mathcal E{=}\{1,4,8,16\}/255$; PGD-10 generator; fixed extrapolation
strength $\alpha$; candidate anchor scales $\Sigma$}
\KwOut{Coefficients $(a,b)$ such that $\sigma_{\mathrm{corr}}(x)=a+b\,r(x)$}

\ForEach{$\epsilon \in \mathcal E$}{

\CalPhaseOne{
craft adversarial set $\mathcal D_\epsilon$ via PGD-10 at budget $\epsilon$
against frozen CLIP\;
}

\CalPhaseTwo{
compute $r(x)$ for all $x\in\mathcal D_\epsilon$\;
$\bar r_\epsilon \gets \mathrm{mean}_{x\in\mathcal D_\epsilon}\, r(x)$\;
}

\CalPhaseThree{
\ForEach{$\sigma_a\in\Sigma$}{
    apply the fixed-anchor correction  with anchor scale $\sigma_a$
    and extrapolation strength $\alpha$ to $\mathcal D_\epsilon$\;
    record robust classification accuracy $\mathrm{Acc}(\sigma_a;\epsilon)$\;
}
$\sigma_\epsilon^\star \gets \arg\max_{\sigma_a\in\Sigma}\mathrm{Acc}(\sigma_a;\epsilon)$\;
store calibration pair $(\bar r_\epsilon,\sigma_\epsilon^\star)$\;
}
}

\CalPhaseFour{
\colorbox{oursbg}{fit $\sigma_{corr}^\star \approx a+b\,\bar r_\epsilon$ by linear regression over $\{(\bar r_\epsilon,\sigma_\epsilon^\star)\}_{\epsilon\in\mathcal E}$}\;
\Return $(a,b)$\;
}
\end{algorithm*}

\subsection{Inference Procedure}
\label{sec:appendix-algorithm-inference}

Algorithm~\ref{alg:react-clip} is the procedure \method{} runs on every
test input, using the coefficients $(a,b)$ fixed by
Algorithm~\ref{alg:calibration}. It operates entirely on the frozen encoder
$F_v$ and requires no backward pass, no adversarial-example generation,
and no knowledge of the perturbation budget used to construct $x$. The
only quantities fixed in advance are the probe scales
$\sigma_{\mathrm{low}},\sigma_{\mathrm{high}}$, the calibrated coefficients
$(a,b)$, the extrapolation strength $\alpha$, and the intervention
threshold $\tau$; all four are set once and held fixed across datasets and
attack budgets. After a trivial initialization step (Phase~0), the
procedure is grouped into four phases: (i) noise-probed drift estimation,
which yields the correction-demand signal $r(x)$; (ii) response-conditioned
anchor construction, which turns $r(x)$ into a sample-specific anchor;
(iii) the selective intervention decision, which fuses $r(x)$ with the
prediction-instability score $J(x)$; and (iv) the conditional correction
and final classification.

\begin{algorithm*}[!htbp]
\DontPrintSemicolon
\LinesNumbered
\SetKwBlock{PhaseZero}{\textcolor{oursaccent}{\bfseries Phase 0 --- Initialization}}{}
\SetKwBlock{PhaseOne}{\textcolor{oursaccent}{\bfseries Phase 1 --- Noise-Probed Drift Estimation}}{}
\SetKwBlock{PhaseTwo}{\textcolor{oursaccent}{\bfseries Phase 2 --- Response-Conditioned Anchor Construction}}{}
\SetKwBlock{PhaseThree}{\textcolor{oursaccent}{\bfseries Phase 3 --- Selective Intervention Decision}}{}
\SetKwBlock{PhaseFour}{\textcolor{oursaccent}{\bfseries Phase 4 --- Conditional Correction \& Classification}}{}
\caption{\method{}: Response-Aware Correction at Test Time (Inference)}
\label{alg:react-clip}
\KwIn{Test image $x$; class prompts $\{P_k\}_{k=1}^K$; frozen encoders
$F_v,F_t$; temperature $\gamma$; probe scales
$\sigma_{\mathrm{low}}{<}\sigma_{\mathrm{high}}$; anchor sample count $M$;
calibrated coefficients $(a,b)$; extrapolation strength $\alpha$; weak
augmentation $\mathcal T$; intervention threshold $\tau$}
\KwOut{Predicted label $\hat y(x)$}

\PhaseZero{
$f_t^k \gets F_t(P_k)/\lVert F_t(P_k)\rVert_2$ for $k=1,\dots,K$
$f_v(x) \gets F_v(x)/\lVert F_v(x)\rVert_2$\;
}

\PhaseOne{
\ForEach{$s \in \{\mathrm{low},\ \mathrm{high}\}$}{
    sample $\eta_s \sim \mathcal N(0,I)$\;
    $x^{(\sigma_s)} \gets x+\sigma_s\eta_s$\;
    $d_s(x) \gets \lVert f_v(x^{(\sigma_s)})-f_v(x)\rVert_2$\;
}
\colorbox{oursbg}{$r(x) \gets \dfrac{d_{\mathrm{high}}(x)-d_{\mathrm{low}}(x)}{d_{\mathrm{low}}(x)}$}
\tcp*{correction-demand signal}
}

\PhaseTwo{
\colorbox{oursbg}{$\sigma_{\mathrm{corr}}(x) \gets a+b\,r(x)$}
\tcp*{sample-specific anchor scale}
\For{$i=1$ \KwTo $M$}{
    sample $\eta_i \sim \mathcal N(0,I)$\;
    $x_i \gets x+\sigma_{\mathrm{corr}}(x)\,\eta_i$\;
    $f_v(x_i) \gets F_v(x_i)/\lVert F_v(x_i)\rVert_2$\;
}
$f_{\mathrm{anc}}(x) \gets \dfrac{1}{M}\sum_{i=1}^M f_v(x_i)$
\tcp*{response-conditioned anchor}
}

\PhaseThree{
$\tilde x \gets \mathcal T(x)$\;
$p_x \gets p(\cdot\mid x)$;\quad $p_{\mathcal T}\gets p(\cdot\mid \tilde x)$
$m \gets \tfrac12(p_x+p_{\mathcal T})$\;
$J(x) \gets \tfrac12\mathrm{KL}(p_x\Vert m)+\tfrac12\mathrm{KL}(p_{\mathcal T}\Vert m)$\;
\colorbox{oursbg}{$s_{\mathrm{int}}(x) \gets r(x)+J(x)$,\quad $g(x) \gets \mathbb 1[s_{\mathrm{int}}(x)\ge\tau]$}
\tcp*{intervention score \& gate}
}

\PhaseFour{
\colorbox{oursbg}{$f_{\mathrm{out}}(x) \gets f_v(x)+g(x)\,\alpha\,[f_{\mathrm{anc}}(x)-f_v(x)]$}
\tcp*{conditional extrapolation}
$p(y{=}k\mid x) \gets \dfrac{\exp\big(\mathrm{sim}(f_{\mathrm{out}}(x),f_t^k)/\gamma\big)}
{\sum_{j=1}^K\exp\big(\mathrm{sim}(f_{\mathrm{out}}(x),f_t^j)/\gamma\big)},\ \ k=1,\dots,K$
\Return $\hat y(x) \gets \arg\max_k p(y{=}k\mid x)$\;
}
\end{algorithm*}

{\small\noindent\textcolor{oursaccent}{\rule{0.28cm}{0.28cm}}~shaded lines
mark the quantities introduced by \method{} (the correction-demand signal
$r(x)$, the sample-specific scale $\sigma_{\mathrm{corr}}(x)$, the fused
intervention score $s_{\mathrm{int}}(x)$ and gate $g(x)$, and the
conditionally applied output feature $f_{\mathrm{out}}(x)$); unshaded lines
implement the noise-averaged feature-correction operator of Eqs.~4--6.\par}

\subsection{Computational Cost}
\label{sec:appendix-algorithm-cost}

Per test image, Algorithm~\ref{alg:react-clip} issues
$1{+}2{+}M{+}1=M{+}4$ forward passes through the frozen visual encoder
(clean feature, two probes, $M$ anchor views, one augmented view. With
the main configuration $M{=}10$, this is $14$ encoder forward passes per
image, none requiring gradient computation or backpropagation.

This cost closely matches the feature-correction baselines that
\method{} builds on. \method{} uses the same number of noise-averaged
views $M$ to construct the anchor as AOM and Defend-CLIP (Eqs.~4--5), and,
like Defend-CLIP, evaluates two additional noise levels to compute the
relative cross-noise drift used for its gate. The only additional cost
\method{} introduces beyond Defend-CLIP is a single forward pass on one
weakly augmented view of the input, used to compute the prediction-instability
score $J(x)$. Relative to AOM, which applies its correction unconditionally
and therefore does not require any drift computation, \method{} trades
these two extra probe passes and one augmented-view pass for the ability to
determine, per input, both how strongly to correct and whether to correct
at all.

More importantly, all forward passes in Algorithm~\ref{alg:react-clip} are
inference-only: \method{} requires no gradient computation, no
backpropagation through the encoder, and no iterative optimization at test
time. This distinguishes it from two other categories of test-time
defense. Counterattack methods such as TTC, DOC, and MAC construct their
corrective perturbation by backpropagating through the encoder to optimize
an input-space counterattack, incurring the cost of one or more
gradient-based optimization steps per image in addition to the forward
passes needed to evaluate them. Prompt-based methods such as SS-TPT
instead optimize the input text prompt itself at test time, which also
requires backpropagation and iterative updates per image. \method{}
requires neither: every quantity in Algorithm~\ref{alg:react-clip},
including the correction-demand signal $r(x)$, the response-conditioned
anchor $f_{\mathrm{anc}}(x)$, and the intervention score $s_{\mathrm{int}}(x)$,
is computed from forward passes alone, making \method{} both training-free
and optimization-free at test time.


\end{document}